\pdfoutput=1

\documentclass[final,times,12pt]{elsarticle}

\usepackage{amssymb}

\usepackage{times}
\usepackage{epsfig}
\usepackage{graphicx}
\usepackage{amsmath}
\usepackage{amssymb}
\usepackage{subfig}
\usepackage{array}
\usepackage{booktabs}
\usepackage{multirow}
\usepackage{makecell}
\usepackage{color}
\usepackage{hyperref}

\DeclareMathOperator*{\argmax}{argmax}
\DeclareMathOperator*{\argmin}{argmin}

\usepackage[normalem]{ulem}
\useunder{\uline}{\ul}{}

\journal{Computer Vision and Image Understanding}

\makeatletter
\def\ps@pprintTitle{%
	\let\@oddhead\@empty
	\let\@evenhead\@empty
	\def\@oddfoot{\footnotesize\itshape
		Accepted by \ifx\@journal\@empty Elsevier
		\else\@journal\fi\hfill\today}%
	\let\@evenfoot\@oddfoot}
\makeatother

\begin{document}

\begin{frontmatter}


\title{Learning a No-Reference Quality Metric\\ for Single-Image Super-Resolution}

\author[a1,a2]{Chao Ma}
\author[a2]{Chih-Yuan Yang}
\author[a1]{Xiaokang Yang}
\author[a2]{Ming-Hsuan Yang}
\address[a1]{Shanghai Jiao Tong University}
\address[a2]{University of California at Merced}
\fntext[label1]{C. Ma and X. Yang are with Institute of Image Communication and
	Network Engineering, Shanghai Jiao Tong University, Shanghai, 200240, P.R. China.
	E-mail: \{chaoma, xkyang\}@sjtu.edu.cn. 
	C. Ma was sponsored by China Scholarship Council and took a two-year study in University of California at Merced.}
\fntext[label2]{C.-Y. Yang and M.-H. Yang are with Electrical Engineering and Computer Science, University
	of California, Merced, CA, 95344. E-mail: yangchihyuan@gmail.com, mhyang@ucmerced.edu.}

\address{}

\begin{abstract}
Numerous single-image super-resolution algorithms have been proposed in the literature, but few studies address the problem of performance evaluation based on visual perception. While most super-resolution images are evaluated by full-reference metrics, the effectiveness is not clear and the required ground-truth images are not always available in practice. To address these problems, we conduct human subject studies using a large set of super-resolution images and propose a no-reference metric learned from visual perceptual scores. Specifically, we design three types of low-level statistical features in both spatial and frequency domains to quantify super-resolved artifacts, and learn a two-stage regression model to predict the quality scores of 
super-resolution images without referring to ground-truth images. Extensive experimental results show that the proposed metric is effective and efficient to assess the quality of super-resolution images based on human perception.

\end{abstract}


\end{frontmatter}


\section{Introduction}

\begin{figure}
	\centering
	\setlength{\tabcolsep}{.2em}
	\small
	\begin{tabular}{ccc}
		\includegraphics[width=.32\textwidth]{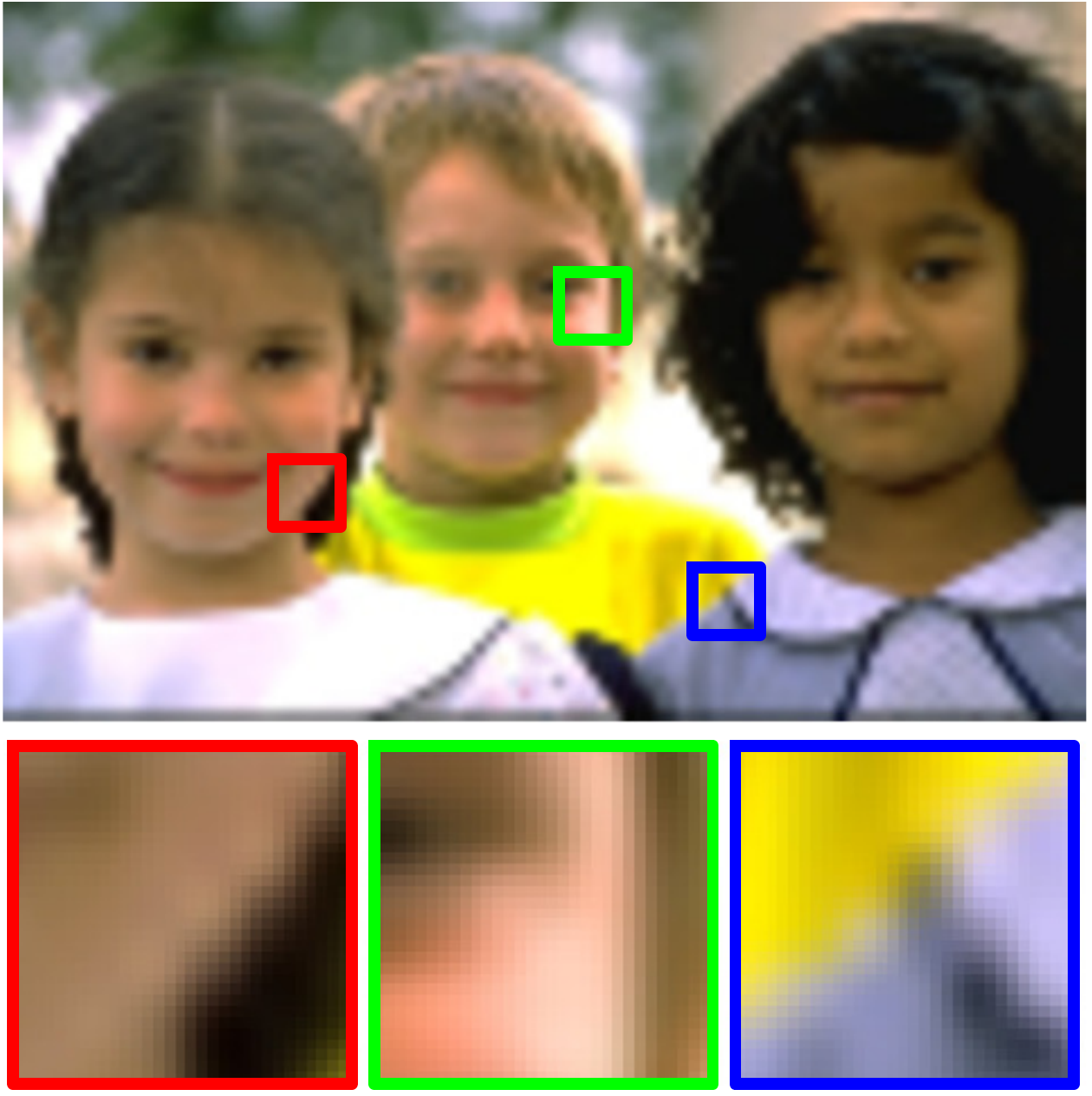}&
		\includegraphics[width=.32\textwidth]{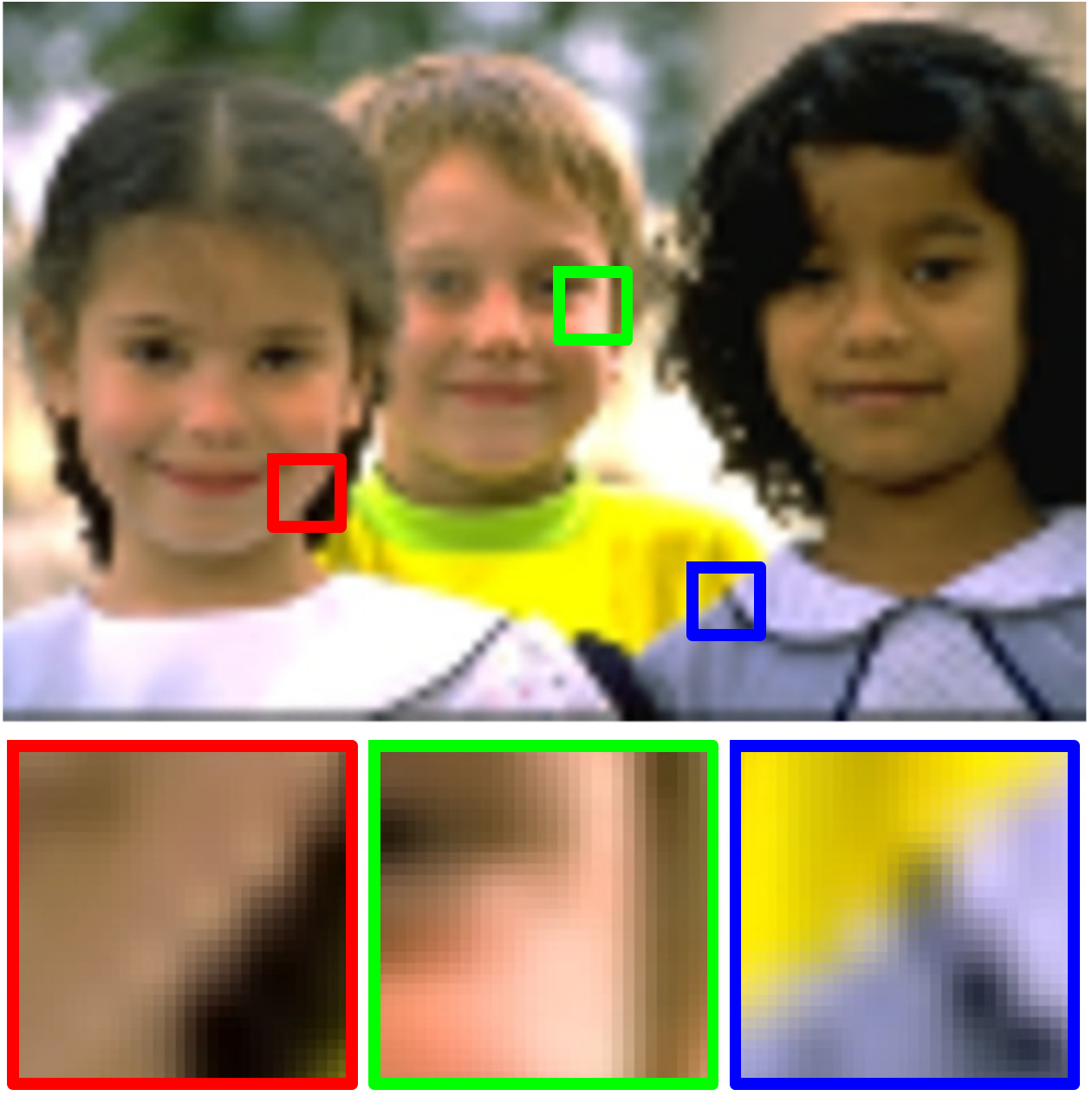} &
		\includegraphics[width=.32\textwidth]{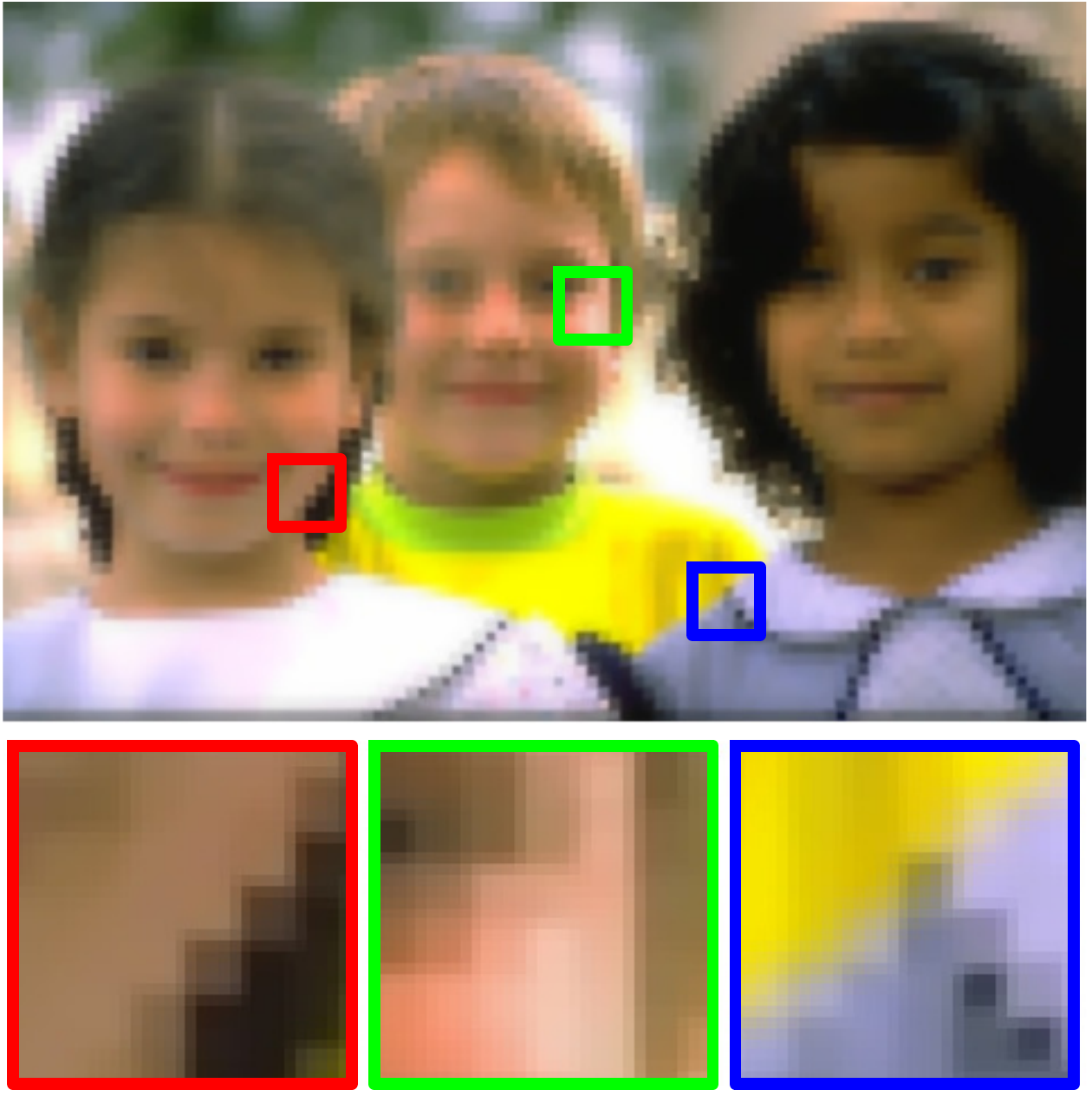}\\
		(a) Bicubic interpolation &
		(b) Back projection (BP)~\cite{DBLP:journals/cvgip/IraniP91} &
		(c) Shan08~\cite{DBLP:journals/tog/ShanLJT08} \\
		\includegraphics[width=.32\textwidth]{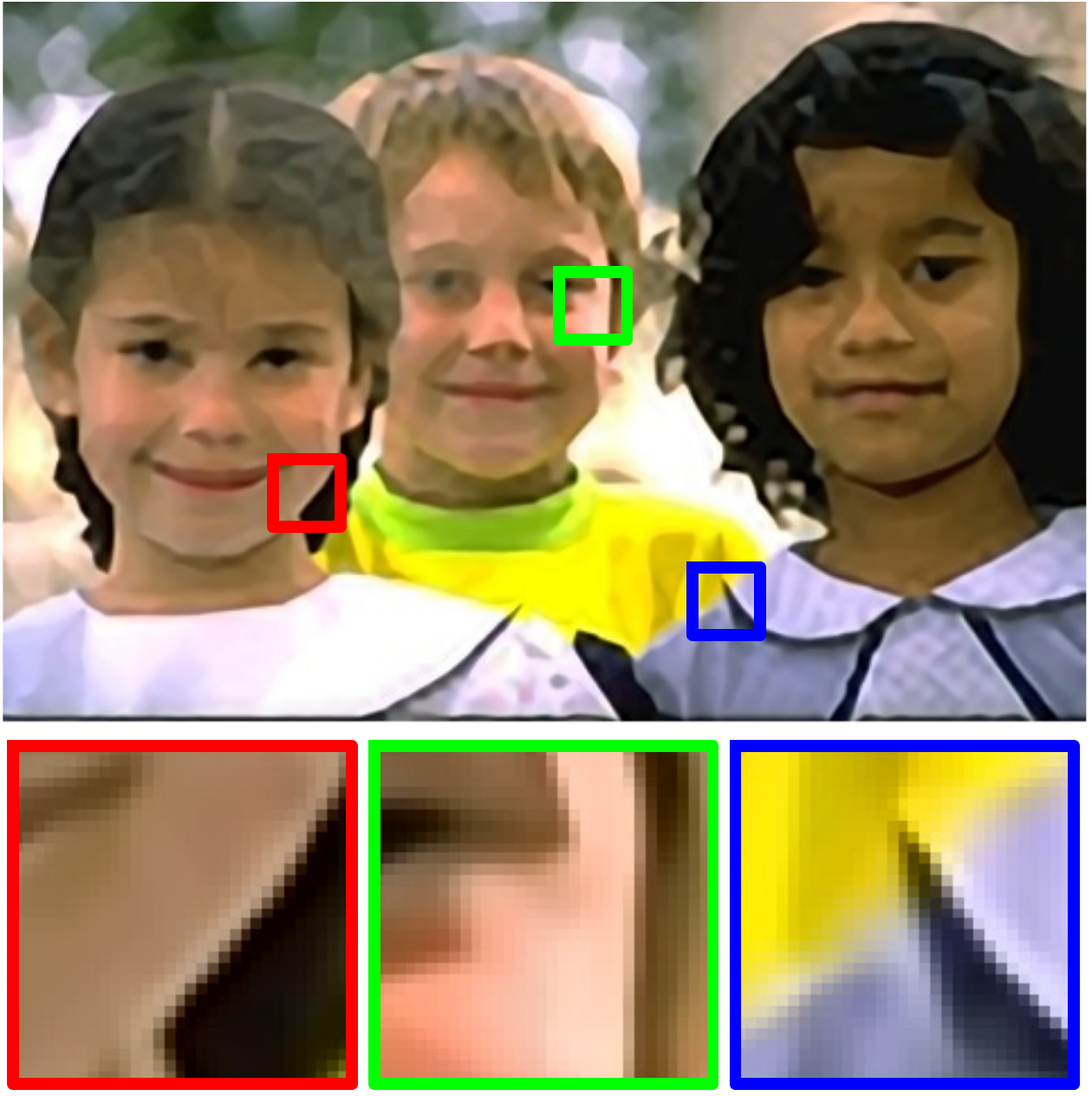}&
		\includegraphics[width=.32\textwidth]{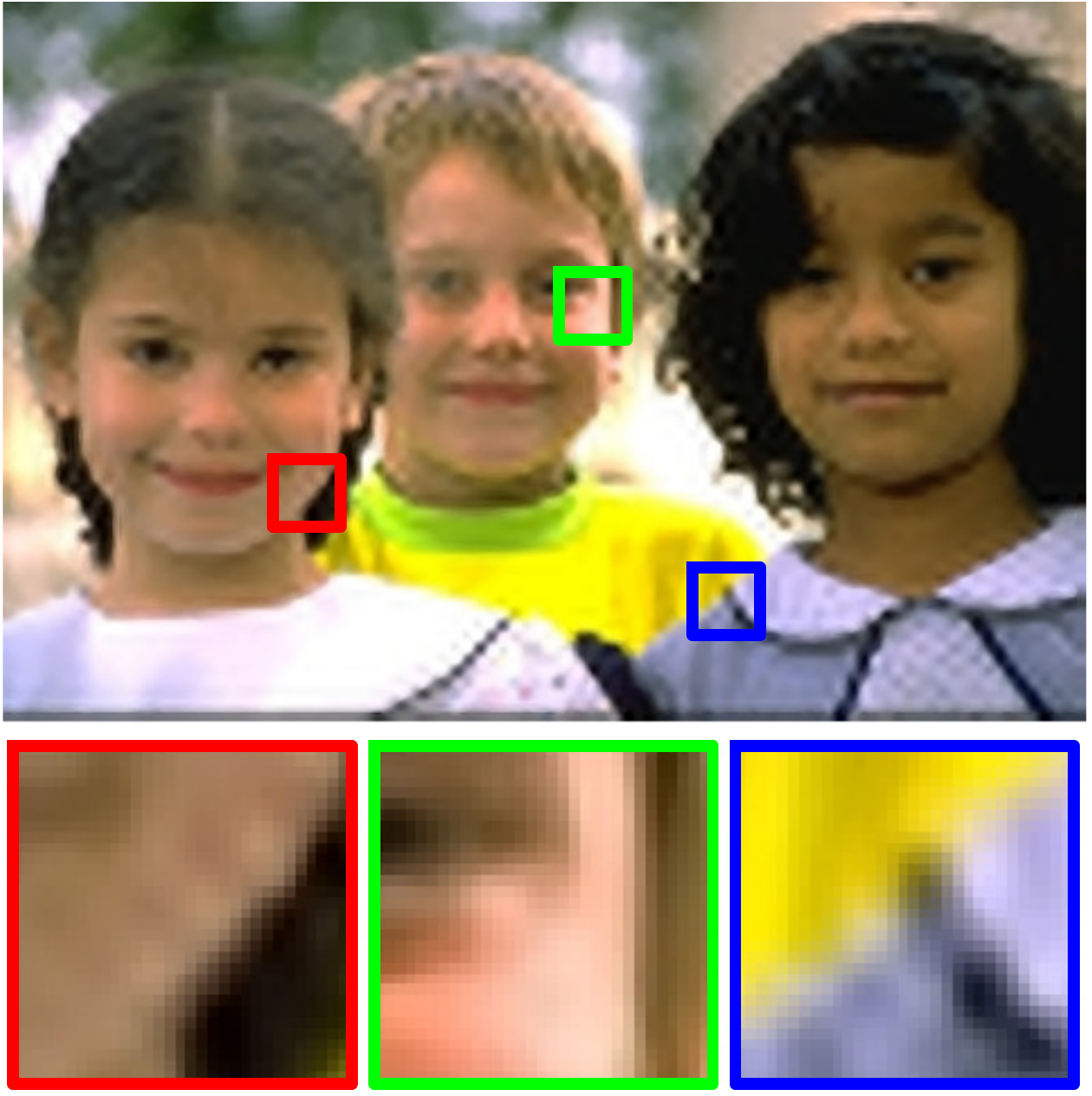}&
		\includegraphics[width=.32\textwidth]{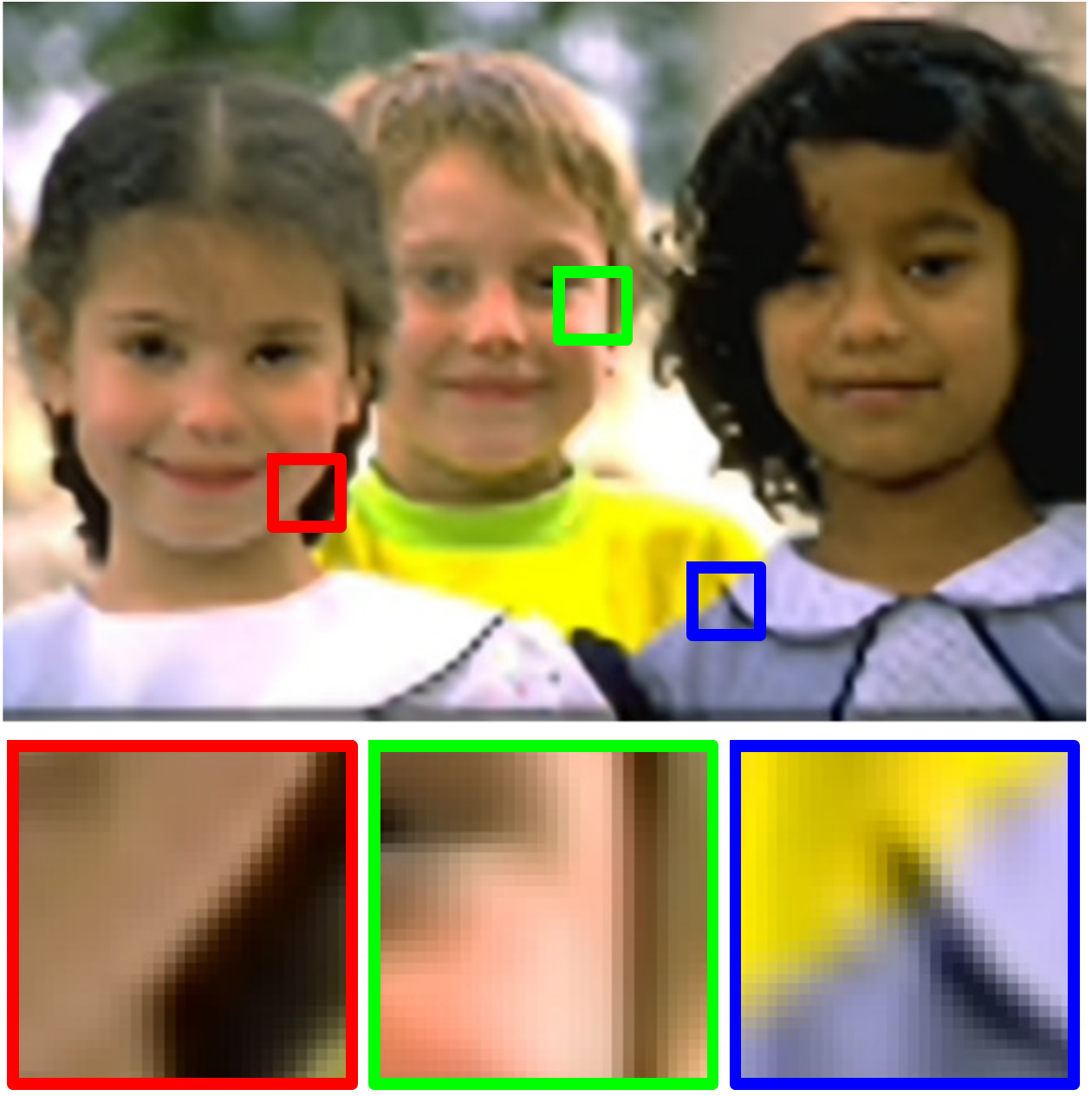}\\
		(d) Glasner09~\cite{Glasner2009} &
		(e) Yang10~\cite{DBLP:journals/tip/YangWHM10} &
		(f) Dong11~\cite{DBLP:journals/tip/DongZSW11} \\
		\includegraphics[width=.32\textwidth]{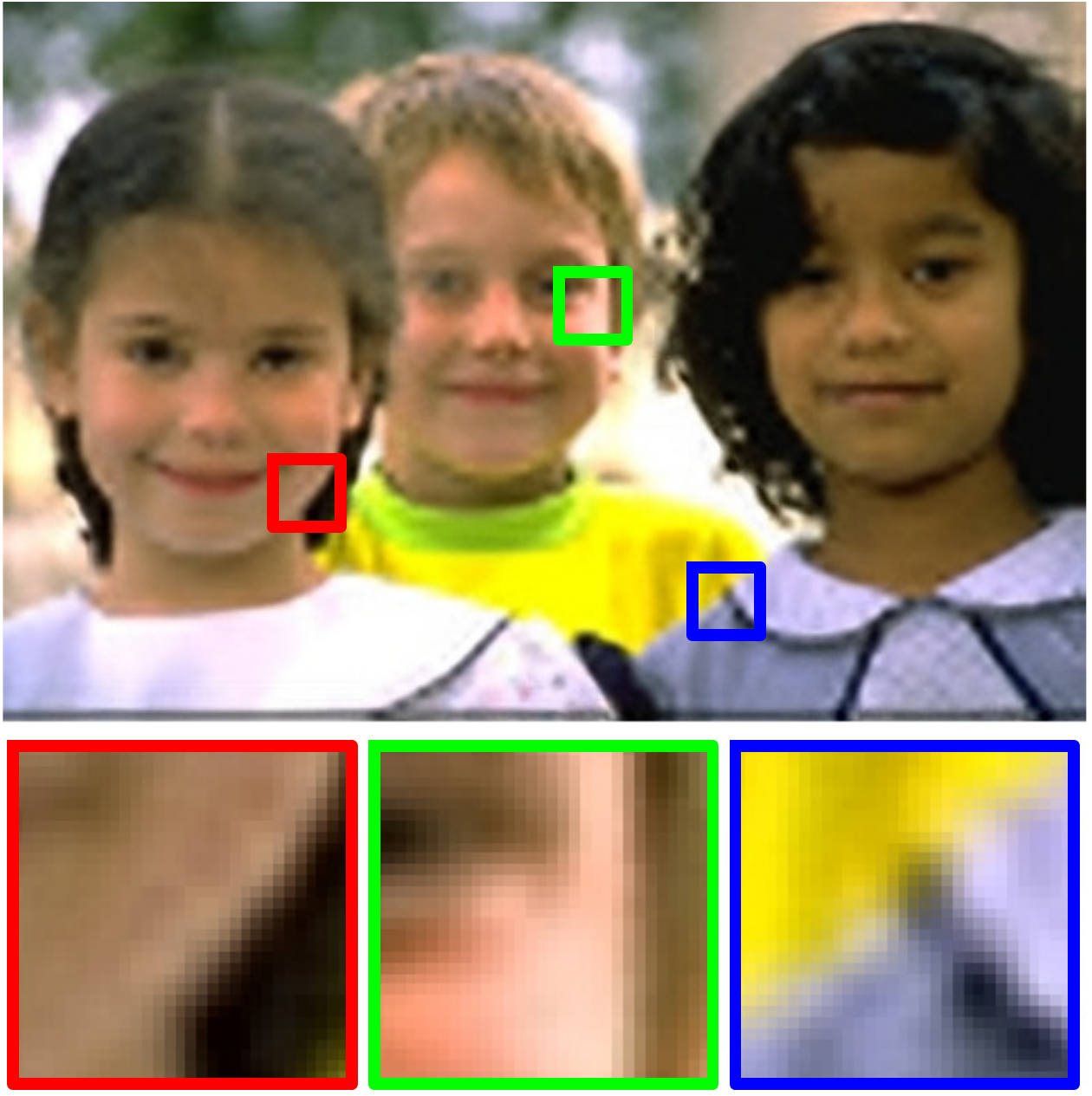}&
		\includegraphics[width=.32\textwidth]{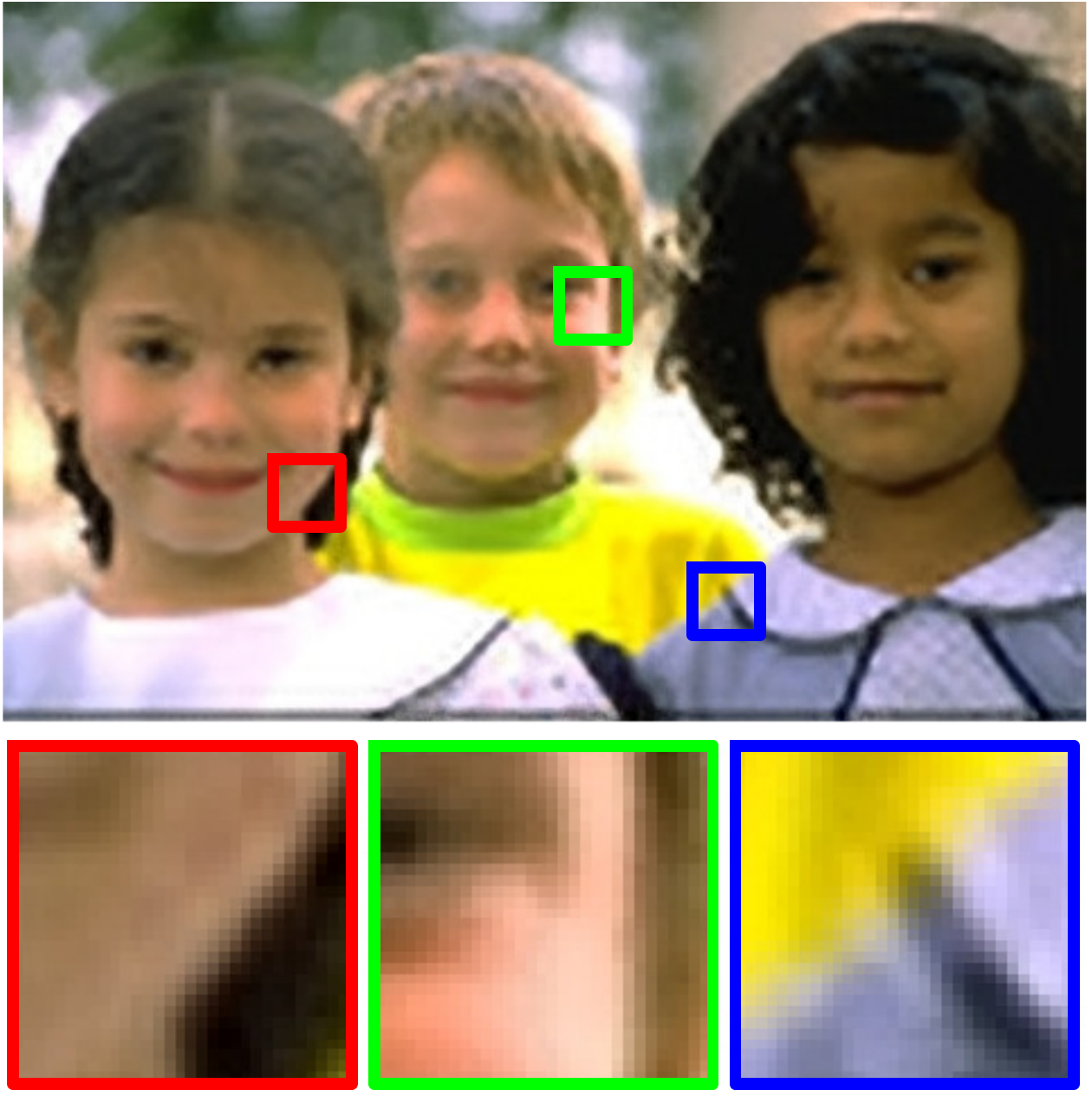}&
		\includegraphics[width=.32\textwidth]{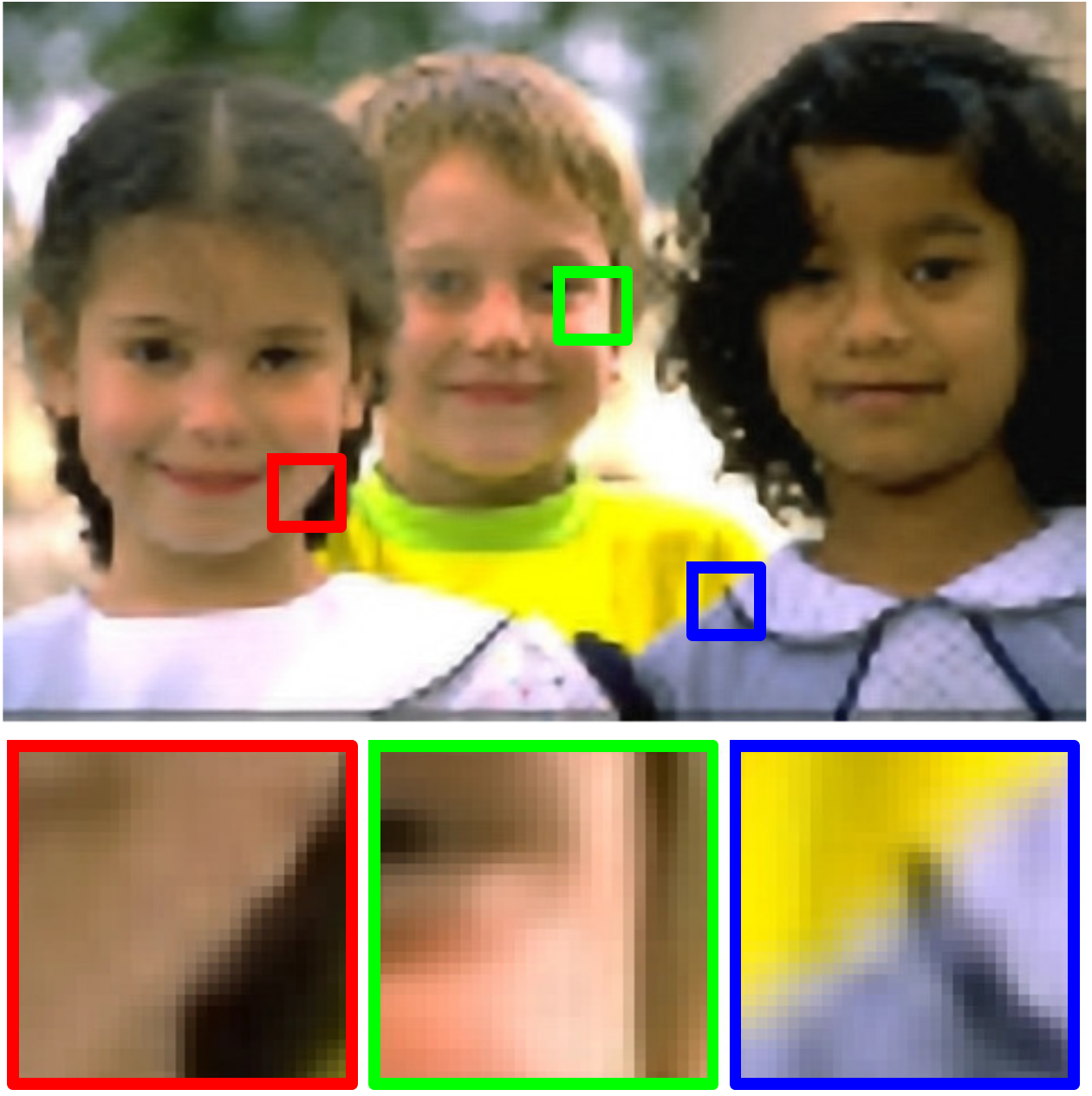}\\
		(g) Yang13~\cite{Yang13_ICCV_Fast}  &
		(h) Timofte13~\cite{DBLP:conf/iccv/TimofteDG13}  &
		(i) SRCNN~\cite{DBLP:conf/eccv/DongLHT14} \\
	\end{tabular}
	\caption{SR images generated from the same LR image using (\ref{eq:downsample}) ($s=4,\sigma=1.2$). The quality scores of these SR images are compared in Table \ref{tb:srscore}. The images are best viewed on a high-resolution displayer with an
		adequate zoom level, where each SR image is shown with at least 320$\times$480 pixels
		(full-resolution).
	}
	\label{fig:SRimage}
\end{figure}

Single-image super-resolution (SR) algorithms aim to construct a high-quality
high-resolution (HR) image from a single low-resolution (LR) input.
Numerous single-image SR
algorithms have been recently proposed for generic images
that exploit priors based on
edges~\cite{DBLP:conf/cvpr/SunXS08}, gradients~\cite{DBLP:journals/tog/ShanLJT08, Kim08_PAMI}, neighboring interpolation \cite{DBLP:journals/cvgip/IraniP91,DBLP:conf/accv/TimofteSG14}, 
regression~\cite{DBLP:conf/eccv/DongLHT14}, 
and patches \cite{Glasner2009,DBLP:journals/tip/YangWHM10,DBLP:journals/tip/DongZSW11,DBLP:journals/tip/SunSXS11,DBLP:conf/cvpr/YangLC13,Yang13_ICCV_Fast,DBLP:journals/imst/FarsiuREM04,DBLP:conf/iccv/TimofteDG13,DBLP:conf/cvpr/SchulterLB15}.
Most SR methods focus on generating sharper edges with richer
textures, and are usually evaluated by measuring the similarity
between super-resolved HR and ground-truth images through full-reference
metrics such as the mean squared error (MSE), peak signal-to-noise ratio
(PSNR) and structural similarity (SSIM)
index~\cite{DBLP:journals/tip/WangBSS04}.
In our recent SR benchmark study~\cite{Yang14_ECCV}, we show that the information
fidelity criterion (IFC)~\cite{DBLP:journals/tip/SheikhBV05} performs 
favorably among full-reference metrics for SR performance evaluation. 
However, full-reference metrics are originally designed to account for image signal and noise
rather than human visual perception~\cite{DBLP:book/Girod}, even for several recently proposed methods .
We present 9 example SR images generated from a same LR image in Figure~\ref{fig:SRimage}. Table \ref{tb:srscore} shows that those full-reference metrics fail to match visual perception of human subjects well for SR performance evaluation.
In addition, full-reference metrics require ground-truth images for evaluation
which are often unavailable in practice. 
The question how we can effectively evaluate the quality of SR images based on visual perception still remains open.
In this work, we propose to learn a no-reference metric 
for evaluating the performance of single-image SR algorithms.
It is because no-reference metrics are designed to mimic visual perception (i.e., learned from large-scale perceptual scores) without requiring ground-truth images as reference. 
With the increase of training data, no-reference metrics have greater potential to match visual perception for SR performance evaluation.  

\begin{table}
	\centering
	\includegraphics[width=\textwidth]{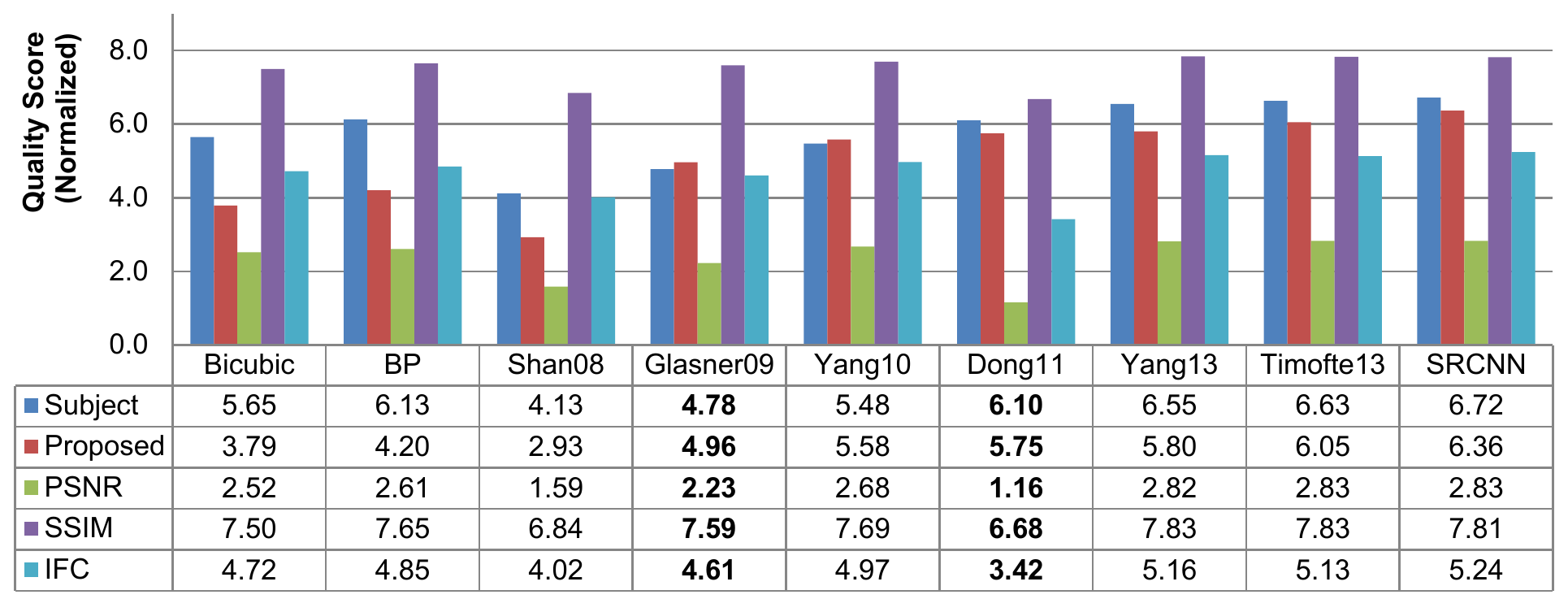} \\
	\caption{Quality scores of SR images in Figure \ref{fig:SRimage} from human subjects, 
		the proposed metric, rescaled PSNR, SSIM and IFC (0 for worst and 10 for
		best). 
		Note that human subjects favor Dong11 over Glasner09 as
		the SR image in Figure \ref{fig:SRimage}(d) is over-sharpened (best viewed on a
		high-resolution displayer).  
		However, the PSNR, SSIM and IFC metrics show opposite results 
		as the image in Figure \ref{fig:SRimage}(f) is misaligned to the reference
		image by 0.5 pixel. 
		In contrast, the proposed metric matches visual perception well.}
	\label{tb:srscore} 
\end{table}

We first conduct human subject studies using a large set of SR
images to collect perceptual scores. 
With these scores for training, 
we propose a novel no-reference quality assessment algorithm that
matches visual perception well.
Our work, in essence, uses the same methodology as that of general image
quality assessment (IQA) approaches. 
However, we evaluate the effectiveness of the signal reconstruction by SR algorithms 
rather than analyzing noise and distortions (e.g., compression and fading) as in
existing IQA methods~\cite{DBLP:journals/spl/MoorthyB10,DBLP:journals/tip/MoorthyB11,DBLP:conf/cvpr/TangJK11, DBLP:journals/tip/SaadBC12,DBLP:conf/cvpr/YeKKD12,DBLP:conf/cvpr/TangJK14}.
We quantify SR artifacts based on their statistical properties in both spatial and frequency domains, and regress them to collected perceptual scores. Experimental results demonstrate the effectiveness of the proposed no-reference
metric in assessing the quality of SR images against existing IQA measures. 

The main contributions of this work are summarized as follows.
First, we propose a novel no-reference IQA metric, which matches visual perception well, to evaluate the performance of SR algorithms. Second, we develop a large-scale dataset of SR images and conduct human subject studies on these images. 
We make the SR dataset with collected perceptual scores publicly available at \url{https://sites.google.com/site/chaoma99/sr-metric}.

\section{Related Work and Problem Context}

The problem how to evaluate the SR performance can be posed as assessing the quality of super-resolved images. 
Numerous metrics for general image quality assessment have been used to evaluate SR performance in the literature. 
According to whether the ground-truth HR images are referred, existing metrics fall into the following three classes.  

\subsection{Full-Reference Metrics}
Full reference IQA methods such as the MSE, PSNR, and SSIM indices~\cite{DBLP:journals/tip/WangBSS04}
are widely used in the SR literature~\cite{DBLP:journals/tog/ShanLJT08,Kim08_PAMI,DBLP:journals/tip/YangWHM10,DBLP:journals/tip/DongZSW11,DBLP:journals/tip/SunSXS11,DBLP:conf/cvpr/YangLC13,Yang13_ICCV_Fast}.
However, these measures are developed for analyzing generic image
signals and do not match human perception (e.g., MSE)~\cite{DBLP:book/Girod}. 
In~\cite{DBLP:conf/icip/ReibmanBG06}, Reibman et al. conduct subject
studies to examine the limitations of SR performance in terms of 
scaling factors using a set of three images and existing metrics.
Subjects are given two SR images each time and asked to select the
preferred one.
The perceptual scores of the whole test SR images are 
analyzed with the Bradley-Terry model~\cite{DBLP:conf/pics/Handley01}.
The results show that while SSIM performs better than others, it
is still not correlated with visual perception well.
In our recent SR benchmark work~\cite{Yang14_ECCV}, we conduct subject studies in a subset of generated SR images, and show that the IFC~\cite{DBLP:journals/tip/SheikhBV05} metric performs well among full-reference measures. 
Since subject studies are always time-consuming and expensive, Reibman et al. use only  six ground-truth images to generate test SR images while we use only 10  in~\cite{Yang14_ECCV}.
It is therefore of great importance to conduct larger subject study to address the question how to effectively evaluate the performance of SR
algorithms based on visual perception.

\subsection{Semi-Reference Metric}
In addition to the issues on matching visual perception, 
full-reference metrics can only be used for assessment 
when the ground-truth images are available.
Some efforts have been made to address this problem by
using the LR input images as references rather than the HR ground-truth
ones, which do not always exist in real-world applications. 
Yeganeh et al.~\cite{DBLP:conf/icip/YeganehRW12} extract two-dimension
statistical features in the spatial and frequency domains to 
compute assessment scores from either a test LR image or a generated
SR image. 
However, only 8 images and 4 SR algorithms are analyzed in their work. 
Our experiments with a larger number of test images and SR algorithms
show that this method is less effective due to the lack of 
holistic statistical features.

\subsection{No-Reference Metrics}
When the ground-truth images are not available, SR images can be
evaluated by the no-reference IQA methods~\cite{DBLP:journals/spl/MoorthyB10,DBLP:conf/cvpr/TangJK11,DBLP:journals/tip/MoorthyB11,DBLP:journals/tip/SaadBC12}
based on the hypothesis that natural images possess certain statistical properties,
which are altered in the presence of distortions (e.g., noise) and
this alternation can be quantified for quality assessment.
In~\cite{DBLP:conf/cvpr/YeKKD12, DBLP:conf/cvpr/TangJK14}, 
features learned from auxiliary datasets are used to quantify the natural 
image degradations as alternatives of statistical properties.
Existing no-reference IQA methods are all learning-based, but the training images are
degraded by noise, compression or fast fading rather than super-resolution.
As a result, the state-of-the-art no-reference IQA methods are less
effective for accounting for the artifacts such as incorrect high-frequency details
introduced by SR algorithms.
On the other hand, since SR images usually contain blur and
ringing artifacts, the proposed algorithm bears some resemblance
to existing metrics for blur and sharpness
estimation~\cite{DBLP:journals/tip/FerzliK09,DBLP:conf/cvpr/ChoJZKSF10,DBLP:journals/tog/LiuWCFR13}.
While the most significant difference lies in
that we focus on SR images, where numerous artifacts are 
introduced by more than one blur kernel.
In this work, we propose a novel
no-reference metric for SR image quality assessment by learning 
from perceptual scores based on subject studies involving
a large number of SR images and algorithms.

\begin{figure}
	\centering
	\includegraphics[width=.28\textwidth]{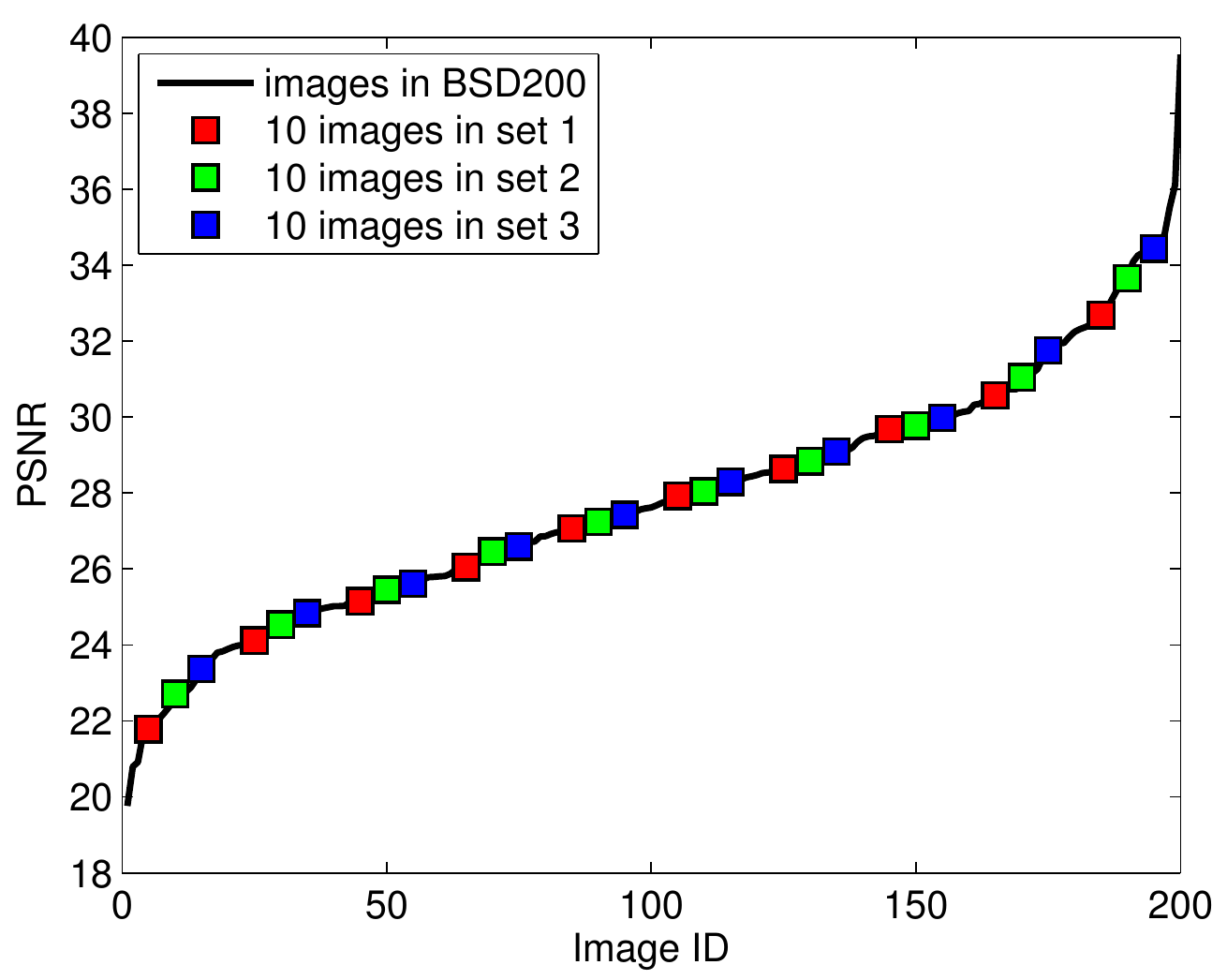}
	\includegraphics[width=.7\textwidth]{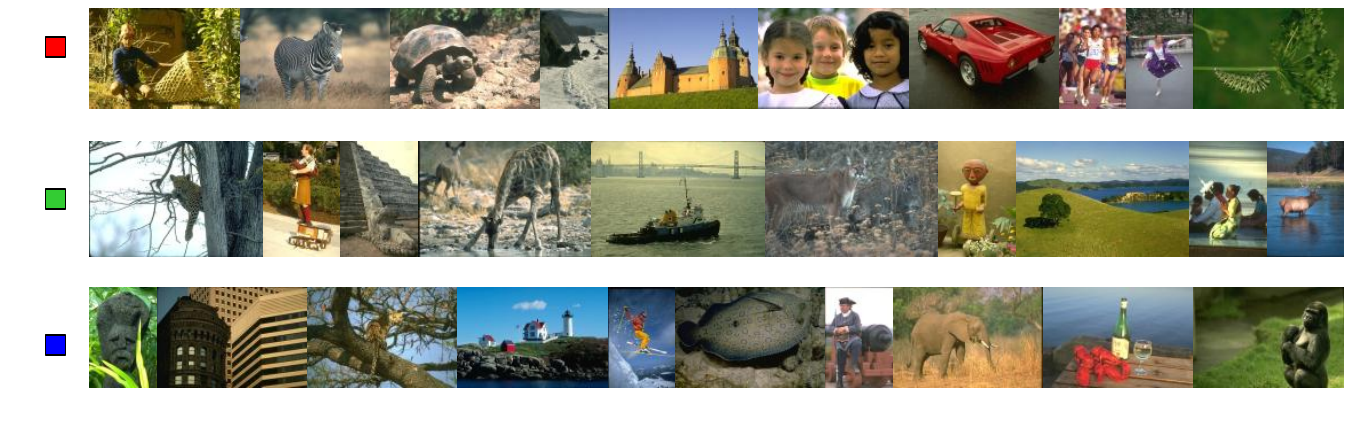}\\
	\caption{Ranked PSNR values on the BSD200 dataset and 
		the evenly selected three sets of images. 
		The PSNR values indicate the quality scores of the SR images 
		generated from the LR images using \eqref{eq:downsample}  
		with scaling factor ($s$) of 2 and Gaussian kernel width ($\sigma$) of 0.8 
		by the bicubic interpolation algorithm.}
	\label{fig:psnrsort}
\end{figure}

\section{Human Subject Studies} 

\label{sec:sujectstudy}
We use the Berkeley segmentation
dataset~\cite{ICCV01:BerklyDataSet} to carry out the experiments as
the images are diverse and widely used for SR
evaluation~\cite{Glasner2009,DBLP:journals/tip/SunSXS11,Yang13_ICCV_Fast}.
For an HR source image $I_h$, let $s$ be a scaling factor, and the width and height of $I_h$ be $s
\rm{\times} n$ and $s \rm{\times} m$.
We generate a downsampled LR image $I_l$ as follows:

\begin{equation}
\label{eq:downsample}
I_l(u, v)=\sum_{x, y}k(x-su, y-sv)I_h(x, y),
\end{equation}
where $u \in \{1,\dots,n\}$ and $v \in \{1,\dots,m\}$ are
indices of $I_l$, and $k$ is a matrix of Gaussian kernel weight determined by
a parameter $\sigma$, e.g.,
$k(\Delta x, \Delta y) = \frac{1}{Z}e^{-(\Delta x^2 + \Delta
	y^2)/2\sigma^2}$,
where $Z$ is a normalization term.
Compared to our benchmark work~\cite{Yang14_ECCV}, we remove the noise term from \eqref{eq:downsample} to reduce uncertainty.
The quality of the super-resolved images from those LR images are used to evaluate the SR performance.
In this work, 
we select 30 ground truth images from the BSD200 dataset~\cite{ICCV01:BerklyDataSet} according to the PSNR values.
In order to obtain a representative image set that covers a wide range of high-frequency details, 
we compute the PSNR values as the quality scores of the SR images generated from the LR images using \eqref{eq:downsample}  
with a scaling factor ($s$) of 2 and a Gaussian kernel width ($\sigma$) of 0.8 by the bicubic interpolation algorithm.
The selected 30 images are evenly divided into three sets 
as shown in Figure~\ref{fig:psnrsort}.

\begin{table}
	\small
	\centering
	\caption{The scaling factors ($s$) in our experiments with their corresponding kernel width values ($\sigma$).}
	\label{fig:pair}
	\setlength{\tabcolsep}{1.4em}
	\begin{tabular}{ |c|c|c|c|c|c|c|}
		\hline
		$s$ & 2 &  3  &  4  & 5  & 6 & 8 \\ \hline 
		$\sigma$  &  0.8 & 1.0  &  1.2 &  1.6   &  1.8  &  2.0  \\ \hline
	\end{tabular}
\end{table}
\begin{figure}[!t]
	\centering
	\small
	\setlength{\tabcolsep}{.2em}
	\begin{tabular}{ccc}
		\includegraphics[width=.32\textwidth]{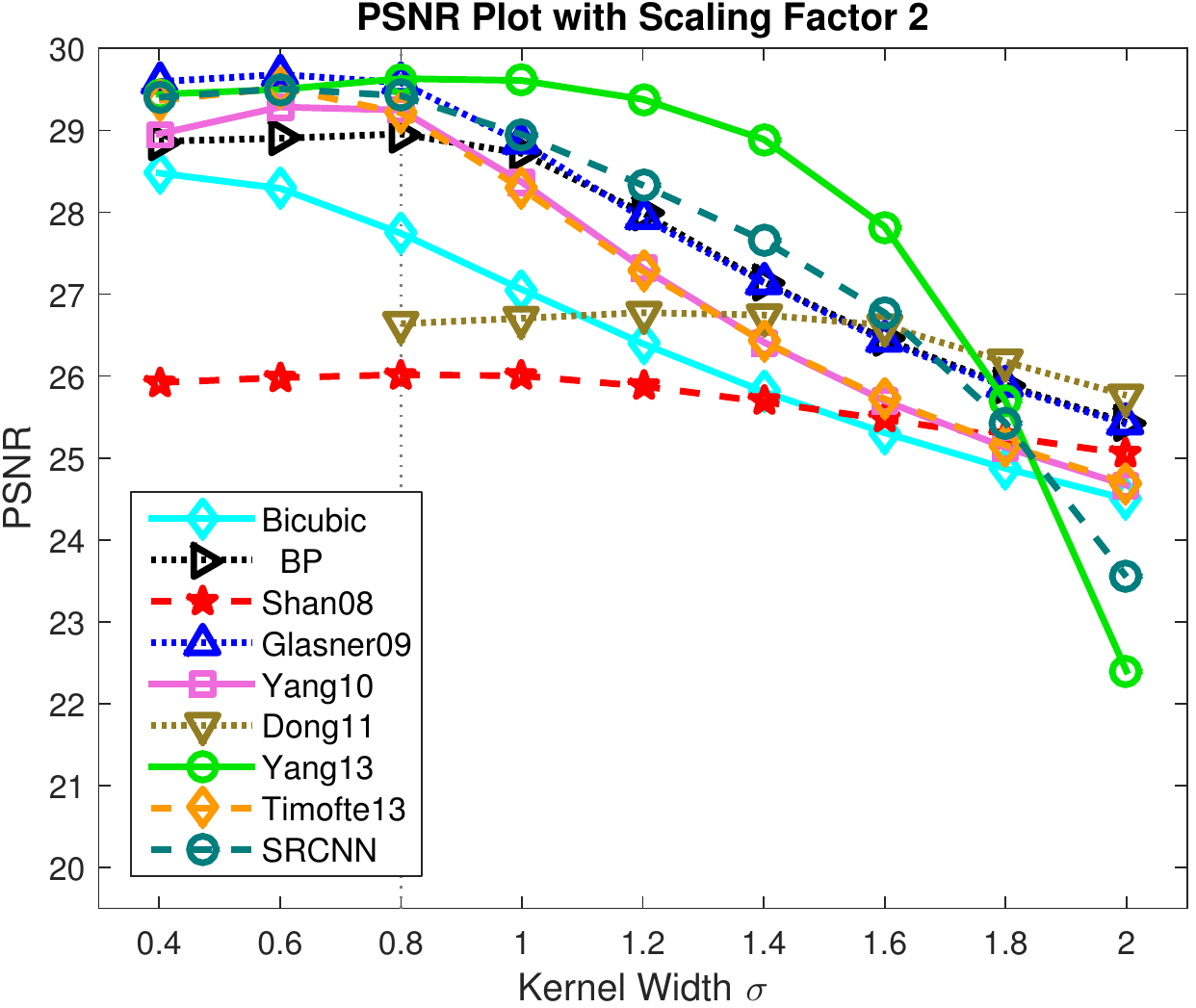}  &
		\includegraphics[width=.32\textwidth]{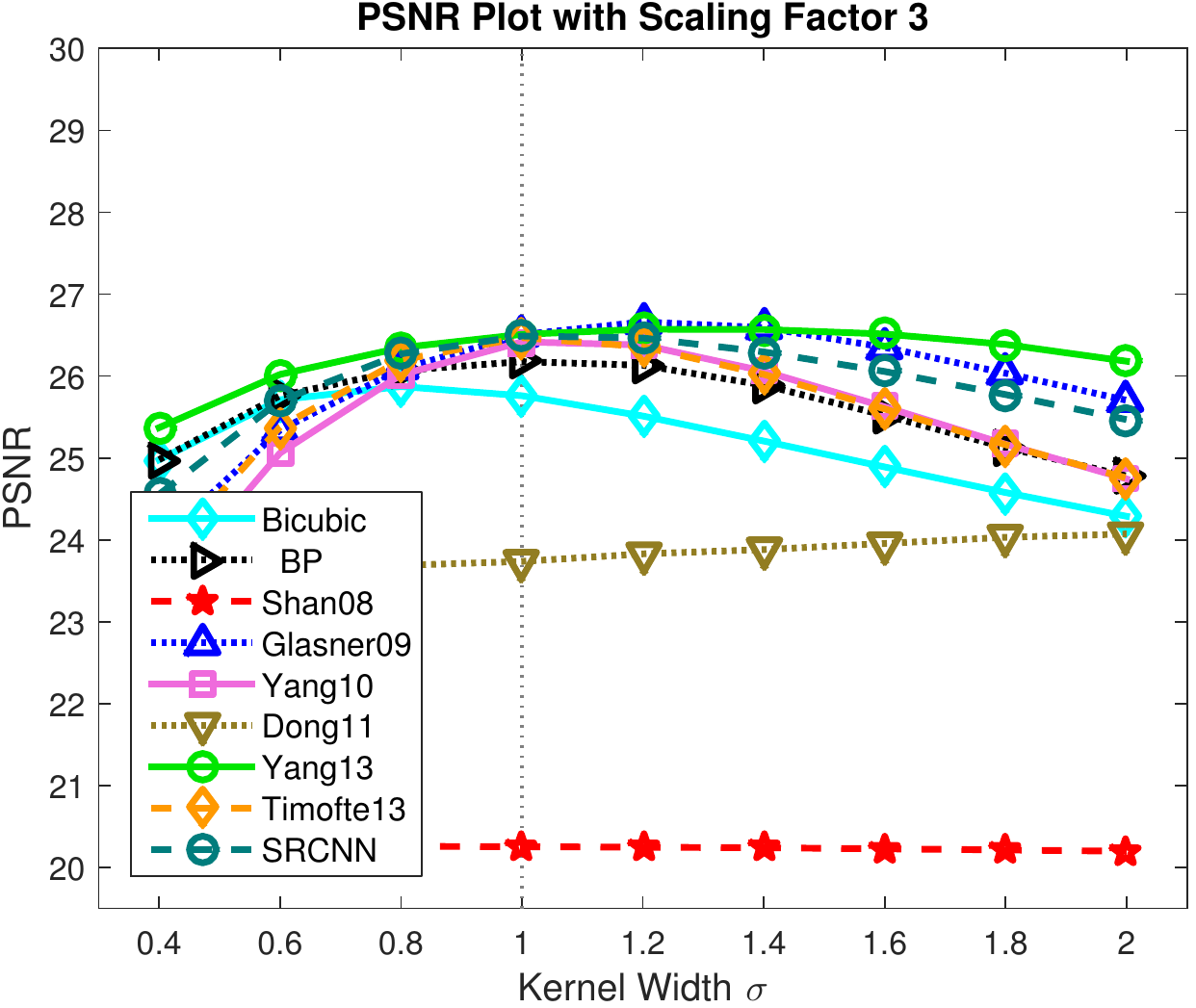}  &
		\includegraphics[width=.32\textwidth]{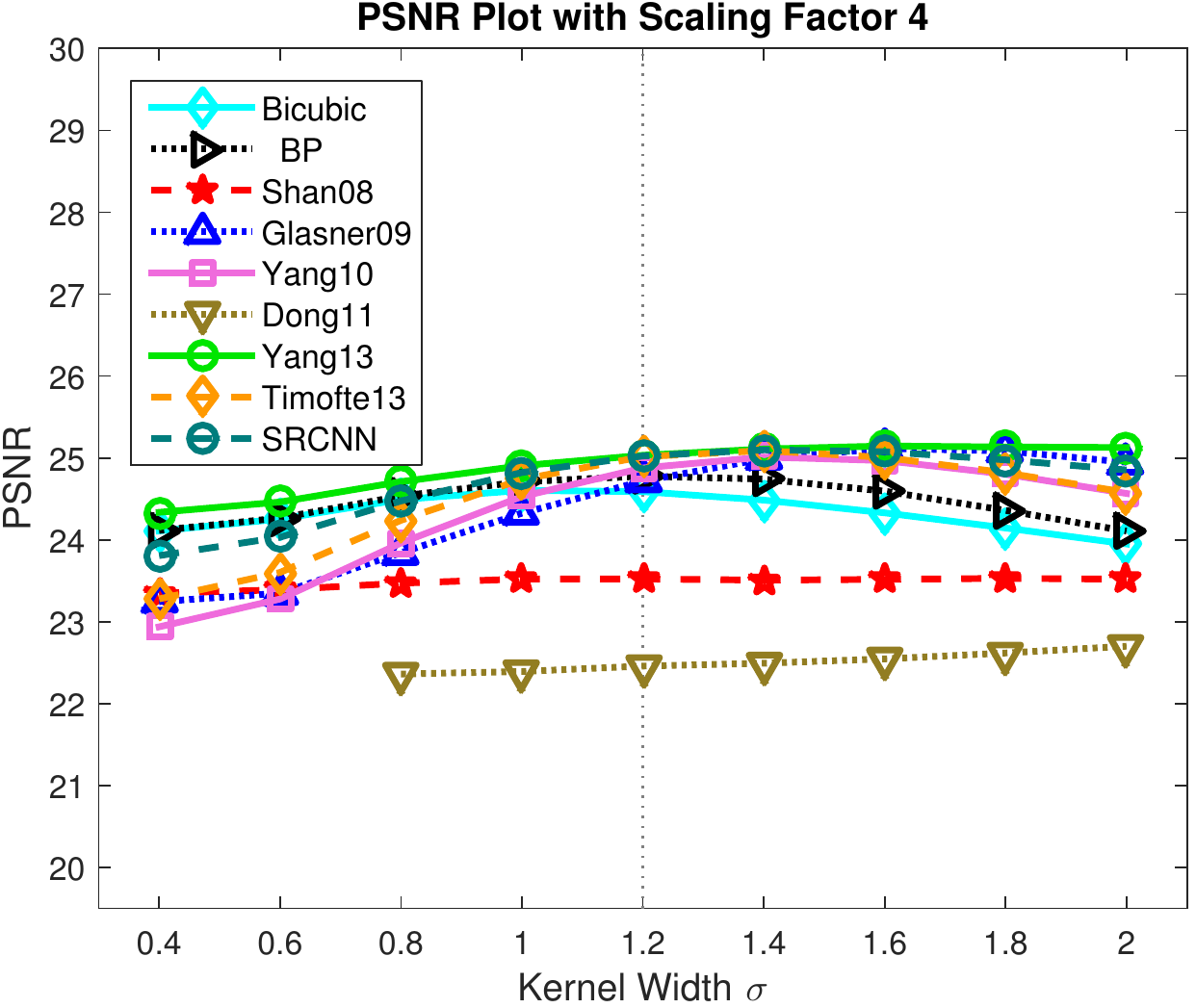} \\
		\includegraphics[width=.32\textwidth]{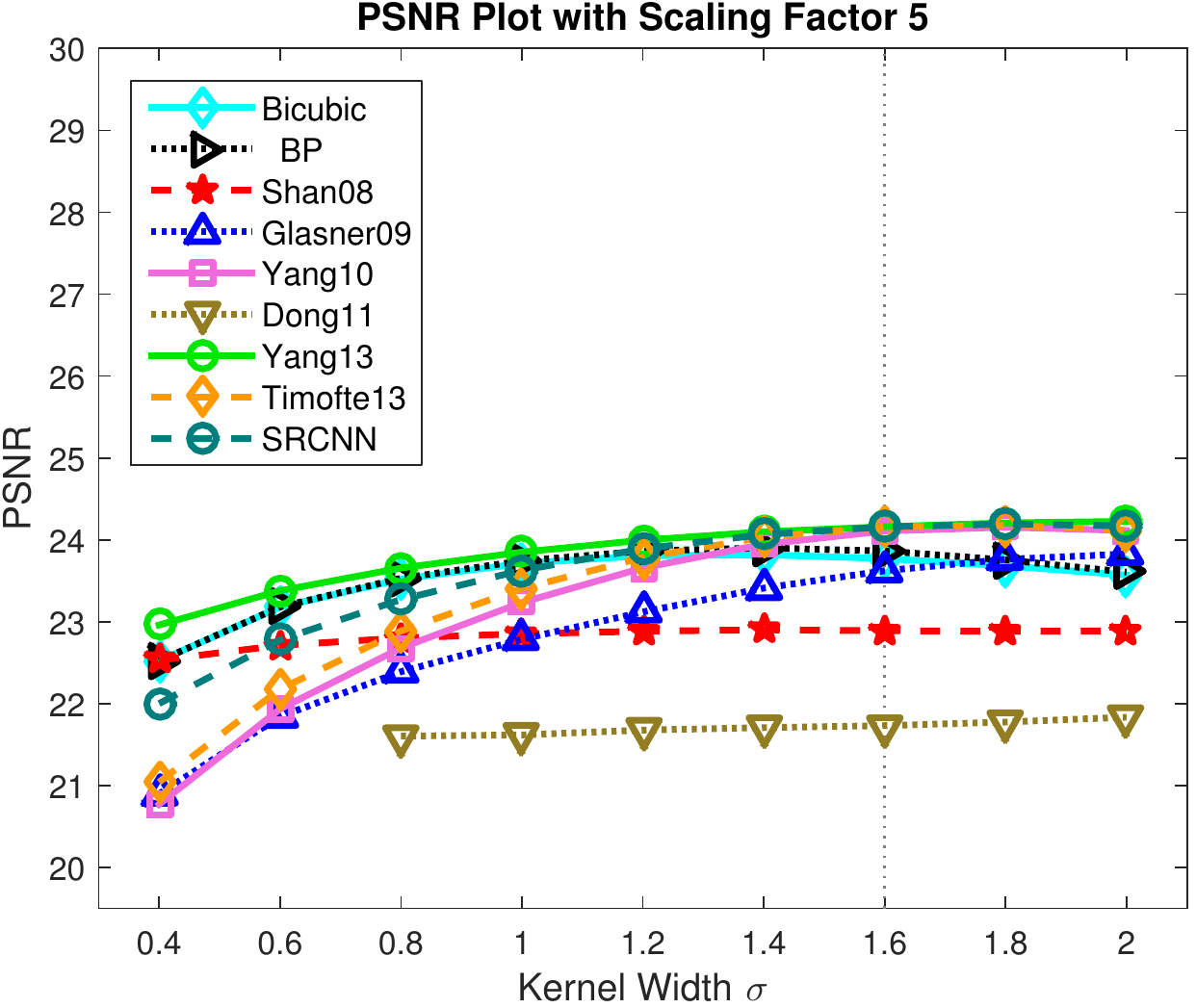} &
		\includegraphics[width=.32\textwidth]{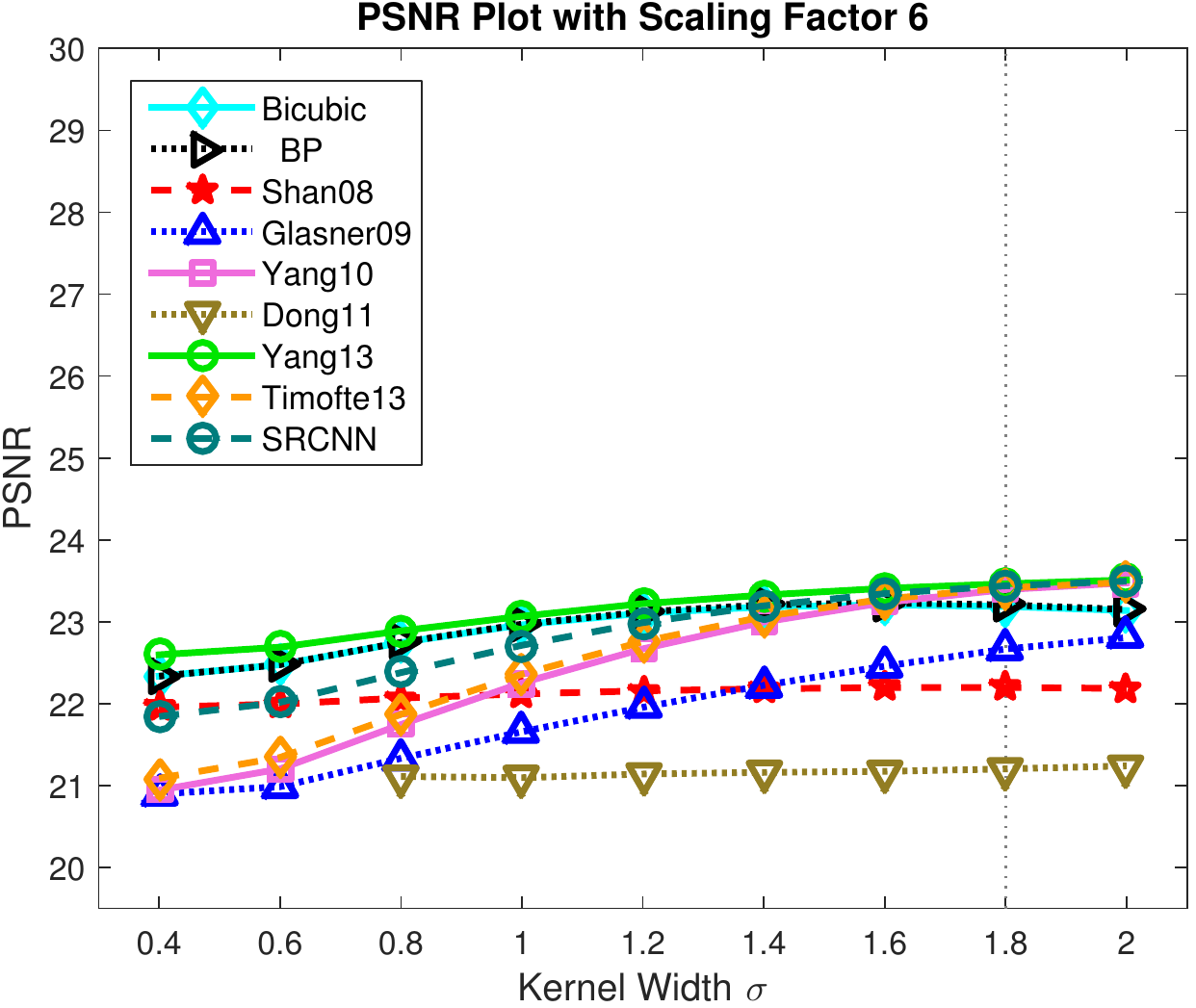} &
		\includegraphics[width=.32\textwidth]{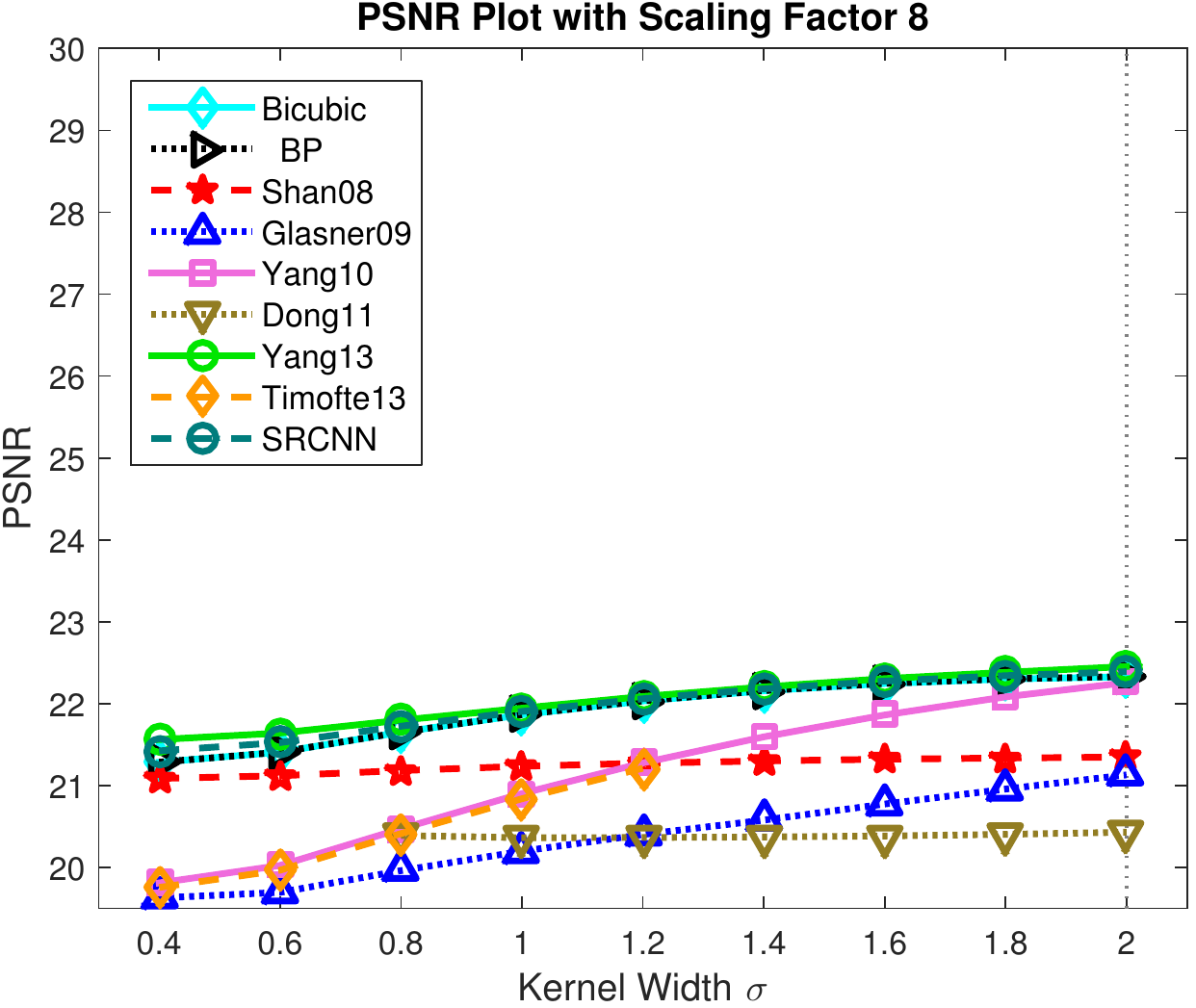}\\
	\end{tabular}
	\caption{Distribution of mean PSNR on the selected images.
		Note the increasing trend of the kernel width along the increase of the scale factor to generate the peak PSNR values.
		%
		The SR algorithm Dong11 does not converge when $\sigma<0.8$. The vertical dash line highlights the optimal kernel width for each scaling factor. 
	}
	\label{fig:kw}
\end{figure}


The LR image formation of (\ref{eq:downsample}) can be viewed as a
combination of a downsampling
and a blurring operation which is determined by the scaling factor $s$ and
kernel width $\sigma$, respectively.
As subject studies are time-consuming and expensive, our current work focuses on large differences caused by scaling factors, which are critical to the quality assessment of SR images.
We focus on how to effectively quantify the upper bound of SR performance based on human perception.
Similar to~\cite{Yang14_ECCV}, we assume the kernel width is known, and
compute the mean PSNR values of the SR images
generated by 9 SR methods under various settings ($s\in\{2,3,4,5,6,8\}$ and 
$\sigma \in \{0.4,0.6,\ldots,2\}$)
using 30 ground truth images.
Figure~\ref{fig:kw} shows that the larger subsampling factor requires larger
blur kernel width for better performance.
We thus select an optimal $\sigma$ for each scaling factor ($s$)
as shown in Table~\ref{fig:pair}.

\begin{table}[t!]
	\centering
	\small
	\caption{Empirical quality scores on SR images generated by bicubic interpolation. GT indicates the ground-truth HR images.}
	\label{tb:inst}
	\setlength{\tabcolsep}{.7em}
	\begin{tabular}{ |c|c|c| c| c |c| c| c| }
		\hline
		$s$  & \ GT \ & 2 &  3  &  4  & 5  & 6 & 8 \\ \hline
		Score ($\approx$)  &   10   &   $8\sim9$   &   $5\sim7$   &   $4\sim6$   &   $3\sim5$    &   $2\sim 4$   &   $<2$   \\ \hline
	\end{tabular}
\end{table}

\begin{figure}
	\centering
	\includegraphics[width=.84\textwidth]{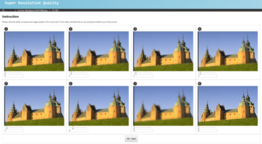}\\
	\caption{One screenshot of human subject study. Subjects assign scores
		between 0 to 10 to displayed SR images. Test images are randomly
		presented in order to reduce bias caused by similarity of image
		contents.}
	\label{fig:screenshot}
\end{figure}

In the subject studies, we use
absolute rating scores rather than 
pairwise comparison scores as we have 1,620 test images, 
which would require millions of pairwise comparisons (i.e., $C_2^{1620}\approx1.3\text{M}$). 
Although the sampling strategy \cite{CVPR14_PengYe} could alleviate this burden partially, pairwise comparison is infeasible given the number of subjects, images and time constraints. 
We note that subject studies in \cite{DBLP:journals/tip/SheikhSB06,Yang14_ECCV} are also based on absolute rating. 
In this work, we develop a user interface (See Figure~\ref{fig:screenshot})
to collect perceptual scores for these SR images. 
At each time, we simultaneously show 9 images generated from one LR image 
by different SR algorithms on a high-resolution display. 
These images are displayed in random order to reduce
bias caused by correlation of image contents. 
Subjects are asked to give scores from 0 to 10 to indicate 
image quality based on visual preference. 
We divide the whole test into 3 sections evenly such that subjects
can take a break after each section and keep high attention span in our studies.
To reduce the inconsistency among the individual quality criterion, 
we design a training process to conduct the test
at the beginning of each section, 
i.e., giving subjects an overview of all the
ground-truth and SR images generated by bicubic interpolation with the referred scale of quality scores 
as shown in Table~\ref{tb:inst}.

\begin{table}
	\caption{Data sets used for image quality assessment based on subject
		studies.} 
	\centering
	\small
		\setlength{\tabcolsep}{0.6em}
		\label{tb:dataset}
		\begin{tabular}{|l|c|c|c|}
			\hline
			Dataset & \# Reference Images  & \ \# Distortions \ & \# Subject Scores  \\ \hline\hline
			LIVE~\cite{DBLP:journals/tip/SheikhSB06}    & 29   & 982 & 22,457 \\ \hline
			ASQA~\cite{CVPR14_PengYe}  & 20   & 120 &  35,700 \\ \hline
			SRAB~\cite{Yang14_ECCV}  & 10   & 540 &  16,200 \\ \hline
			Our study    & 30   & 1,620 & 81,000 \\ \hline
		\end{tabular}
\end{table}

\begin{figure}
	\centering
	\small
	\setlength{\tabcolsep}{.5em}
	\begin{tabular}{cc}
		\includegraphics[width=.42\textwidth]{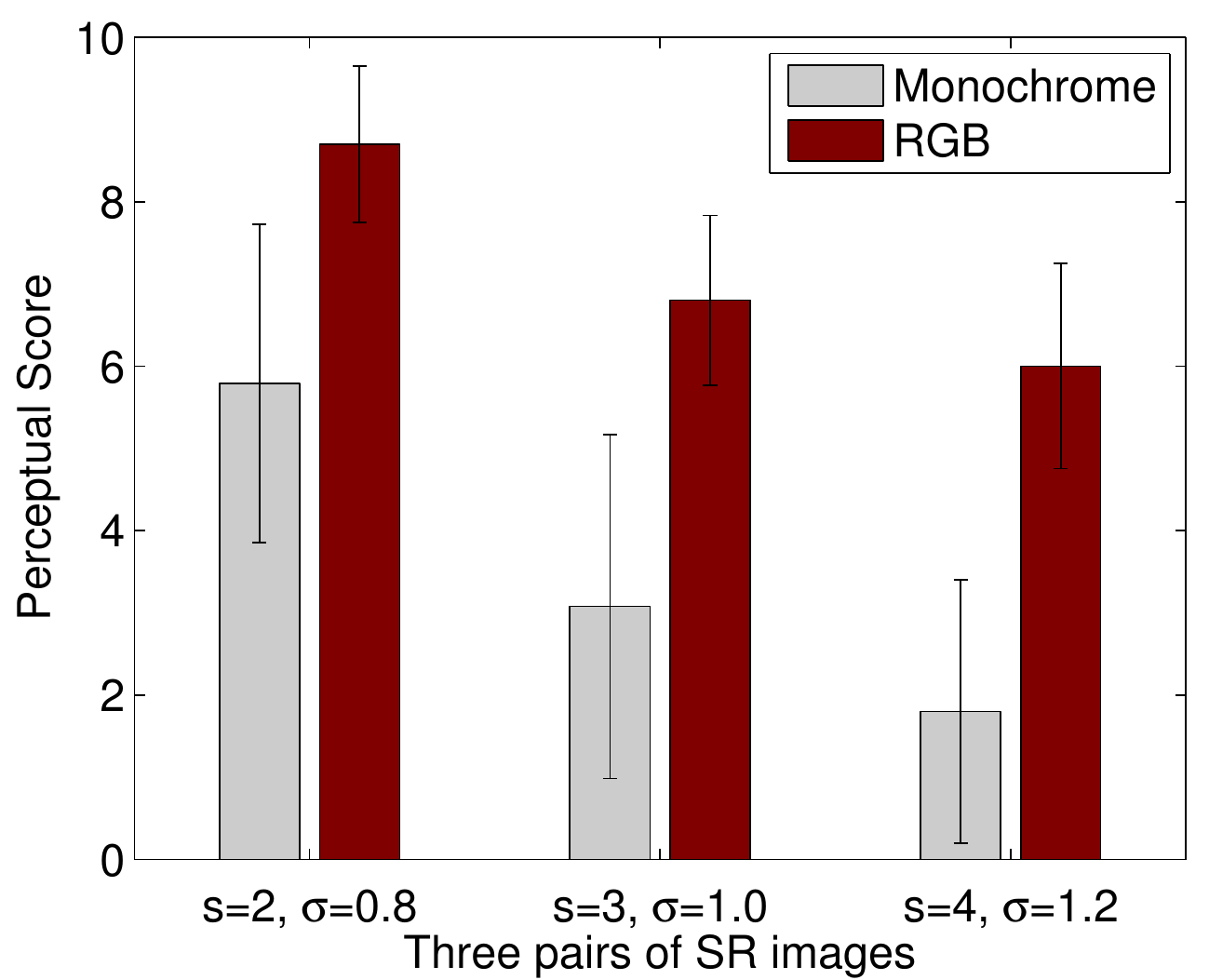} &
		\includegraphics[width=.42\textwidth]{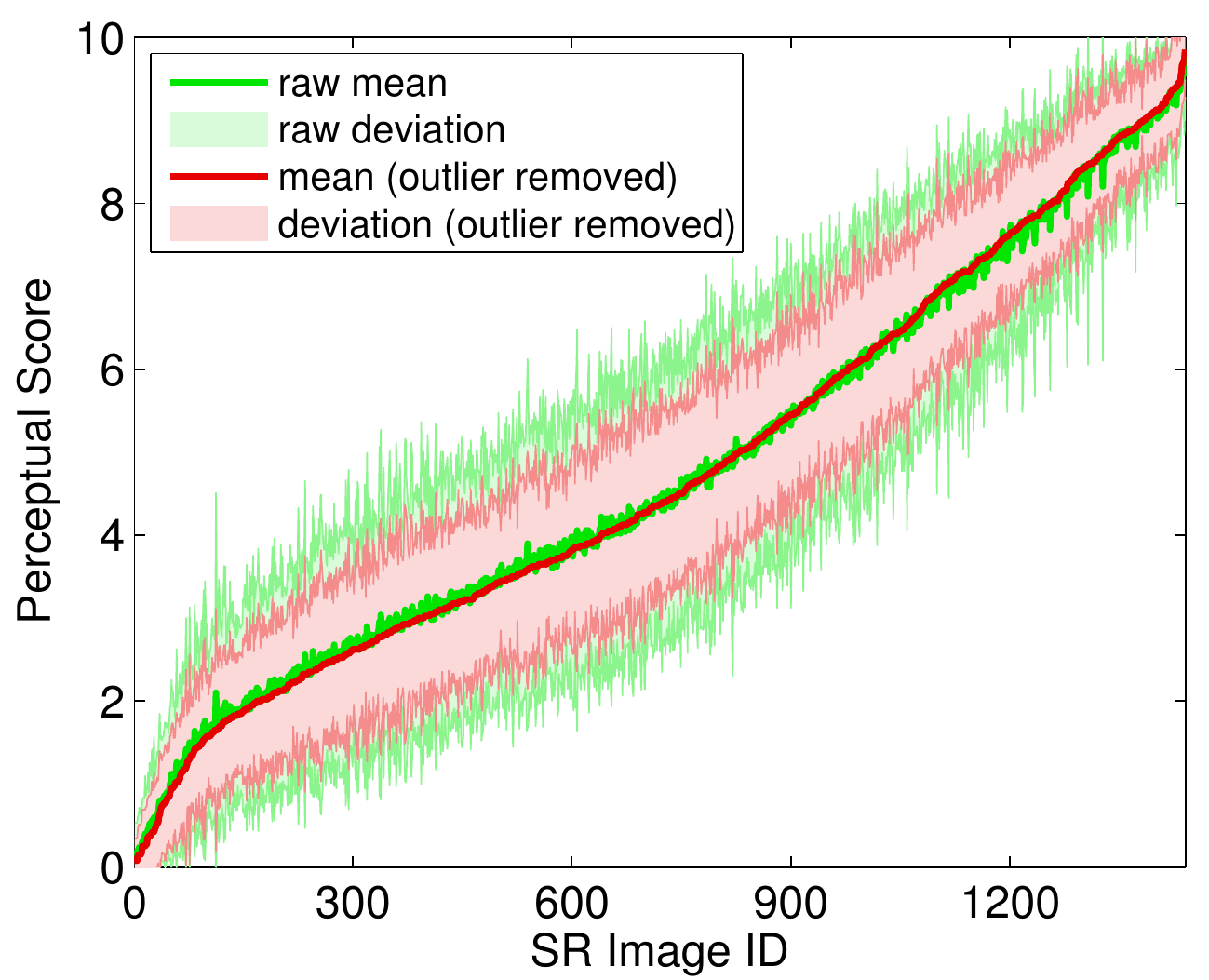} \\
		(a) & (b)
	\end{tabular}
	\caption{(a) Deviation of 50 perceptual scores
		on three pairs of SR images generated by bicubic interpolation from the same
		test image in Figure~\ref{fig:SRimage}.
		(b) Sorted mean perceptual scores and deviations before and after removing
		outliers. } 
	\label{fig:meanerror}
\end{figure}

We collect 50 scores from 50 subjects for each image, and 
compute the perceptual quality index as the mean of the median 40
scores to remove outliers.
To the best of our knowledge, our subject study is the largest so far in
terms of SR images, algorithms, and subject scores (See Table~\ref{tb:dataset}).
%
In addition to using more images than~\cite{Yang14_ECCV}, we
present subjects color SR images for evaluation as we observe that monochrome SR images introduce larger individual bias as demonstrated in Figure~\ref{fig:meanerror}(a).
It is reasonable that gray-scale images are rear in daily life and subjects hold different quality criterion. 
Figure~\ref{fig:meanerror}(b) shows that
the mean perceptual scores are more stable after removing outliers.

\begin{figure}[!t]
	\centering
	\small
	\begin{minipage}{\textwidth}
		\includegraphics[width=.495\textwidth]{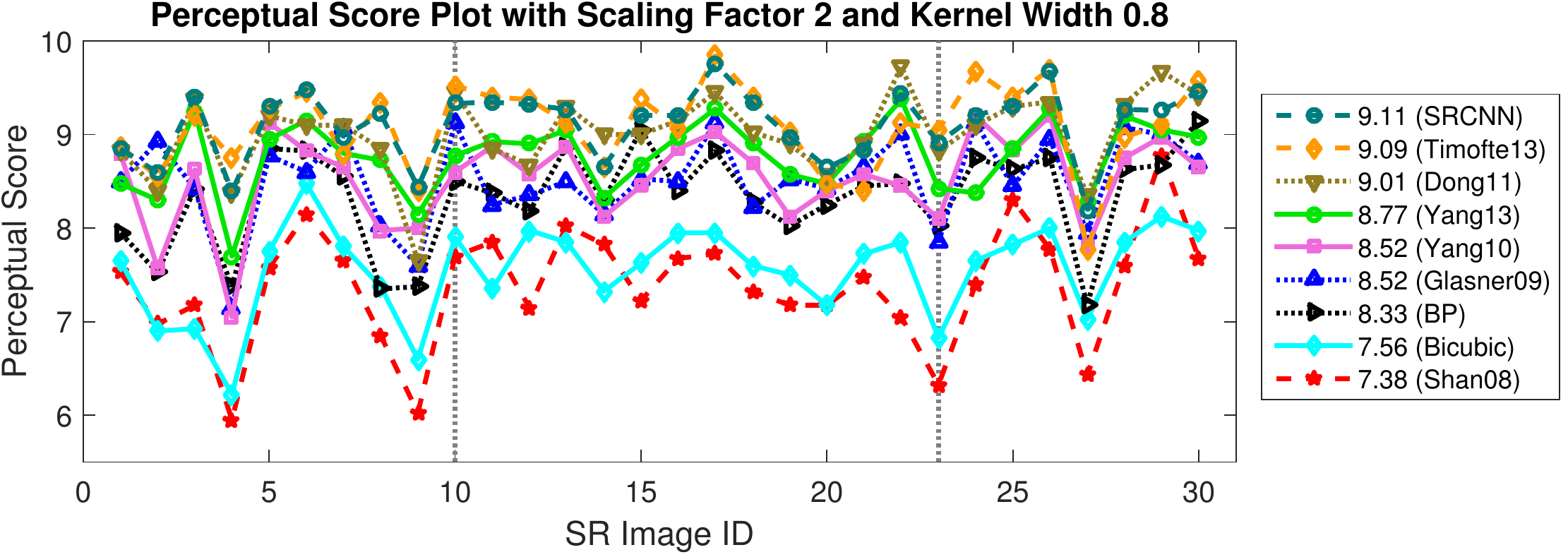}
		\includegraphics[width=.495\textwidth]{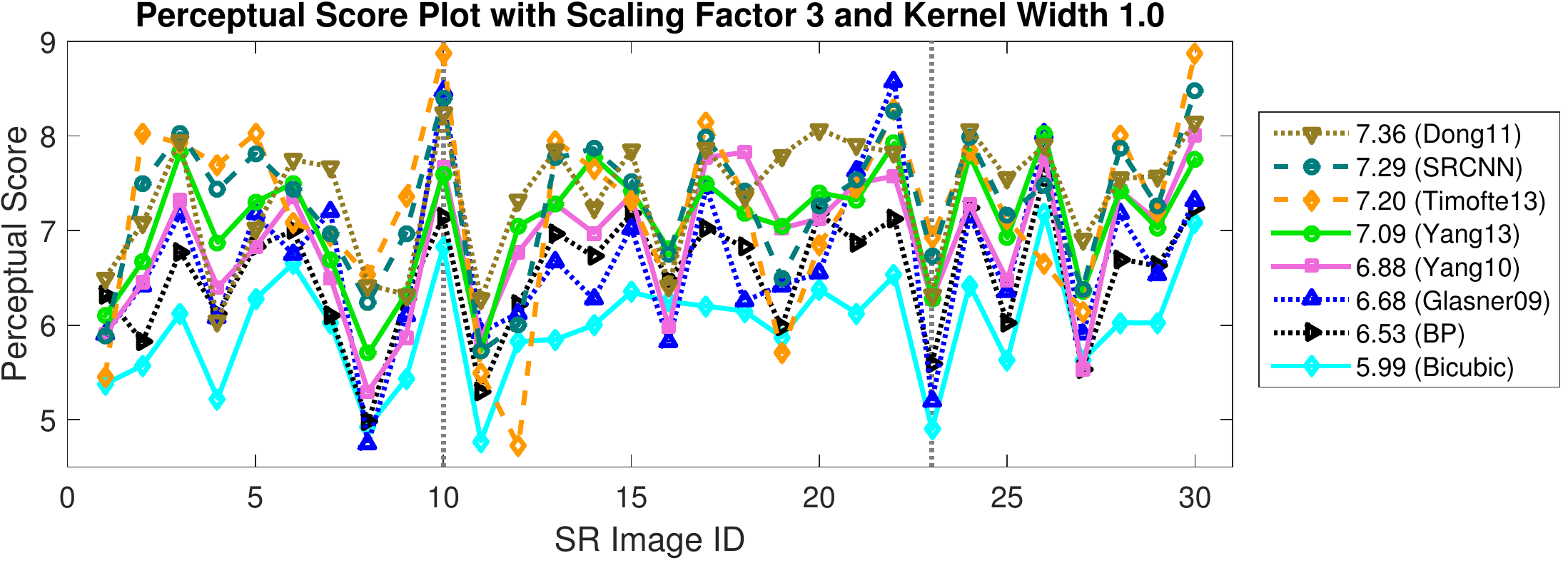}\\
		\includegraphics[width=.495\textwidth]{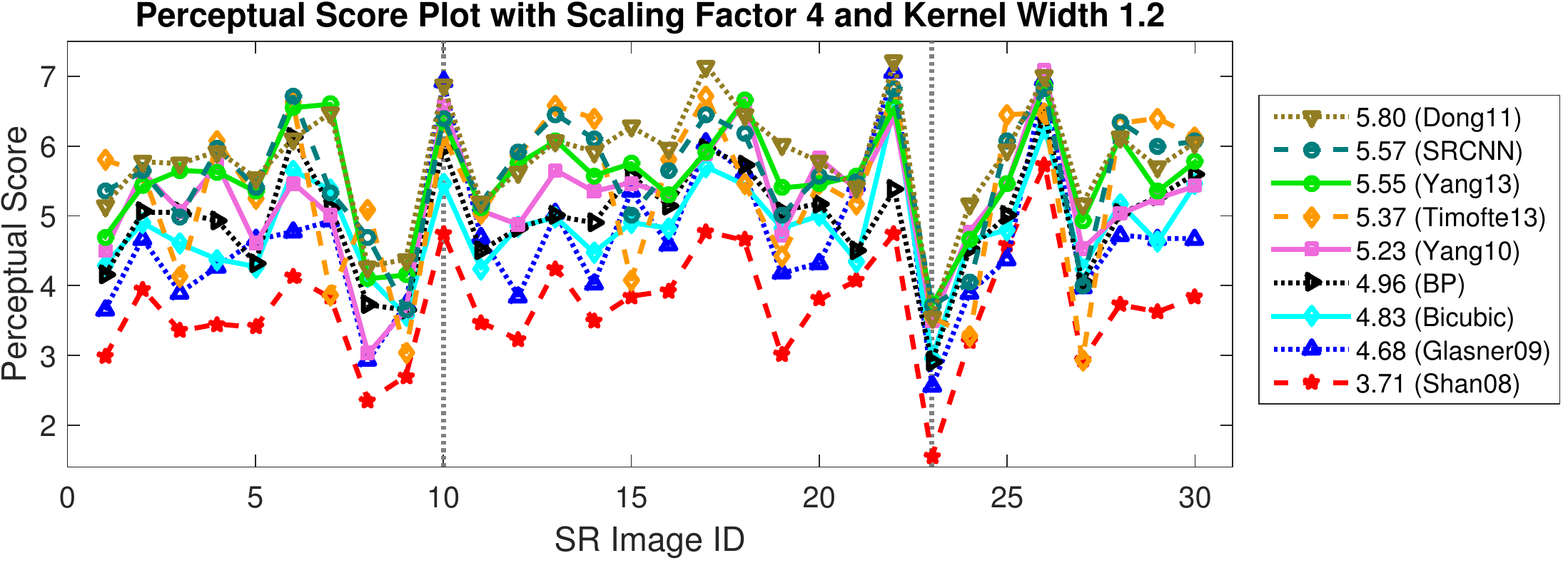}
		\includegraphics[width=.495\textwidth]{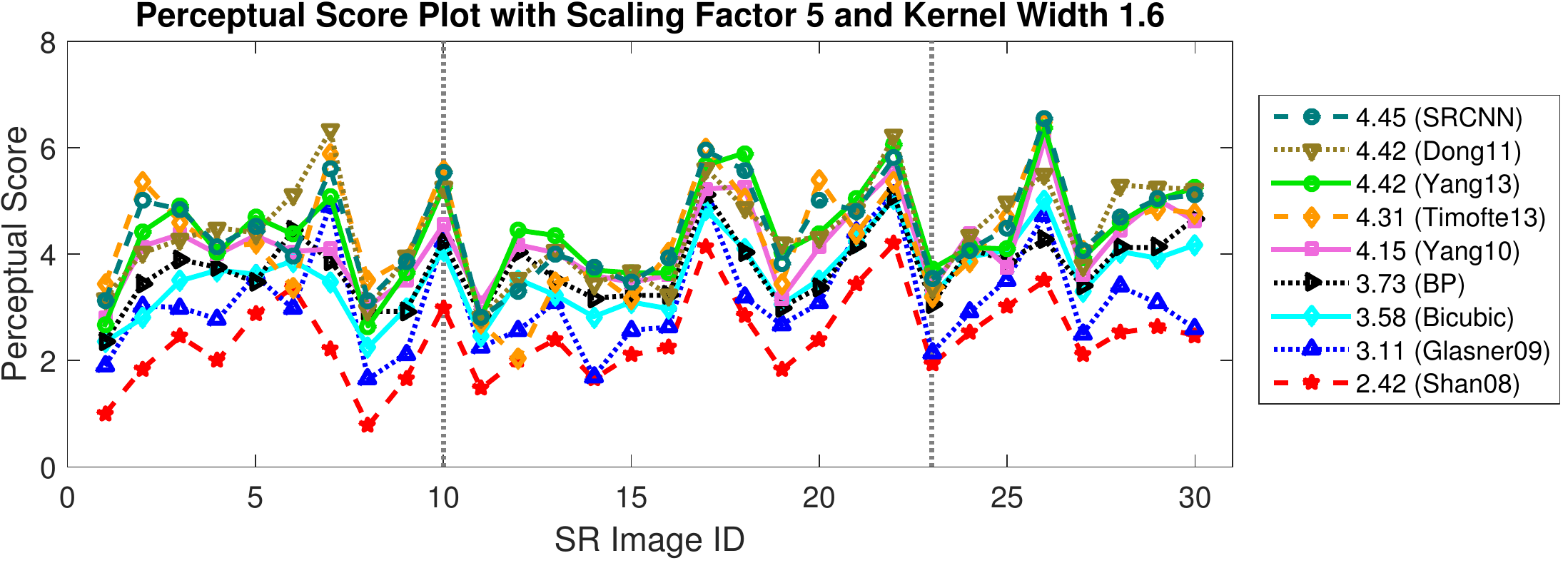}\\
		\includegraphics[width=.495\textwidth]{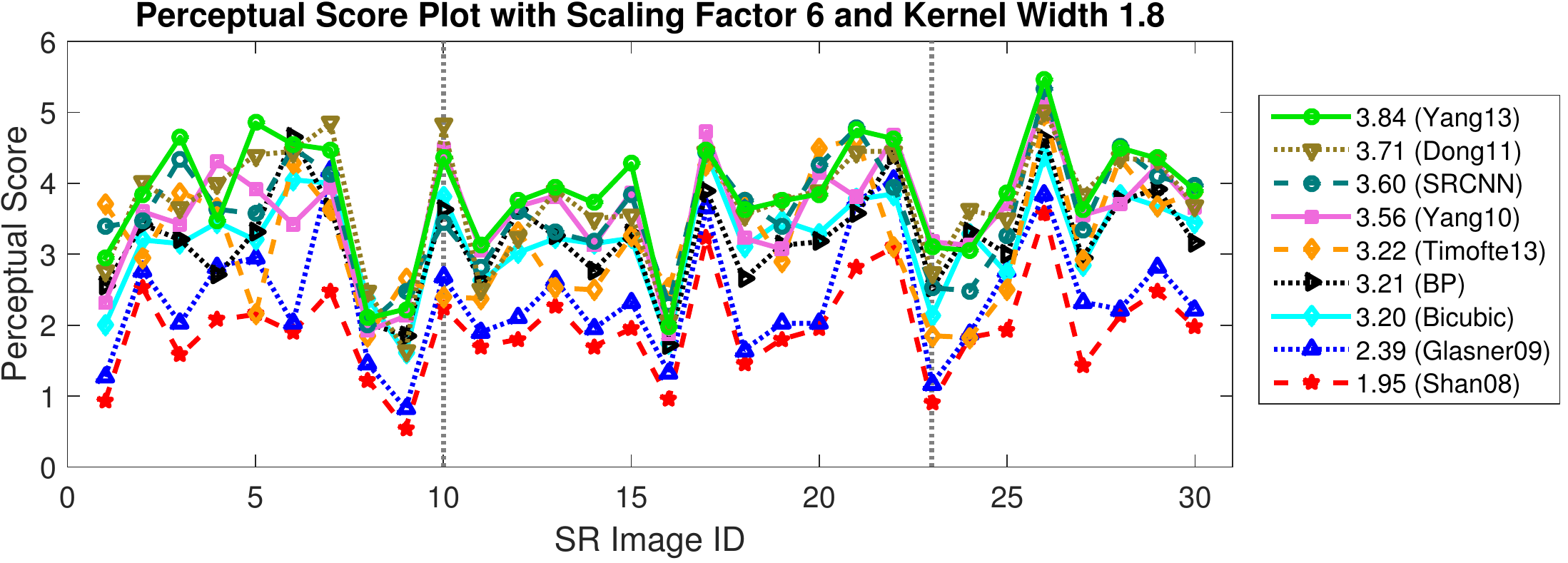}
		\includegraphics[width=.495\textwidth]{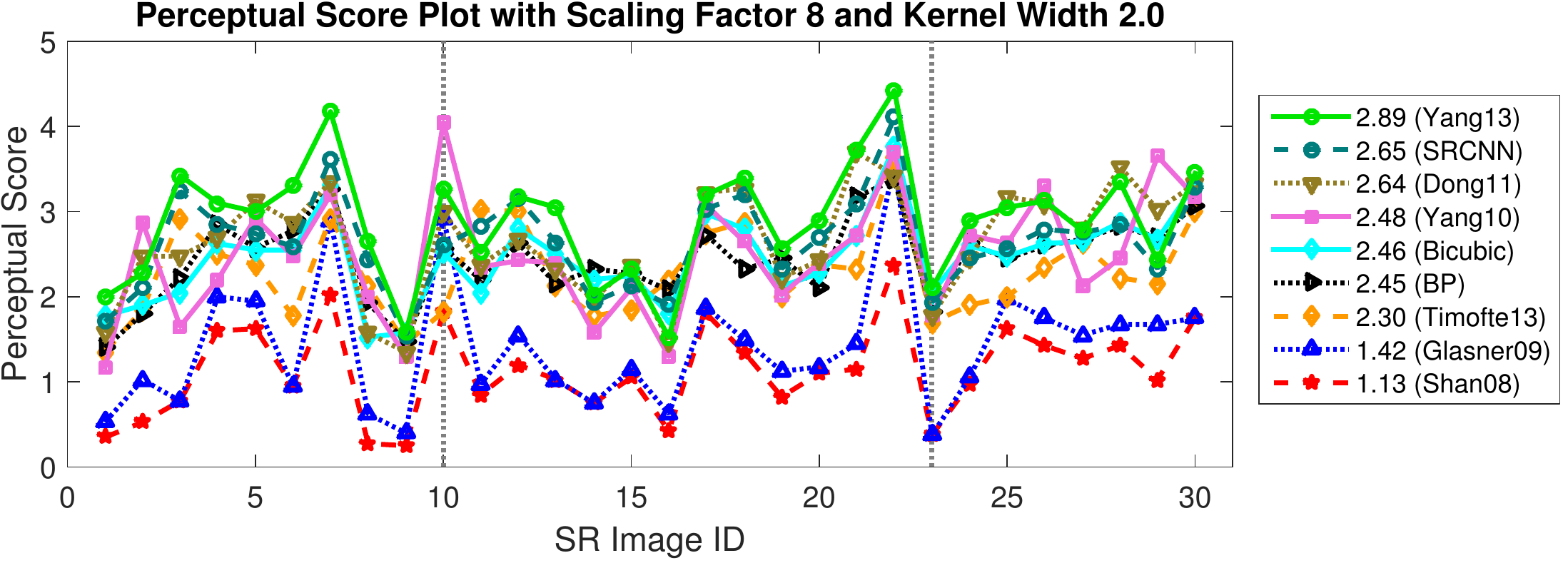}\\
		\caption{Perceptual scores of SR images under 6 pairs of scaling factor ($s$) and kernel width ($\sigma$). 
			The performance rank of SR algorithms remains relatively consistent,
			even while score values change under different scaling factors and kernel widths. 
			The average perceptual scores of each SR algorithm are shown in the legend
			(Shan08 with $s=3$, $\sigma=1.0$ 
			is excluded as the SR images contain severe noise and their perceptual scores are close to 0)}. 
		\vspace{1em}
		\label{fig:score}
	\end{minipage}
	\begin{minipage}{\textwidth}
		\centering
		\small
		\setlength{\tabcolsep}{1em}
		\begin{tabular}{cc}
			\includegraphics[width=.42\textwidth]{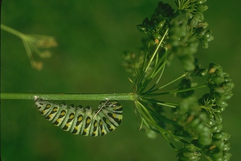}&
			\includegraphics[width=.42\textwidth]{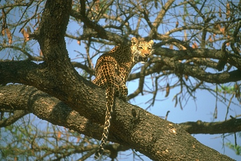}\\
			(a) Image ID 10  & (b) Image ID 23
		\end{tabular}
		\caption{The scores in Figure~\ref{fig:score} indicated by vertical dash lines for the SR images generated from (a) are much higher than that of (b).}
		\label{fig:twoimages}
	\end{minipage}
\end{figure}


Figure~\ref{fig:score} shows the computed mean perceptual quality indices
in terms of scaling factor and kernel width.
From the human subject studies, we have the following observations.
First, the performance rank of 9 SR algorithms remains the same (i.e.,
the curves are similar)
across all images in Figure \ref{fig:score}(a)-(f),
which shows consistency of perceptual scores on evaluating SR
algorithms. 
Second, the performance rank changes with scaling factors,
e.g., Glasner09 outperforms Bicubic with higher perceptual scores in
Figure~\ref{fig:score}(a) while it is the opposite in Figure~\ref{fig:score}(c). 
Since the image quality degradation caused by scaling factors is larger 
than that by different SR methods, 
the statistical properties for quantifying SR artifacts have to be discriminative 
to both scaling variations and SR algorithms. 
Third,  SR results generated from LR images with more smooth
contents have higher perceptual scores, e.g., the score of the image in
Figure~\ref{fig:twoimages}(a) is higher than that of Figure~\ref{fig:twoimages}(b).
This may be explained by the fact that visual perception is sensitive
to edges and textures and most algorithms do not perform well for
images such as Figure~\ref{fig:twoimages}(b).

\section{Proposed Algorithm}

\begin{figure}
	\centering
	\includegraphics[width=\textwidth]{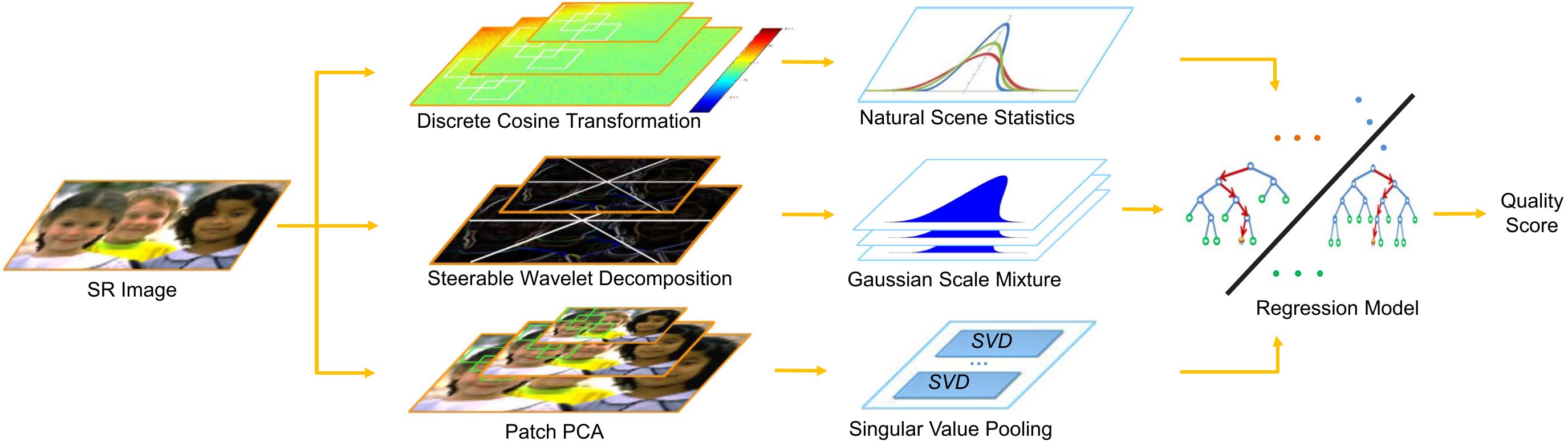}
	\caption{Main steps of the proposed no-reference metric.
		For each input SR image, statistics computed from the 
		spatial and frequency domains are used as features to represent SR
		images. Each set of extracted features are trained in separate ensemble regression trees,
		and a linear regression model is used to predict a
		quality score by learning from a large number of visual perceptual scores.}
	\label{fig:flowchart}
\end{figure}

We exploit three types of statistical properties as features, 
including local and global frequency variations and spatial discontinuity,
to quantify artifacts and assess the quality of SR images. 
Each set of statistical features is computed on a pyramid to alleviate the
scale sensitivity of SR artifacts.
Figure~\ref{fig:flowchart} shows the main steps of the proposed
algorithm for learning no-reference quality metric. 
Figure~\ref{fig:feat} shows an overview of the statistical properties 
of each type of features. 

\subsection{Local Frequency Features}
\begin{figure}
	\small
	\centering
	\setlength{\tabcolsep}{0.1em}
	\begin{tabular}{ccc}
		\includegraphics[width=.33\textwidth]{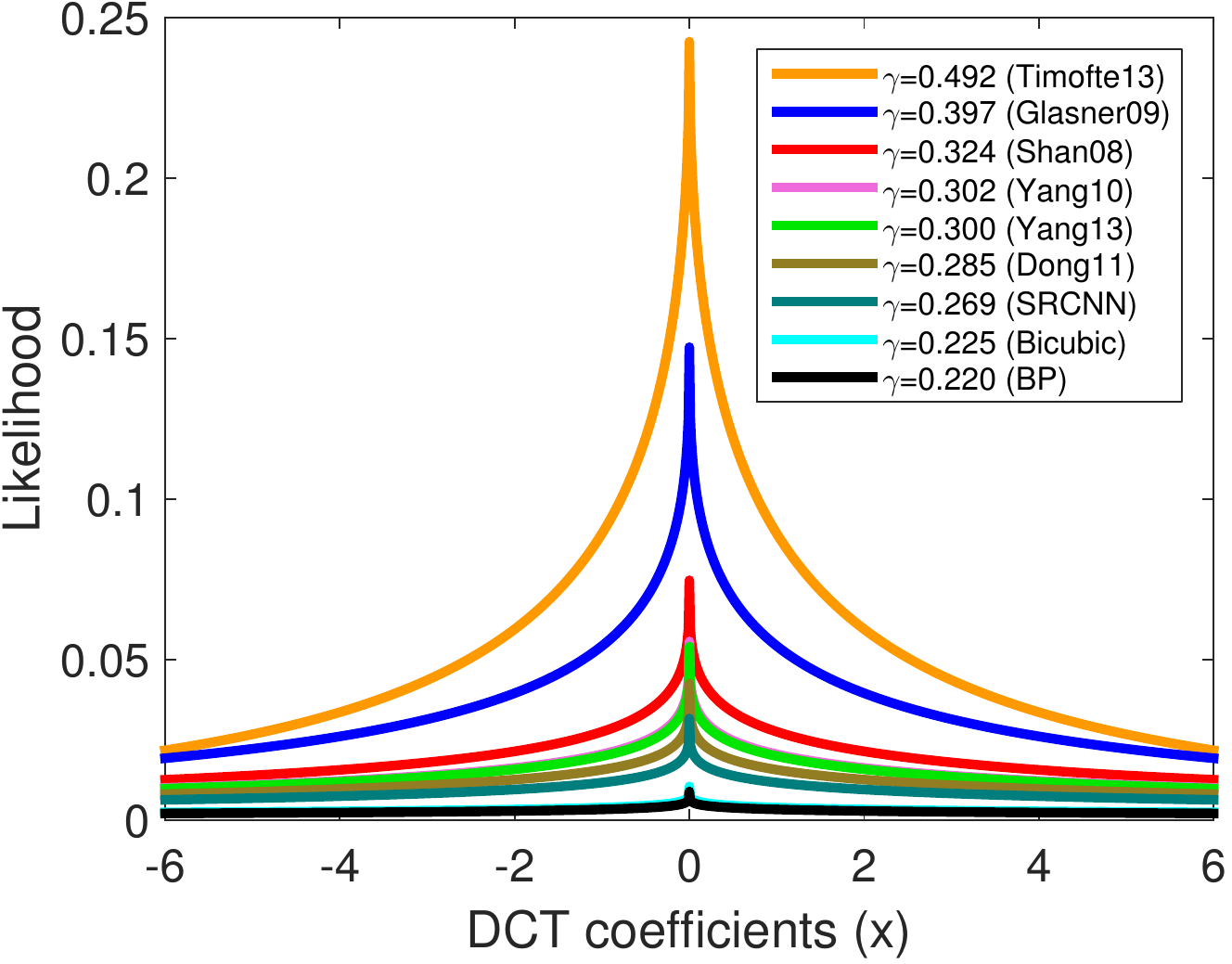} & 
		\includegraphics[width=.332\textwidth]{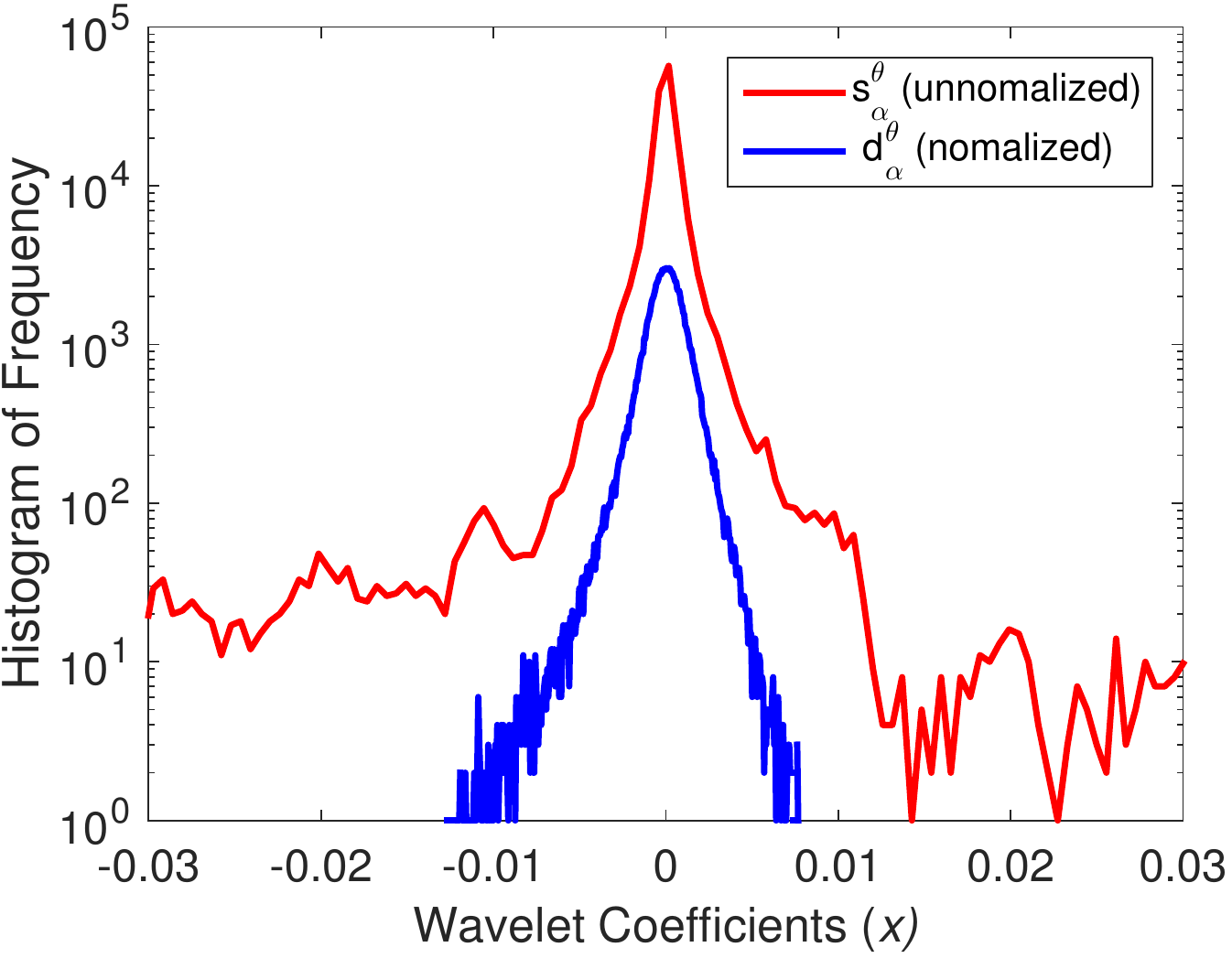} &
		\includegraphics[width=.318\textwidth]{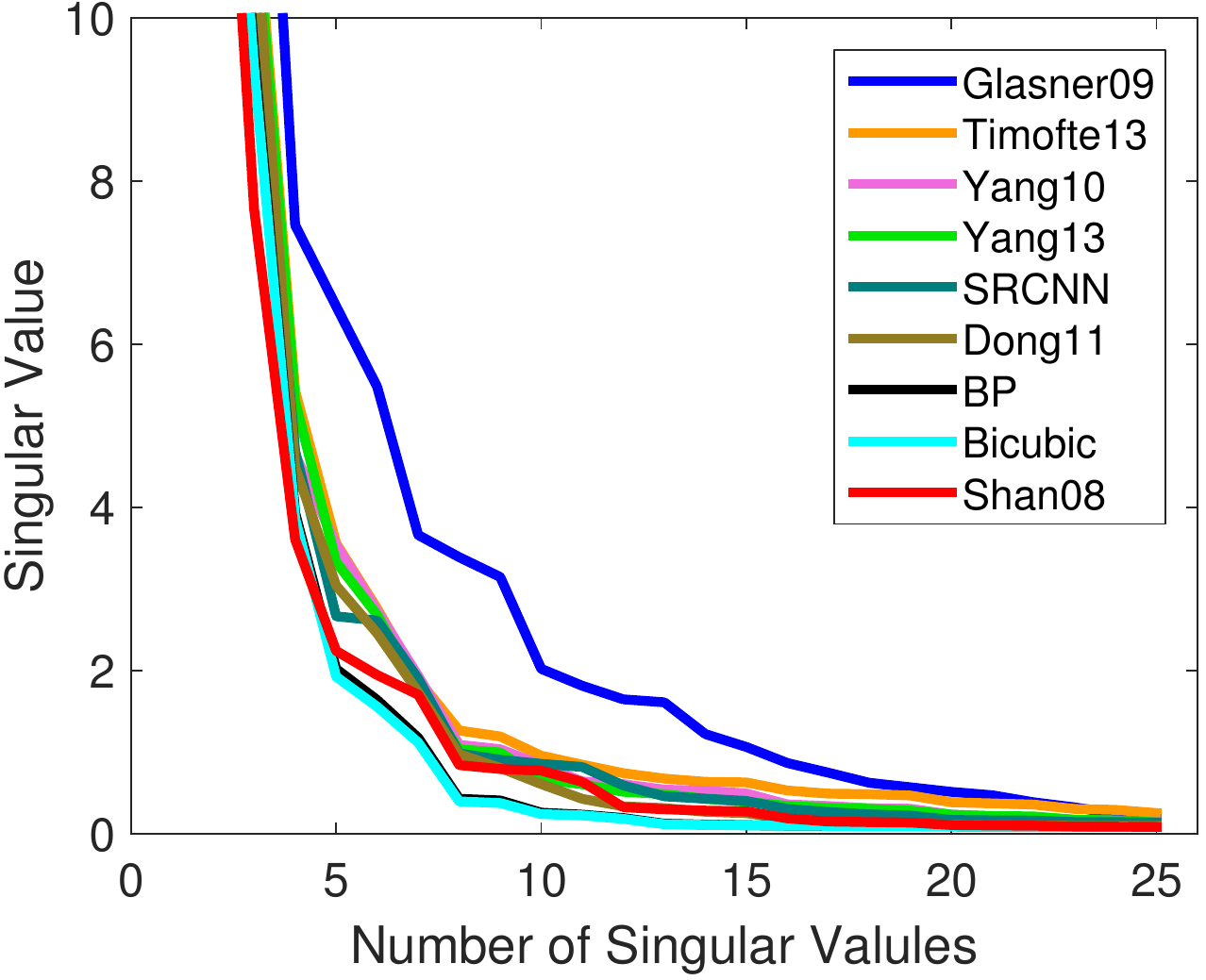} \\
		(a) & (b) & (c)\\
	\end{tabular}
	\caption{(a) Estimated GGD distribution of the normalized DCT
		coefficient in the first block of the images from Figure~\ref{fig:SRimage}.
		Note that the shape parameter $\gamma$ effectively characterizes the
		distribution difference between SR algorithms ($\mu$ is disregarded).
		(b) Wavelet coefficient distribution in one
		subband. 
		The GSM makes the distribution of subband more Gaussian-like (blue).
		(c) Distribution of patch singular values of SR images in Figure~\ref{fig:SRimage}.
		For SR images generated by Bicubic and BP containing more edge blur
		(smoothness), their singular values fall off more rapidly. 
		In contrast, Glasner09 strengthens the sharpness and the singular
		values of its generated SR image decrease more slowly. }
	\label{fig:feat}	
\end{figure}

\begin{figure}
	\small
	\centering
	\setlength{\tabcolsep}{2em}
	\begin{tabular}{cc}
		\includegraphics[width=.25\textwidth]{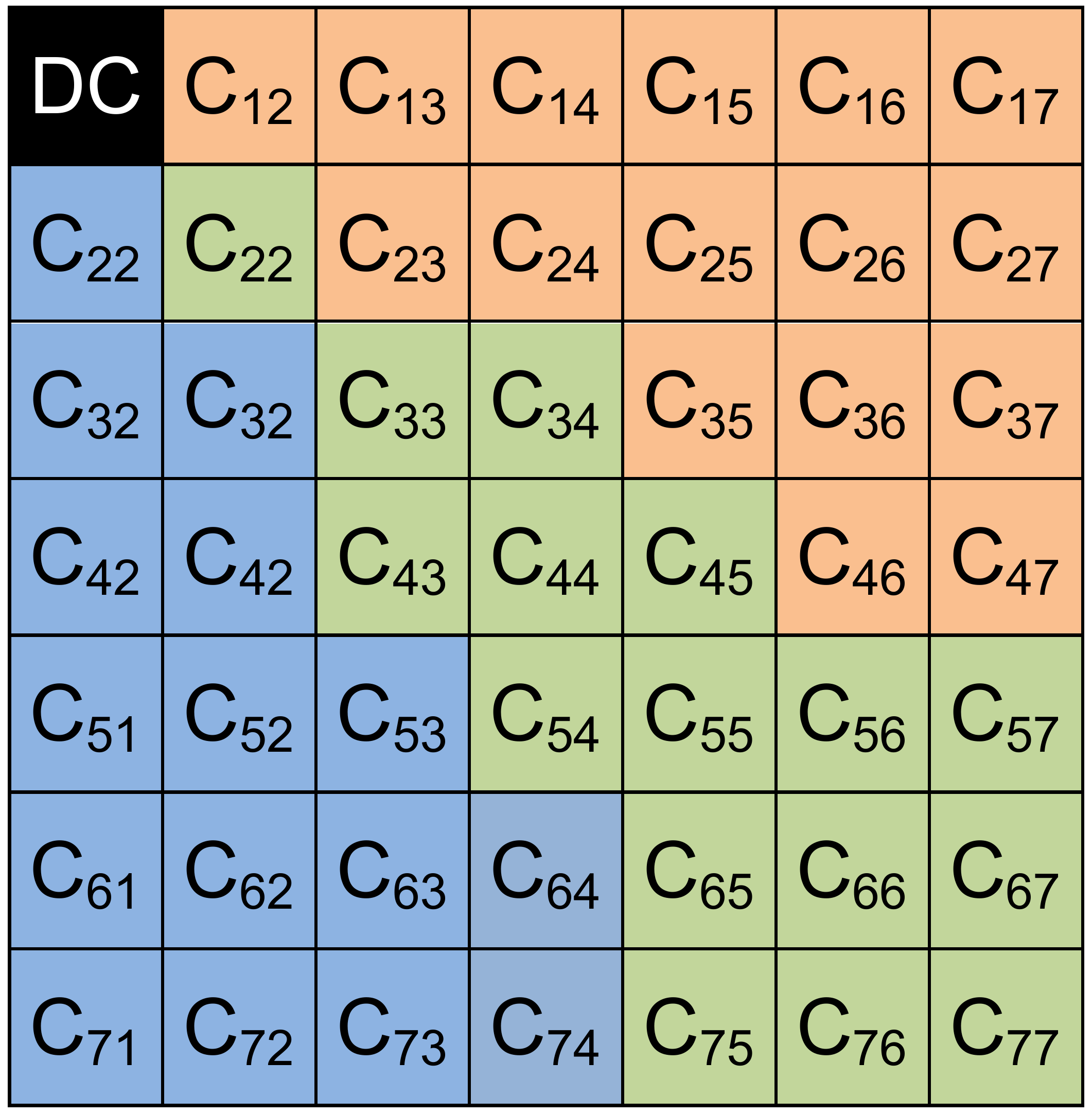} &
		\includegraphics[width=.4\textwidth]{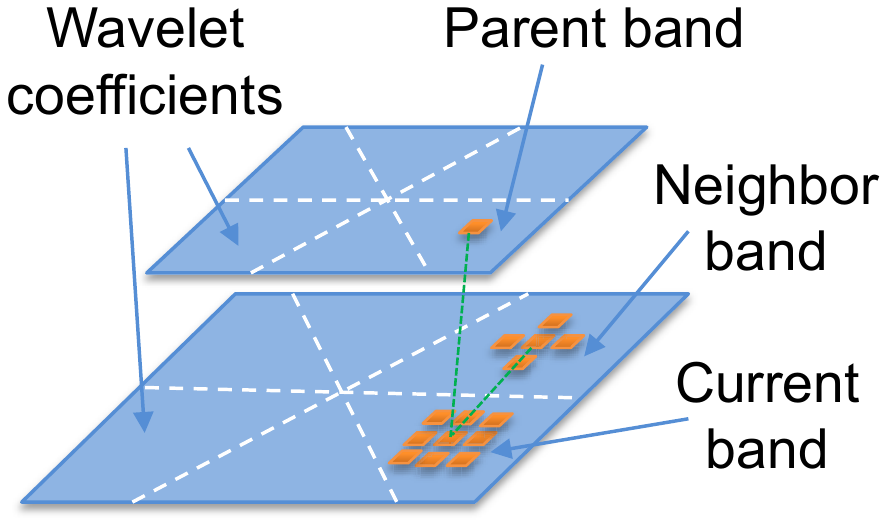} \\
		(a) & (b)
	\end{tabular}
	\caption{
		(a) Three groups of DCT coefficients for one block are shown by
		different colors. The DC coefficients are excluded.
		(b) $N = 15$ neighboring filters. $3\times3$ adjacent positions
		in the current band, 5 locations in the neighboring band 
		and 1 from the parent band.}
	\label{fig:bb}
\end{figure}

The statistics of coefficients from the discrete cosine transform (DCT) 
have been shown to effectively quantify the degree and type of image
degradation~\cite{DBLP:journals/jmiv/SrivastavaLSZ03},  
and used for natural image quality assessment~\cite{DBLP:journals/tip/SaadBC12}. 
Since SR images are generated from LR inputs, the task can be
considered as a restoration of high-frequency components on LR images.
To quantify the high-frequency artifacts introduced by SR restoration, 
we propose to transform SR images into the DCT domain and fit the DCT
coefficients by the generalized Gaussian distribution (GGD) as in \cite{DBLP:journals/tip/SaadBC12}.

\begin{equation}
\label{eq:GGD}
f(x|\mu,\gamma)= \frac{1}{2\Gamma(1+\gamma^{-1})}
e^{-(|x-\mu|^\gamma)},
\end{equation}
where $\mu$ is the mean of the random variable $x$,
$\gamma$ is the shape parameter and $\Gamma(\cdot)$
is the gamma function, e.g., $\Gamma(z)=\int_0^\infty t^{z-1}e^{-t}dt$.
We observe that the shape factor $\gamma$ is more
discriminative than the mean value $\mu$ to characterize the distribution
of DCT coefficients (See Figure~\ref{fig:feat}(a)).
We thus select the value of $\gamma$ as one statistical
feature to describe SR images.
Let $\sigma$ be the standard deviation of a DCT block,
we use $\bar{\sigma}=\frac{\sigma}{\mu}$  
to describe the perturbation within one block.
We further group DCT coefficients of each block 
into three sets (See Figure~\ref{fig:bb}(a)) and compute the
normalized deviation $\bar{\sigma}_i$ ($i=1,2,3$) of each set and their
variation $\Sigma$ of $\{\bar{\sigma}_i\}$ as features.
As all the statistics are computed on 
individual blocks, large bias is likely to be introduced if these measures
are simply concatenated.
We thus pool those block statistics and use the mean values
to represent each SR image.
To increase their discriminative strength, we add the first and
last 10\% pooled variations as features.

\subsection{Global Frequency Features}
The global distribution of the wavelet coefficients 
of one SR image might not be fitted well by a specific distribution (e.g., GGD). 
We sort to the Gaussian scale mixture (GSM) model, which shows effective 
in describing the marginal and joint statistics of natural
images~\cite{NIPS/Wainwright99b,DBLP:journals/tip/MoorthyB11} 
using a set of neighboring wavelet bands.
An $N$-dimensional random vector $Y$ belongs to a GSM if $Y\equiv
z\cdot U$, where $\equiv$ denotes equality in probability
distribution, 
and $U$ is a zero-mean Gaussian random vector with covariance $Q$.
The variable $z$ is a non-negative mixing multiplier.
The density of $Y$ is given by an integral as

\begin{equation}
p_Y(y)=\int\frac{1}{(2\pi)^{N/2}|z^2Q|^{1/2}} e^{\left
	(-\frac{Y^TQ^{-1}Y}{2z^2}\right )}p_z(z)dz,
\end{equation}
where $p_z(\cdot)$ is the probability of the mixing variable $z$.
We first apply the steerable pyramid
decomposition~\cite{DBLP:journals/tit/SimoncelliFAH92}
on an SR image to generate neighboring wavelet coefficients.
Compared to~\cite{NIPS/Wainwright99b,DBLP:journals/tip/MoorthyB11},
we apply the decomposition in both the real and imaginary domains 
rather than only in the real domain.
We observe that the wavelet coefficients in the complex domain have
better discriminative strength.
As shown in Figure~\ref{fig:bb}(b), we assume that $N$ (e.g.,
$N=15$) filters in neighborhoods that share a mixer estimated by
$\hat{z}=\sqrt{Y^TQ^{-1}Y/N}$.  
Such estimation is identical to divisive normalization 
\cite{DBLP:journals/tip/WangBSS04,DBLP:journals/tip/MoorthyB11}
and makes the probability distribution of wavelet band more
Gaussian-like (See Figure~\ref{fig:feat}(b)).
Let $d_\alpha^\theta$ be the normalized wavelet subband 
with scale $\alpha$ and orientation $\theta$.
We estimate the shape parameter $\gamma$ using (\ref{eq:GGD}) on 
$d_\alpha^\theta$ and concatenated bands $d^\theta$ across scales.
In addition, 
we compute the structural
correlation~\cite{DBLP:journals/tip/WangBSS04,DBLP:journals/tip/MoorthyB11} 
between high-pass response and their band-pass counterparts to measure 
the global SR artifacts. 
Specifically, the band-pass and high-pass responses 
are filtered across-scale by a $15\times15$ Gaussian window with
kernel width $\sigma=1.5$. 
The structural correlation is computed by
$\rho=\frac{2\sigma_{xy}+c_0}{\sigma_x^2+\sigma_y^2+c_0}$, 
where $\sigma_{xy}$ is the cross-covariance between the windowed regions;
$\sigma_x$ as well as $\sigma_y$ are their windowed variances; 
and $c_0$ is a constant for stabilization.

\subsection{Spatial Features}
Since the spatial discontinuity of pixel intensity is closely related
to perceptual scores for SR images in subject studies (See
Figure~\ref{fig:score}), 
we model this property in a way similar
to~\cite{DBLP:conf/icip/YeganehRW12}.
We extract features from patches rather than pixels to increase
discriminative strength. 
We apply principal component analysis (PCA) on patches
and use the corresponding singular values to describe the spatial
discontinuity. 

Singular values of images with smooth contents are squeezed to zero
more rapidly than for those with sharp contents (as they correspond
to less significant eigenvectors).
Figure~\ref{fig:feat}(c) shows the singular values of SR images 
generated from Bicubic and BP fall off more rapidly 
as the generated contents tend to be smooth.

\subsection{Two-stage Regression Model}
We model the features of local frequency,
global frequency and spatial discontinuity with three independent
regression forests~\cite{DBLP:journals/ml/Breiman01,MSR-TR-2011-114}.
Their outputs are linearly regressed on perceptual scores to predict the quality
of evaluated SR images.
Let $x_n$ ($n=1,2,3$) denote one type of low-level features, 
and $y$ be the perceptual scores of SR images.  
The $j$-th node of the $t$-th decision tree ($t=1,2,\ldots, T$) 
in the forest is learned as:
\begin{equation}
\theta_j^{n*}=\argmax_{\theta_j^n\in \mathcal{T}_j}I_j^n,
\end{equation}
where $\mathcal{T}_j$ 
controls the size of a random subset of training data to train node $j$.
The objective function $I_j^n$ is defined as:
\begin{equation}
I_j^n=\sum_{x_n\in\mathcal{S}_j}\log(|\Lambda_y(x_n)|)-\sum_{i\in
	\{L,R\}}\big(\sum_{x_n\in \mathcal{S}_j^i}\log(|\Lambda_y(x_n)|)\big) 
\end{equation}
with $\Lambda_y$ the conditional covariance matrix computed from probabilistic linear fitting,
where $\mathcal{S}_j$ denotes the set of training data arriving at node $j$, 
and $\mathcal{S}_j^L$, $\mathcal{S}_j^R$ the left and right split sets.
We refer readers to \cite{MSR-TR-2011-114} for more details about regression forest.
The predicted score $\hat{y}_n$ is thus
computed by averaging the outputs of $T$ regression trees as:
\begin{equation}
\hat{y}_n=\frac{1}{T}\sum_t^Tp_t(x_n|\Theta).
\end{equation}
Consequently, we linearly regress the outputs from all three types of features to perceptual scores, 
and estimate the final quality score as $\hat{y}=\sum_n\lambda_n\cdot\hat{y}_n$,
where the weight $\lambda$ is learned by minimizing 
\begin{equation}
\lambda^*=\argmin_{\lambda}(\sum_n\lambda_n\cdot\hat{y}_n-y)^2.
\end{equation}

\section{Experimental Validation}
\label{sec:experiment}
In the human subject studies, we generate 1,620 SR images from 180 LR
inputs using 9 different SR algorithms, and collect their perceptual scores
from 50 subjects. The mean of the median 40 subject scores is used as perceptual score.
We randomly split the dataset into 5 sets, and 
recursively select one set for test and the remaining for training. 
After this loop, we obtain the quality scores estimated by the proposed metric for all SR images. 
We then compare the Spearman rank correlation coefficients between the 
predicted quality scores and perceptual scores.
In addition to the 5-fold cross validation, we split the training and test sets according to
the reference images and SR methods 
to verify the generality of the proposed metric.
Given that there are 30 reference images and 9 SR methods,
we leave $6$ reference images or 
2 methods out in each experiment.
Several state-of-the-art no-reference IQA methods and 4 most
widely used full-reference metrics for SR images are included for
experimental validation.
More results and the source code of the proposed metric can be found at  \url{https://sites.google.com/site/chaoma99/sr-metric}.

\begin{table}
	\caption{List of features used in this work.}
	\label{tb:feature}
	\centering
	\small
	\setlength{\tabcolsep}{1em}
	\begin{tabular}{ |c|c|c| }
		\hline
		Feature domain & Feature Description  &  \# \\
		\hline\hline
		\multirow{3}*{Local frequency} &  $\gamma$ (mean, first 10\% percentile) & 6  \\
		\cline{2-3}
		&     $\bar{\sigma}$ (mean, last 10\% percentile)                       &  6   \\
		\cline{2-3}
		& $\Sigma$ (mean, last 10\% percentile)            & 6   \\
		\hline
		\multirow{3}*{Global frequency} &  $\gamma$ for each band $d_\alpha^\theta$ and $d^\theta$  & 18   \\
		\cline{2-3}
		&  Across-scale  correlation             & 12    \\
		\cline{2-3}
		&  Across-band  correlation                   & 15    \\
		\hline
		Spatial discontinuity & Singular values of patches & 75  \\
		\hline
		Total & & 138 \\
		\hline
	\end{tabular}
\end{table}

\begin{figure}
	\centering
	\small
	\includegraphics[width=0.8\textwidth]{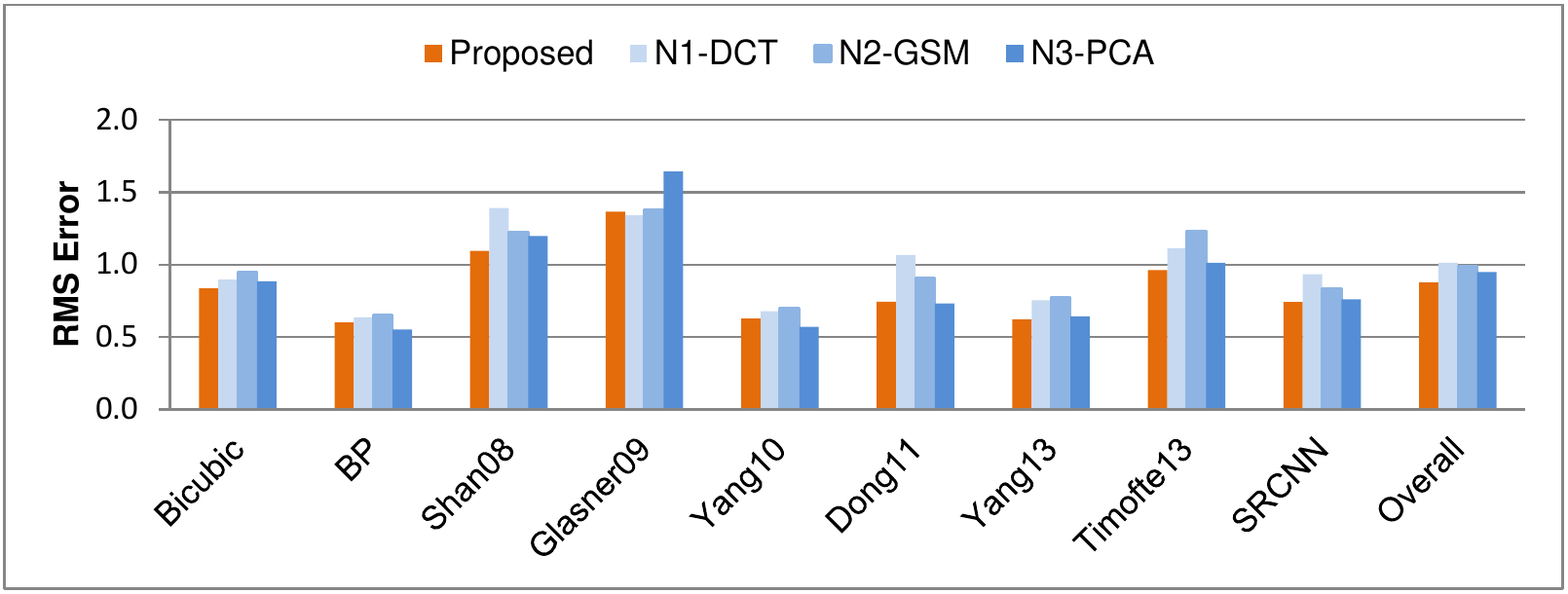}  \\
	(a) 5-fold cross validation \\
	\includegraphics[width=0.8\textwidth]{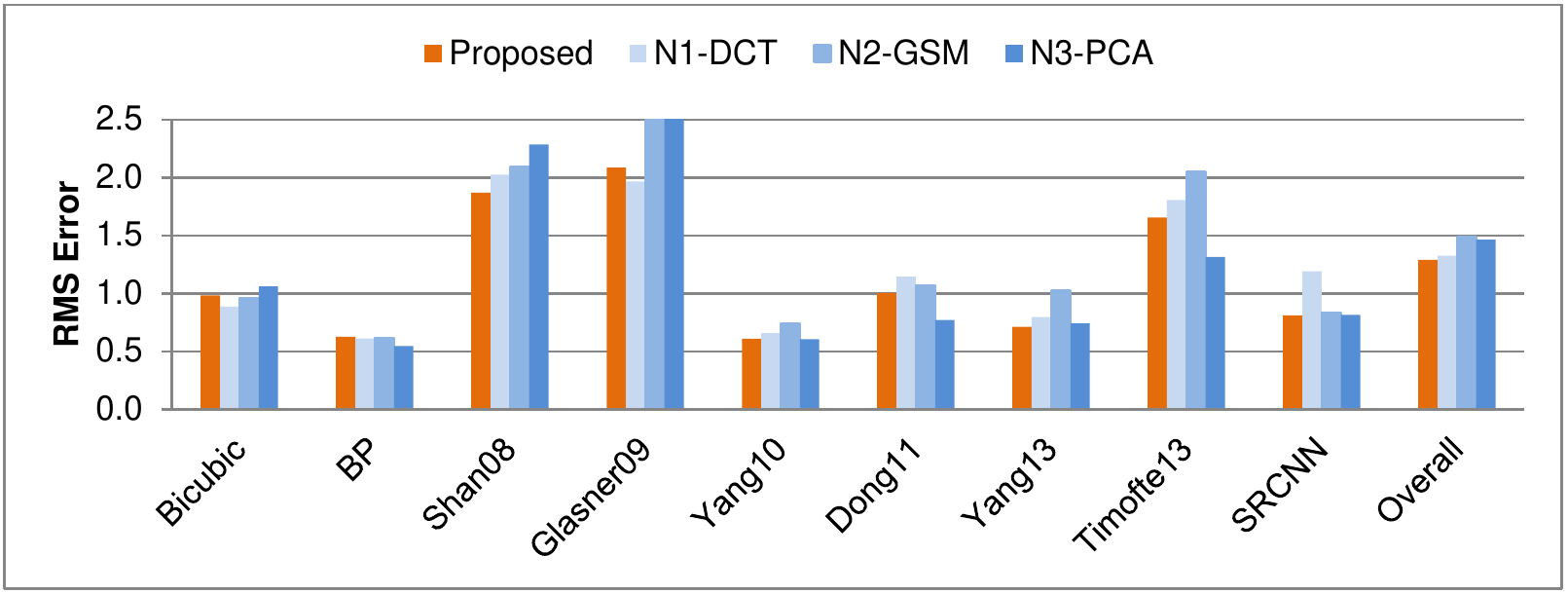} \\
	(b) Leaving 6 reference images out \\
	\includegraphics[width=0.8\textwidth]{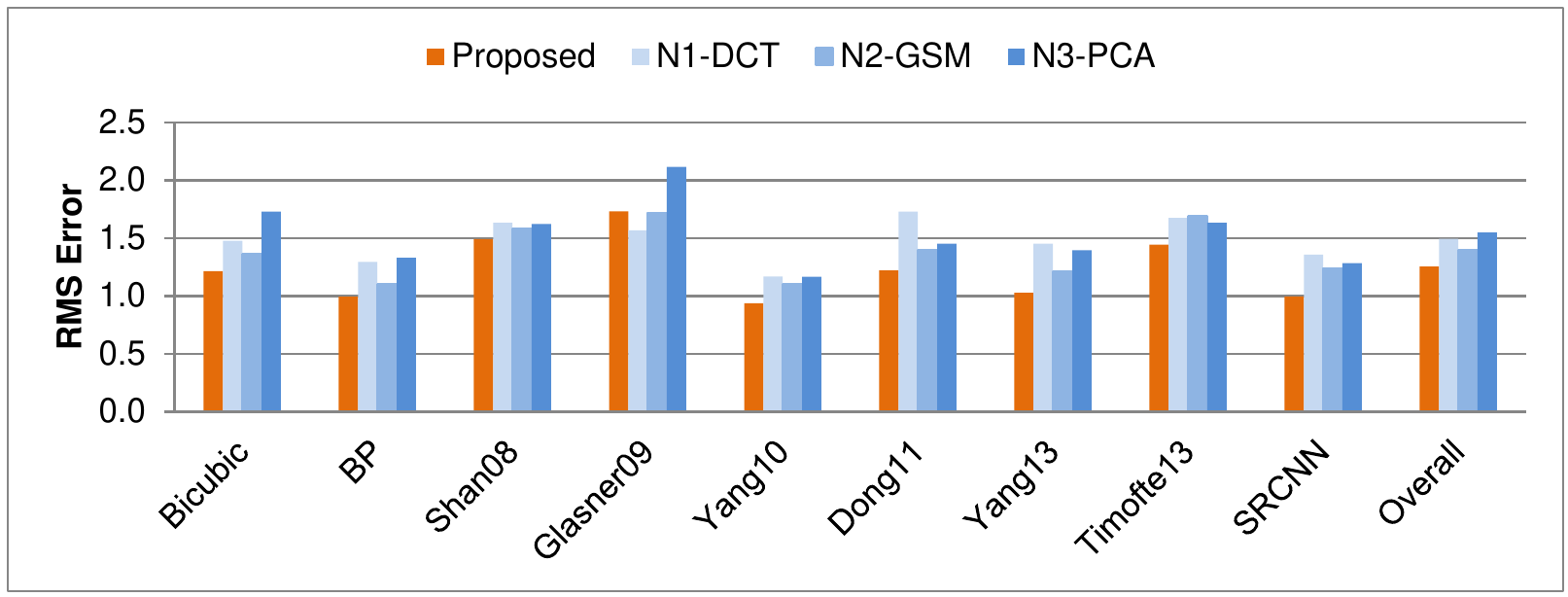} \\
	(c) Leaving 2 SR methods out \\ 
	\caption{Root-mean-square error between the estimated score and
		the subjective score 
		(measures with smaller values are closer to human
		visual perception) using 3 validation schemes. 
		Note that the proposed two-stage regression model (orange bar) on
		three types of low-level  
		features (blue bar) reduces the error between perceptual scores
		significantly. } 
	\label{fig:RMSE}
\end{figure}

\subsection{Parameter Settings}
We use a three-level 
pyramid on $7\times7$ blocks of DCT coefficients
to compute local frequency features.
For steerable pyramid wavelet decomposition, we set
$\alpha$ and $\theta$ to be 2 and 6, respectively. 
The resulting 12 subbands are denoted by $s_\alpha^\theta$, where
$\alpha\in\{1,2\}$ and
$\theta\in\{0^\circ,30^\circ,60^\circ,90^\circ,120^\circ,150^\circ\}$.
We set the number $N$ of neighboring filters to 15,
i.e., $3\times3$ adjacent positions in the current band,
5 adjacent locations in the neighboring band and 1 from the parent
band share a mixer (See Figure~\ref{fig:bb}(b)).
For spatial discontinuity, we compute singular values on $5\times 5$
patches on a three-level pyramid.
We list the detailed feature information in Table~\ref{tb:feature}.
We vary the parameter $T$ of regression trees 
from 100 to 5000 with a step of 50 and 
find the proposed algorithm performs best when $T$ is set to 2000.

\begin{table}
	\caption{Spearman rank correlation coefficients~\cite{hogg2005introduction}
		(metric with higher coefficient matches perceptual score better). The random forest regression (RFR) uniformly performs better than the support vector regression (SVR) for each type of features or the concatenation (-con) of three type of features. The proposed two-stage regression approach (-all) combining three types of features improves the accuracy for both RFR and SVR. Bold: best; underline: second best.} 
	\label{tb:rfrsvr}
	\resizebox{\textwidth}{!}{
		\setlength{\tabcolsep}{.0em}
		\centering
		\small
		\begin{tabular}{ | p{5em} || *{5}{p{4.5em}<{\centering}|} | *{5}{p{4.5em}<{\centering}|} }
			\hline 
          & Ours           & RFR-con     & RFR-DCT & RFR-GSM & RFR-PCA        & SVR-all & SVR-con & SRV-DCT & SVR-GSM & SVR-PCA \\\hline
          ~ Bicubic   & \textbf{0.933} & 0.922       & 0.910   & 0.898   & {\ul 0.923}    & 0.851   & 0.772   & 0.630   & 0.713   & 0.862   \\
          ~ BP        & \textbf{0.966} & {\ul 0.962} & 0.956   & 0.952   & \textbf{0.966} & 0.881   & 0.876   & 0.776   & 0.838   & 0.889   \\
          ~ Shan08    & \textbf{0.891} & {\ul 0.887} & 0.830   & 0.870   & 0.874          & 0.504   & 0.373   & 0.499   & 0.522   & 0.044   \\
          ~ Glasner09 & \textbf{0.931} & {\ul 0.926} & 0.911   & 0.897   & 0.878          & 0.841   & 0.717   & 0.766   & 0.685   & 0.599   \\
          ~ Yang10    & {\ul 0.968}    & 0.961       & 0.954   & 0.948   & \textbf{0.969} & 0.929   & 0.905   & 0.874   & 0.834   & 0.877   \\
          ~ Dong11    & {\ul 0.954}    & 0.946       & 0.922   & 0.929   & \textbf{0.960} & 0.885   & 0.892   & 0.792   & 0.883   & 0.874   \\
          ~ Yang13    & \textbf{0.958} & {\ul 0.955} & 0.937   & 0.932   & \textbf{0.958} & 0.898   & 0.855   & 0.801   & 0.770   & 0.874   \\
          ~ Timofte13 & \textbf{0.930} & {\ul 0.928} & 0.911   & 0.880   & 0.927          & 0.883   & 0.814   & 0.859   & 0.628   & 0.839   \\
          ~ SRCNN     & \textbf{0.949} & 0.938       & 0.917   & 0.936   & {\ul 0.945}    & 0.866   & 0.853   & 0.778   & 0.816   & 0.843   \\\hline
          ~ Overall   & \textbf{0.931} & {\ul 0.921} & 0.909   & 0.913   & {\ul 0.921}    & 0.752   & 0.696   & 0.711   & 0.616   & 0.663 \\\hline
		\end{tabular}
	}
\end{table}

\begin{table}
	\caption{Spearman rank correlation coefficients~\cite{hogg2005introduction}
		(metric with higher coefficient matches perceptual score better). The compared no-reference metrics are re-trained on our SR dataset using the 5-fold cross validation. The proposed metric performs favorably against state-of-the-art methods. Bold: best; underline: second best.} 
	\label{tb:cv}
	\resizebox{\textwidth}{!}{
		\setlength{\tabcolsep}{0em}
		\centering
		\small
		\begin{tabular}{ | p{5.5em} || *{6}{p{4.2em}<{\centering}|} | *{6}{p{4.2em}<{\centering}|}}
			\hline
			&  Ours & BRISQUE & BLIINDS & CORNIA & CNNIQA  & NSSA & DIVINE & BIQI & IFC & SSIM &  FSIM & PSNR\\
			& & \cite{DBLP:journals/tip/MittalMB12} 
			& \cite{DBLP:journals/tip/SaadBC12}
			& \cite{DBLP:conf/cvpr/YeKKD12}
			& \cite{DBLP:conf/cvpr/KangYLD14}
			& \cite{DBLP:conf/icip/YeganehRW12}
			& \cite{DBLP:journals/tip/MoorthyB11}
			& \cite{DBLP:journals/spl/MoorthyB10}
			& \cite{DBLP:journals/tip/SheikhBV05}
			& \cite{DBLP:journals/tip/WangBSS04}
			& \cite{DBLP:journals/tip/ZhangZMZ11} & \\
			\hline\hline
			~ Bicubic   & \textbf{0.933} & 0.850 & 0.886 & 0.889          & {\ul 0.926} & -0.007 & 0.784 & 0.770 & 0.884 & 0.588 & 0.706 & 0.572 \\
			~ BP        & \textbf{0.966} & 0.917 & 0.931 & 0.932          & {\ul 0.956} & 0.022  & 0.842 & 0.740 & 0.880 & 0.657 & 0.770 & 0.620 \\
			~ Shan08    & {\ul 0.891}    & 0.667 & 0.664 & \textbf{0.907} & 0.832       & -0.128 & 0.653 & 0.254 & 0.934 & 0.560 & 0.648 & 0.564 \\
			~ Glasner09 & \textbf{0.931} & 0.738 & 0.862 & {\ul 0.918}    & 0.914       & 0.325  & 0.426 & 0.523 & 0.890 & 0.648 & 0.778 & 0.605 \\
			~ Yang10    & \textbf{0.968} & 0.886 & 0.901 & 0.908          & {\ul 0.943} & 0.036  & 0.525 & 0.556 & 0.866 & 0.649 & 0.757 & 0.625 \\
			~ Dong11    & \textbf{0.954} & 0.783 & 0.811 & 0.912          & {\ul 0.921} & 0.027  & 0.763 & 0.236 & 0.865 & 0.649 & 0.765 & 0.634 \\
			~ Yang13    & \textbf{0.958} & 0.784 & 0.864 & 0.923          & {\ul 0.927} & 0.168  & 0.537 & 0.646 & 0.870 & 0.652 & 0.768 & 0.631 \\
			~ Timofte13 & \textbf{0.930} & 0.843 & 0.903 & 0.911          & {\ul 0.924} & 0.320  & 0.122 & 0.563 & 0.881 & 0.656 & 0.756 & 0.620 \\
			~ SRCNN     & \textbf{0.949} & 0.812 & 0.843 & 0.898          & {\ul 0.908} & 0.165  & 0.625 & 0.617 & 0.885 & 0.660 & 0.780 & 0.645 \\\hline
			~ Overall   & \textbf{0.931} & 0.802 & 0.853 & {\ul 0.919}    & 0.904       & 0.076  & 0.589 & 0.482 & 0.810 & 0.635 & 0.747 & 0.604 \\\hline
		\end{tabular}
	}
\end{table}

\subsection{Quantitative Validations}
We run the proposed measure 100 times in each validation and choose
the mean values as the estimated quality scores.
We compare the contribution of each feature type 
using root-mean-square errors (RMSEs) in Figure~\ref{fig:RMSE}. 

The small overall error values, 0.87 in (a) and less than 1.4 in (b)
and (c) compared to the score range (0 to 10), indicate the
effectiveness of the proposed method 
by linearly combining three types of statistical features. 
In addition, 
we carry out an ablation study replacing the random forest regression (RFR) by the support vector regression (SVR) on each type of features. The SVR model is widely used in existing no-reference image quality metrics \cite{DBLP:journals/tip/MittalMB12, DBLP:journals/tip/SaadBC12, DBLP:journals/tip/MoorthyB11, DBLP:journals/spl/MoorthyB10, 
	DBLP:conf/cvpr/YeKKD12}. 
Table \ref{tb:rfrsvr} shows that RFR is more robust to the outliers than SVR on each type of features or a simple concatenation of three types of features. The proposed two stage-regression model effectively exploits three types of features and performs best. 

\begin{table}
	\caption{Spearman rank correlation coefficients~\cite{hogg2005introduction}
		(metric with higher coefficient matches perceptual score better). The compared metrics are retrained on our SR dataset under the leave-image-out validation. Bold: best; underline: second best.} 
	\label{tb:leaveimage}
	\centering
	\resizebox{.82\textwidth}{!}{
		\setlength{\tabcolsep}{0em}
		\small
		\begin{tabular}{| p{6em} || p{5em}<{\centering}|p{5em}<{\centering}|p{5em}<{\centering}|p{5em}<{\centering}|p{5em}<{\centering}|p{5em}<{\centering}|}
			\hline
			&  Ours & BRISQUE & BLIINDS & CORNIA & CNNIQA  & NSSA \\
			& & \cite{DBLP:journals/tip/MittalMB12} 
			& \cite{DBLP:journals/tip/SaadBC12}
			& \cite{DBLP:conf/cvpr/YeKKD12}
			& \cite{DBLP:conf/cvpr/KangYLD14}
			& \cite{DBLP:conf/icip/YeganehRW12} \\\hline\hline
			~ Bicubic   & \textbf{0.805} & 0.423 & 0.522 & {\ul 0.761}    & 0.736       & 0.093  \\
			~ BP        & \textbf{0.893} & 0.539 & 0.476 & {\ul 0.873}    & 0.853       & -0.046 \\
			~ Shan08    & {\ul 0.800}    & 0.442 & 0.474 & \textbf{0.832} & 0.742       & 0.048  \\
			~ Glasner09 & \textbf{0.867} & 0.277 & 0.399 & {\ul 0.859}    & 0.803       & 0.023  \\
			~ Yang10    & \textbf{0.904} & 0.625 & 0.442 & 0.843          & {\ul 0.867} & 0.012  \\
			~ Dong11    & \textbf{0.875} & 0.527 & 0.411 & 0.819          & {\ul 0.849} & -0.101 \\
			~ Yang13    & \textbf{0.885} & 0.575 & 0.290 & {\ul 0.843}    & 0.841       & 0.108  \\
			~ Timofte13 & {\ul 0.815}    & 0.500 & 0.406 & \textbf{0.828} & 0.740       & -0.035 \\
			~ SRCNN     & \textbf{0.904} & 0.563 & 0.383 & 0.827          & {\ul 0.850} & 0.042  \\\hline
			~ Overall   & \textbf{0.852} & 0.505 & 0.432 & {\ul 0.843}    & 0.799       & 0.017  \\\hline
		\end{tabular}
	}
\end{table}

\begin{table}
	\caption{Spearman rank correlation coefficients~\cite{hogg2005introduction}
		(metric with higher coefficient matches perceptual score better). The compared metrics are retrained on our SR dataset under the leave-method-out validation. Bold: best; underline: second best.} 
	\label{tb:leavemethod}
	\centering
	\resizebox{.82\textwidth}{!}{
		\setlength{\tabcolsep}{0em}
		\small
		\begin{tabular}{| p{6em} || p{5em}<{\centering}|p{5em}<{\centering}|p{5em}<{\centering}|p{5em}<{\centering}|p{5em}<{\centering}|p{5em}<{\centering}|}
			\hline
			&  Ours & BRISQUE & BLIINDS & CORNIA & CNNIQA  & NSSA \\
			& & \cite{DBLP:journals/tip/MittalMB12} 
			& \cite{DBLP:journals/tip/SaadBC12}
			& \cite{DBLP:conf/cvpr/YeKKD12}
			& \cite{DBLP:conf/cvpr/KangYLD14}
			& \cite{DBLP:conf/icip/YeganehRW12} \\
			\hline\hline
			~ Bicubic   & {\ul 0.932}    & 0.850 & 0.929       & 0.893          & \textbf{0.941} & 0.036  \\
			~ BP        & {\ul 0.967}    & 0.934 & 0.953       & 0.938          & \textbf{0.971} & 0.021  \\
			~ Shan08    & \textbf{0.803} & 0.534 & 0.471       & {\ul 0.799}    & 0.767          & -0.087 \\
			~ Glasner09 & \textbf{0.913} & 0.677 & 0.805       & 0.817          & {\ul 0.883}    & 0.393  \\
			~ Yang10    & \textbf{0.965} & 0.834 & 0.895       & 0.914          & {\ul 0.930}    & -0.054 \\
			~ Dong11    & \textbf{0.932} & 0.774 & 0.780       & 0.917          & {\ul 0.920}    & -0.062 \\
			~ Yang13    & \textbf{0.944} & 0.716 & 0.845       & {\ul 0.911}    & 0.906          & 0.147  \\
			~ Timofte13 & 0.774          & 0.760 & {\ul 0.849} & \textbf{0.898} & 0.845          & 0.382  \\
			~ SRCNN     & \textbf{0.933} & 0.771 & 0.806       & {\ul 0.908}    & 0.890          & 0.149  \\\hline
			~ Overall   & \textbf{0.848} & 0.644 & 0.763       & {\ul 0.809}    & 0.797          & 0.053  \\\hline
		\end{tabular}
	}
\end{table}

\begin{figure}
	\centering
	\small
	\setlength{\tabcolsep}{0em}
	\begin{tabular}{ccc}
		\includegraphics[width=.33\textwidth]{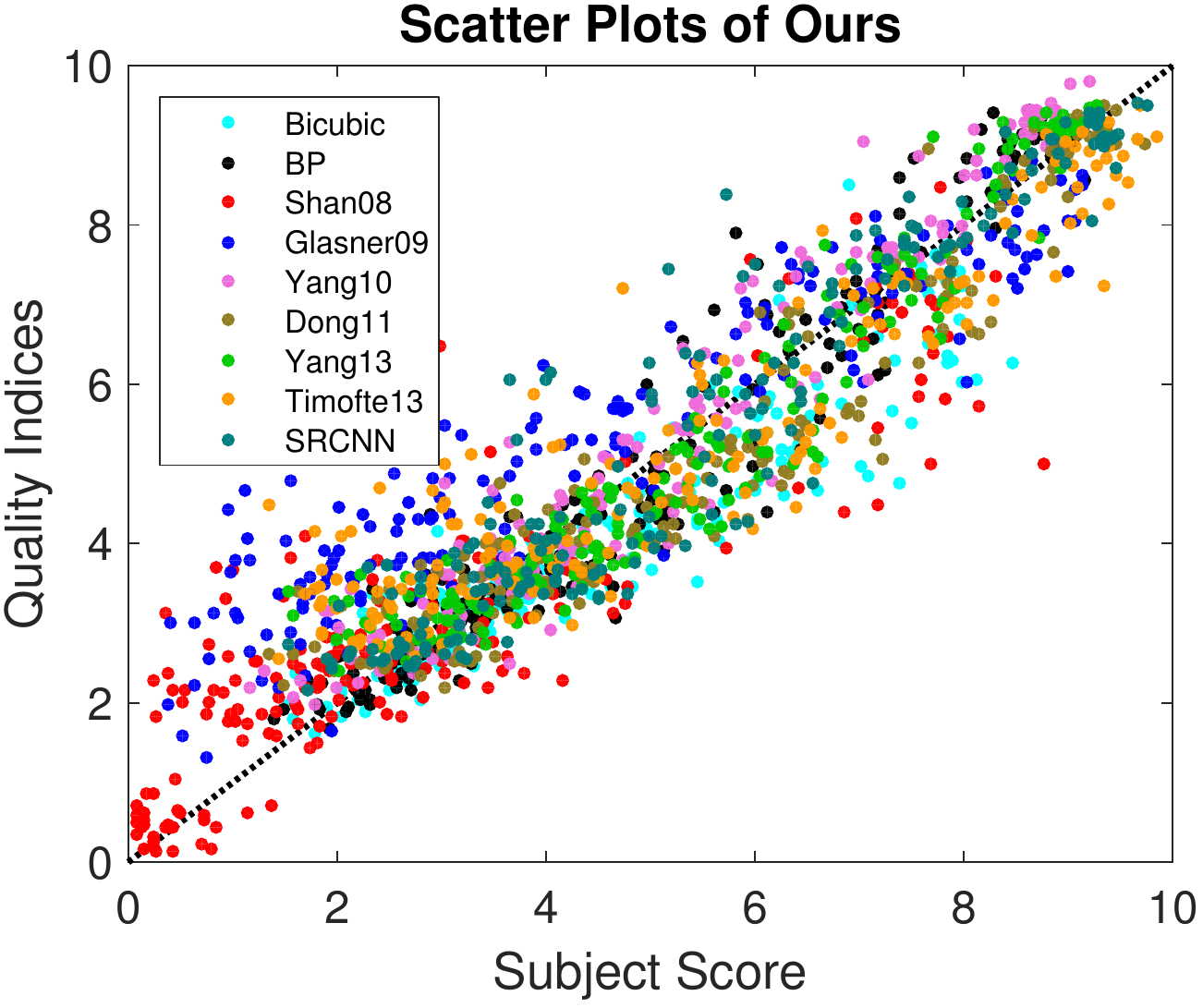}&
		\includegraphics[width=.33\textwidth]{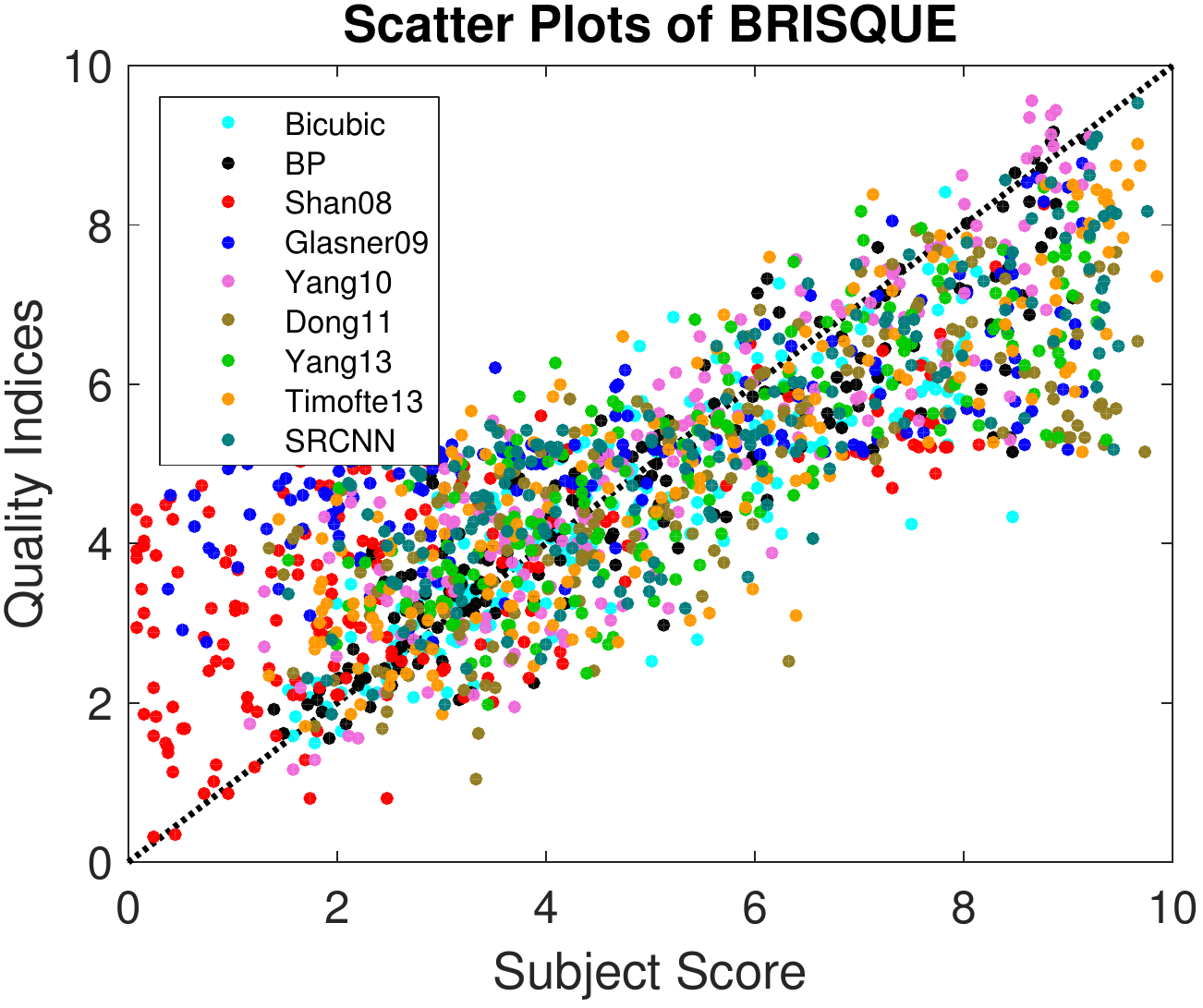}&
		\includegraphics[width=.33\textwidth]{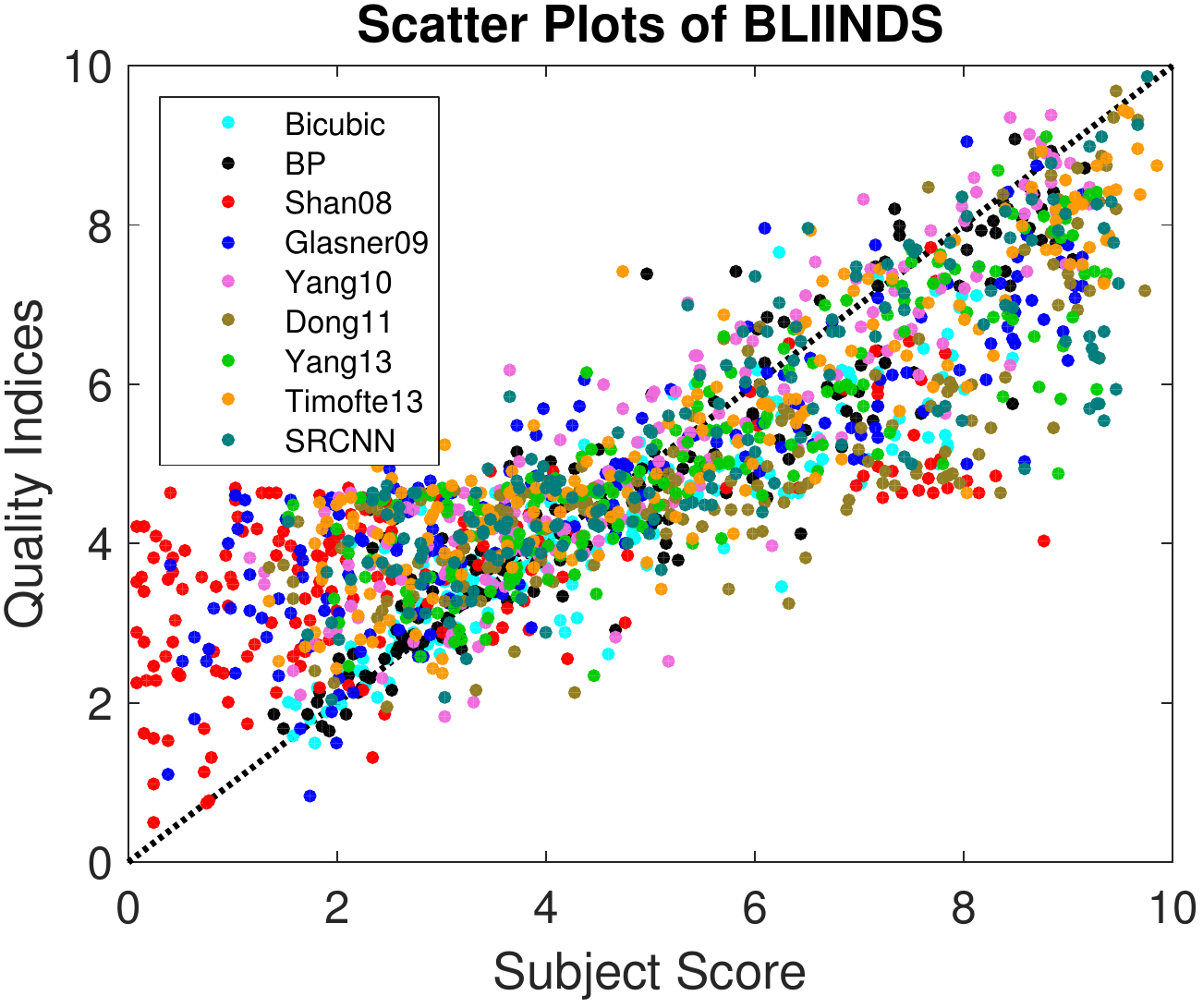} \\
		\includegraphics[width=.33\textwidth]{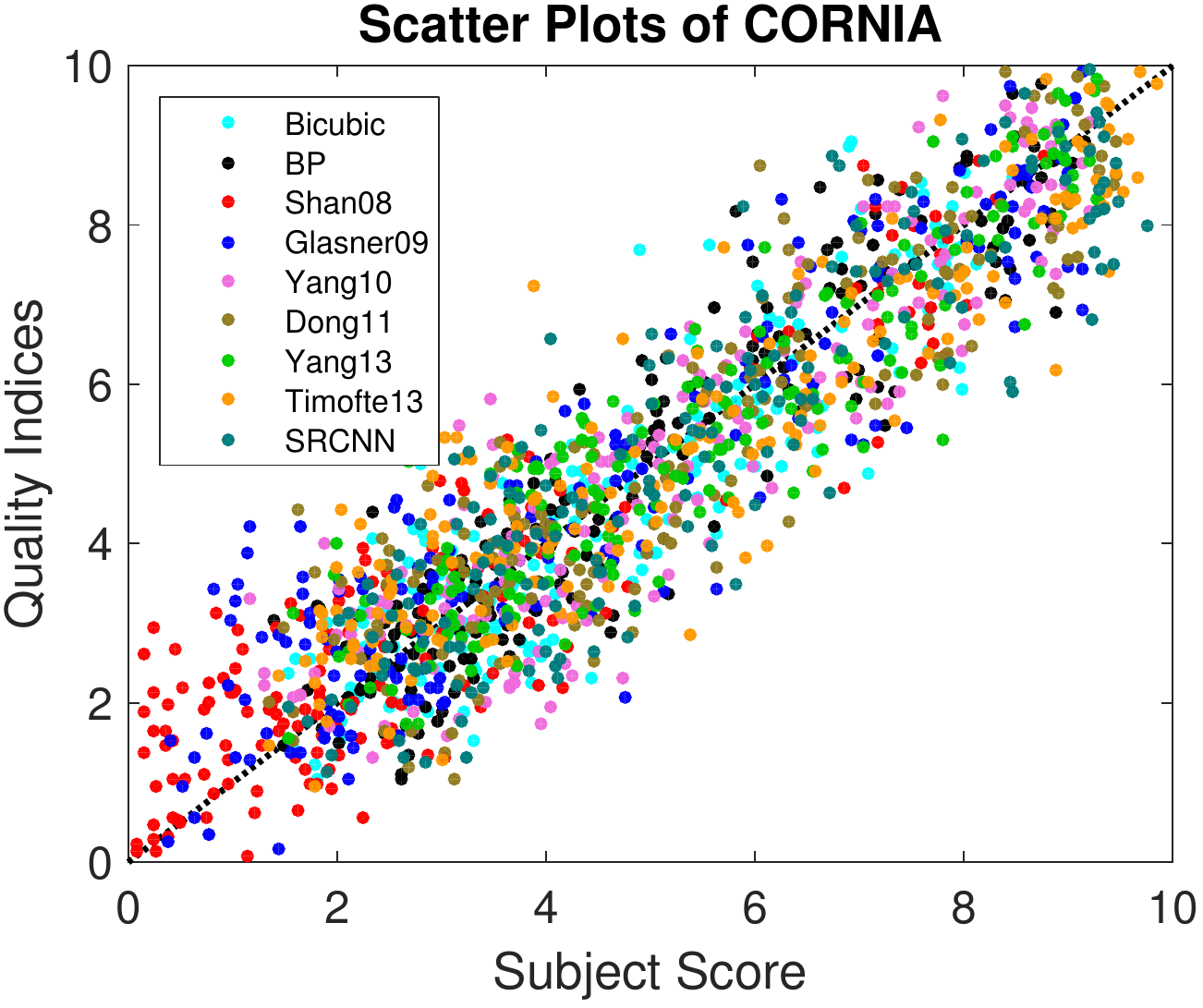} &
		\includegraphics[width=.33\textwidth]{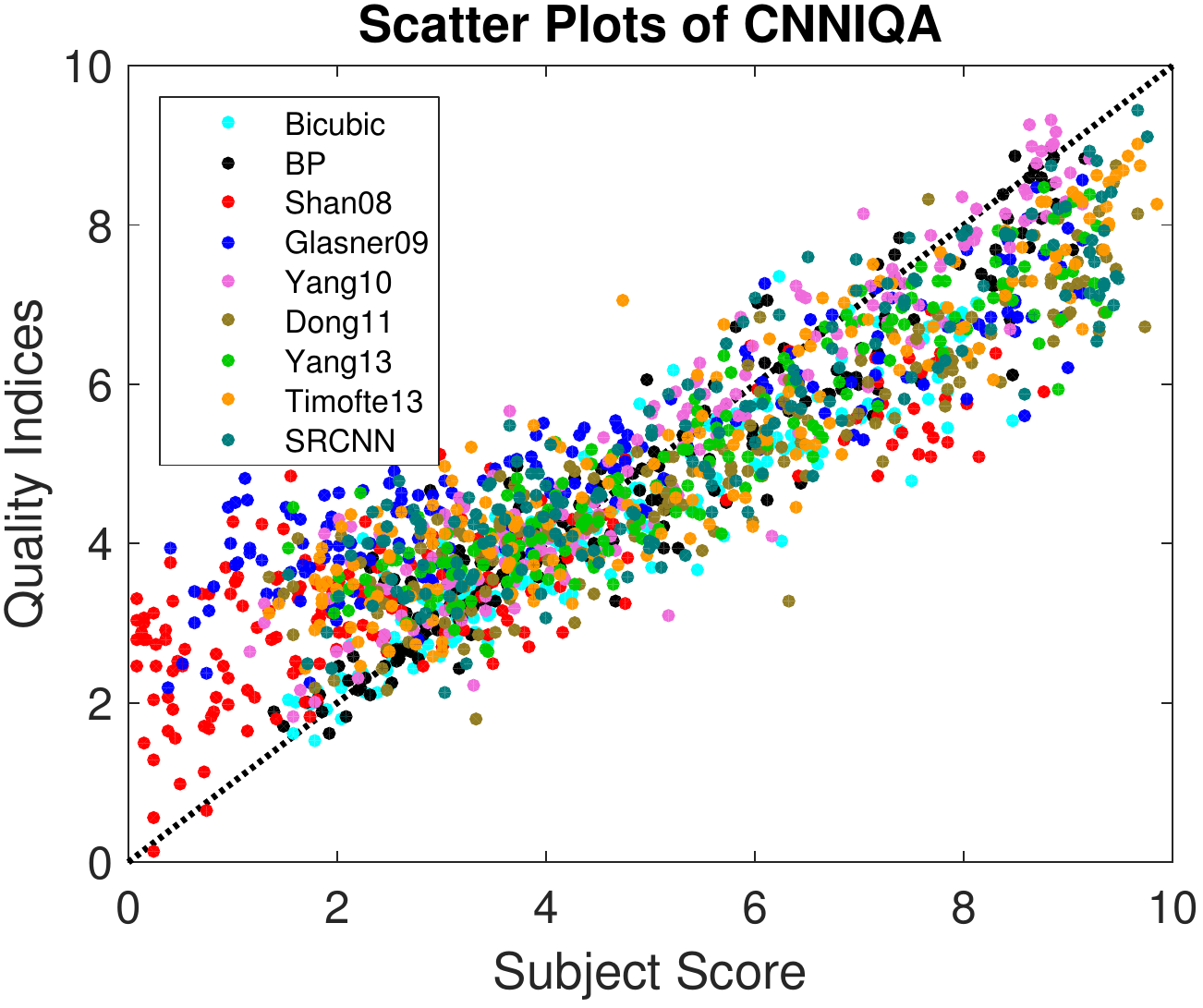} &
		\includegraphics[width=.33\textwidth]{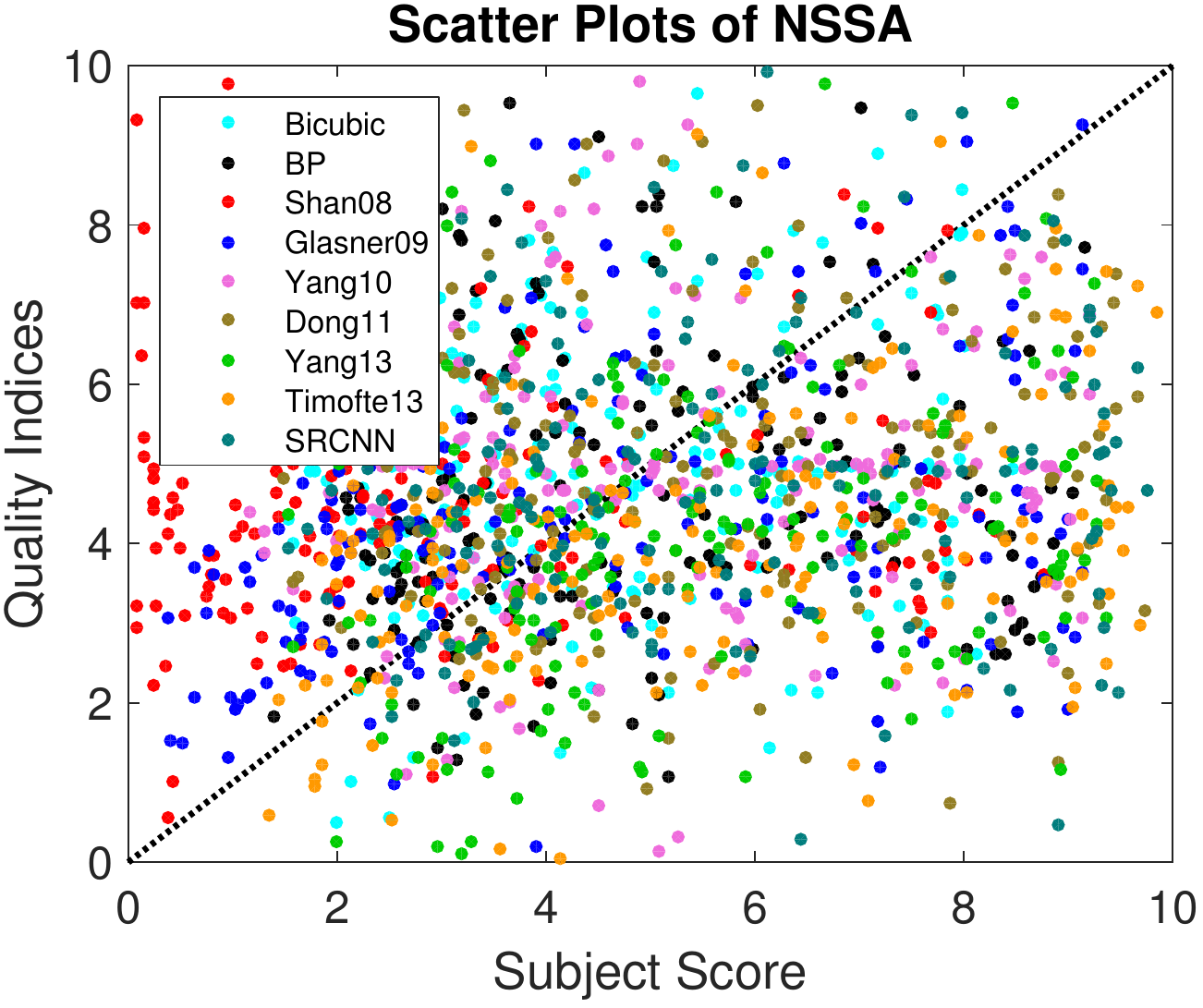} \\
		\includegraphics[width=.33\textwidth]{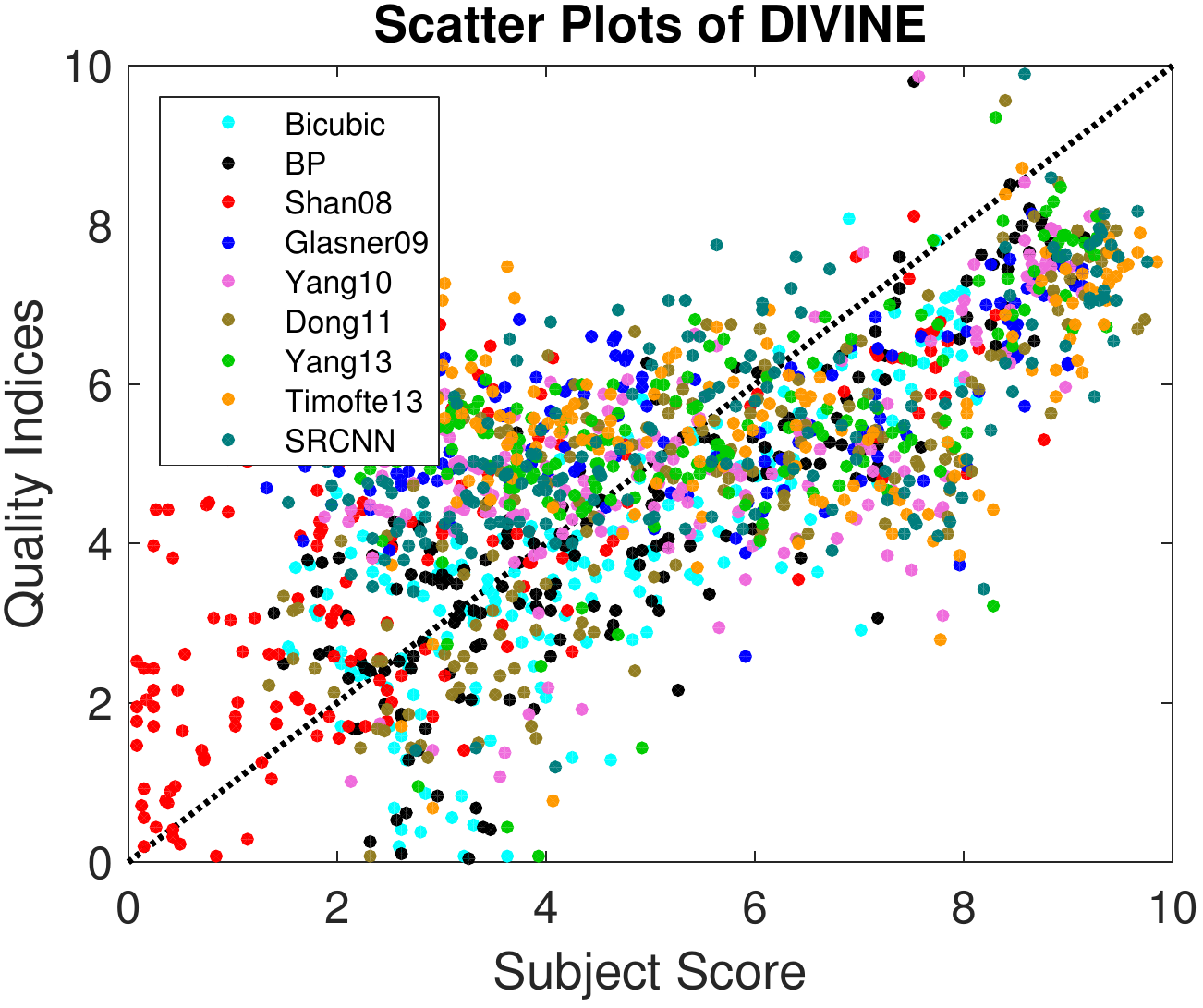} &
		\includegraphics[width=.33\textwidth]{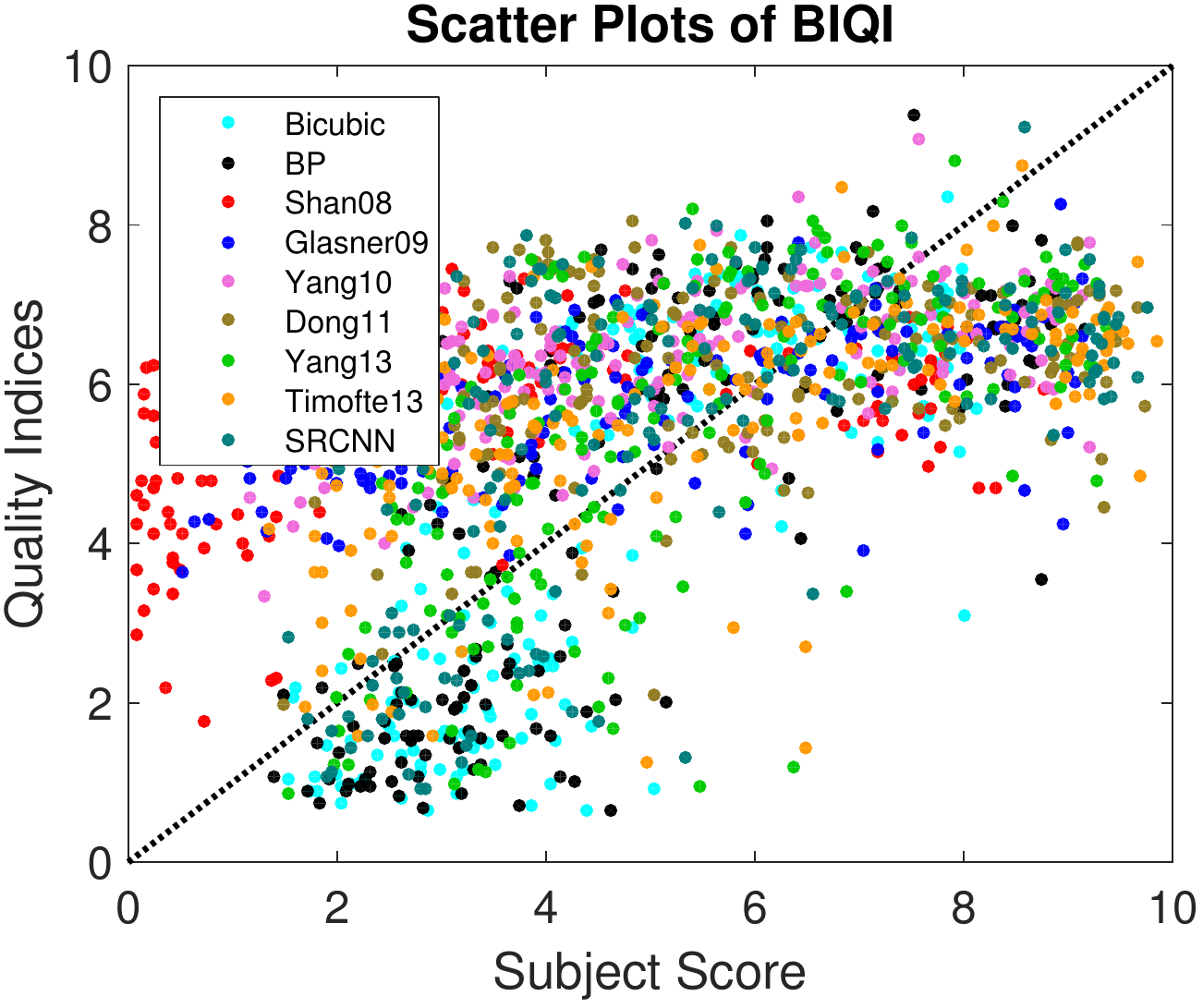} &
		\includegraphics[width=.33\textwidth]{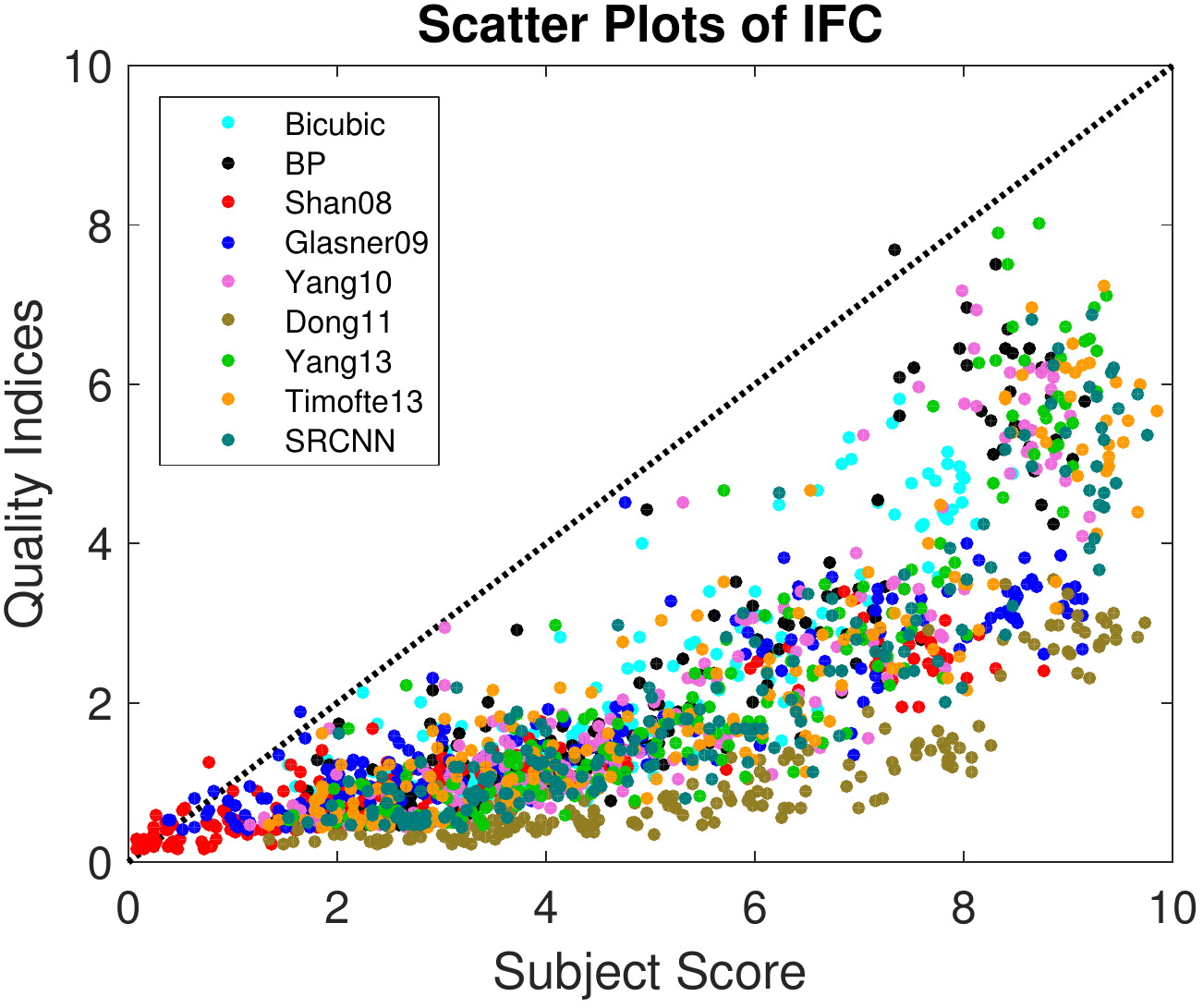}\\
		\includegraphics[width=.33\textwidth]{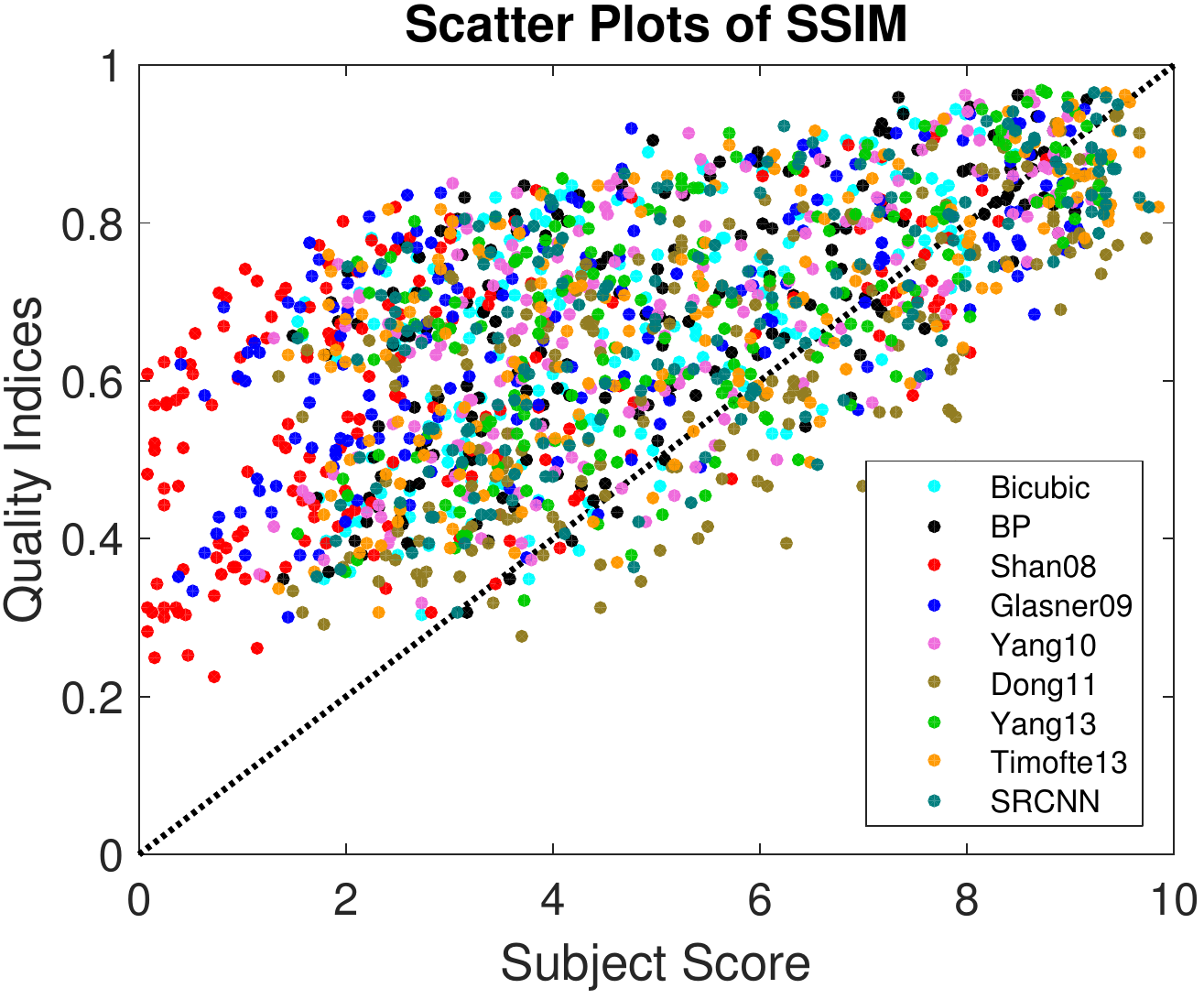} &
		\includegraphics[width=.33\textwidth]{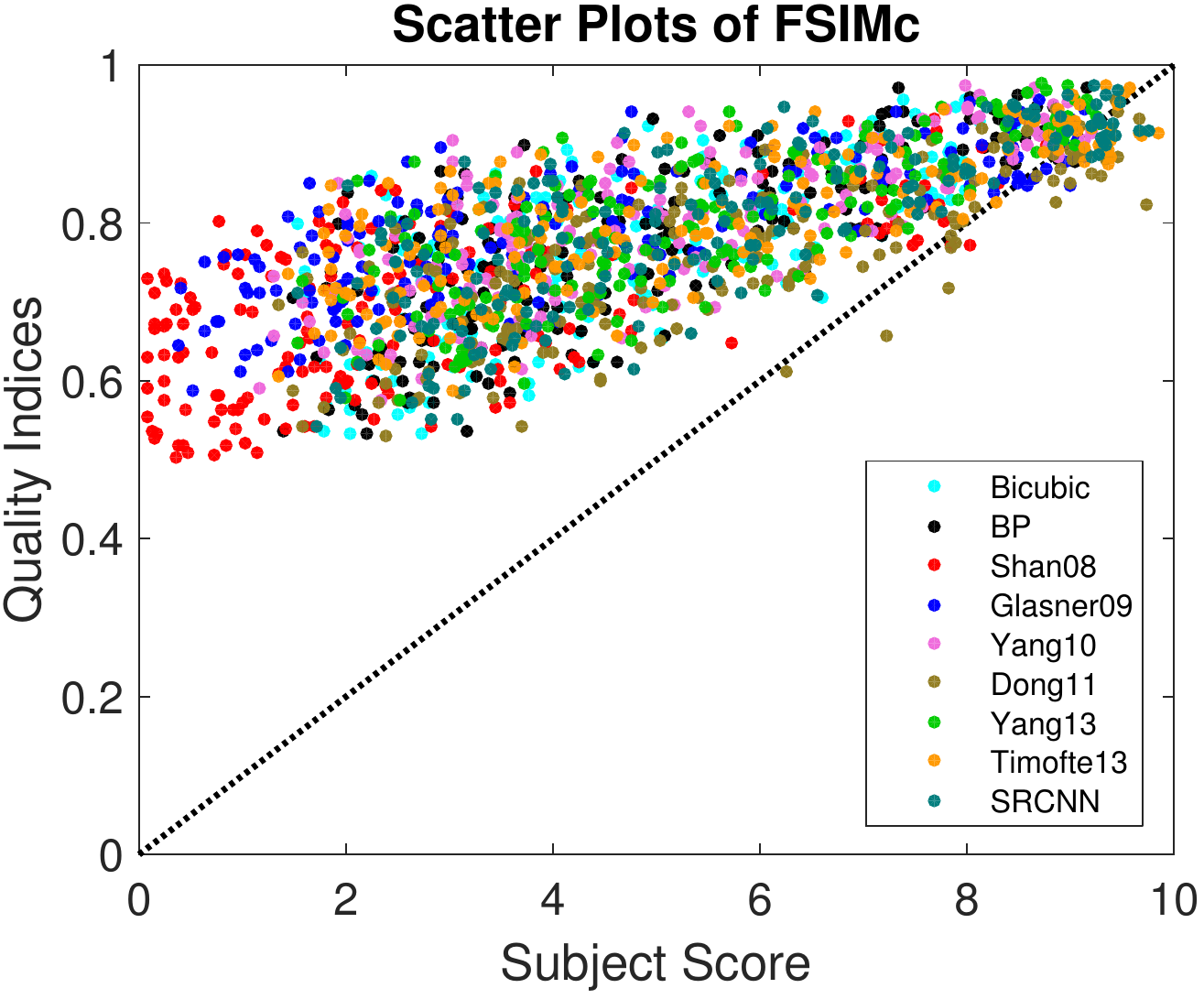} &
		\includegraphics[width=.33\textwidth]{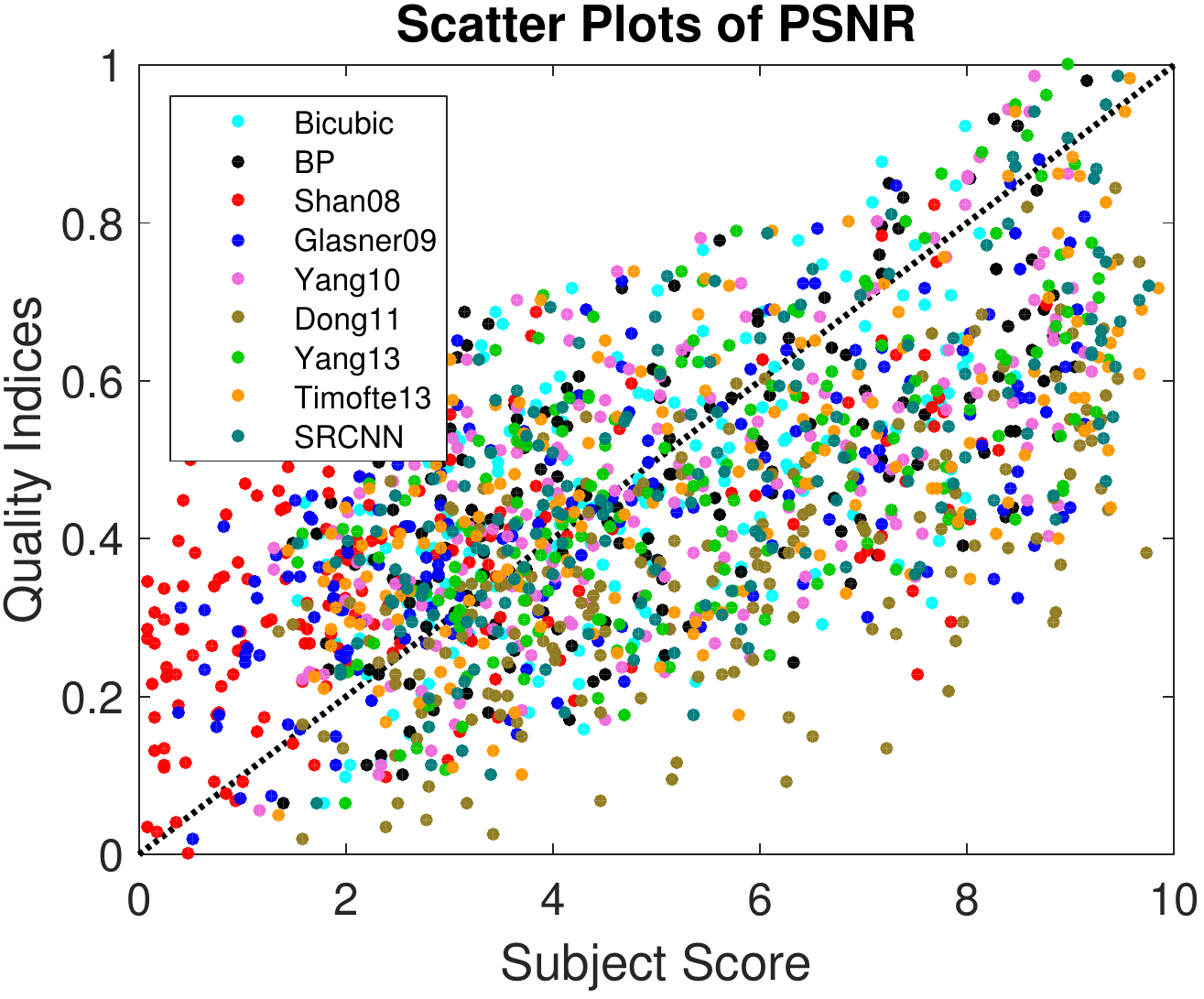} \\
	\end{tabular}
	\caption{Quality indices generated by different methods to perceptual scores.
		The proposed metric and other no-reference baseline methods (except DIVINE and BIQI) are leaned under 5-fold cross validation. 
		A metric matches visual perception well if the distribution is compact
		and spreads out along the diagonal.
	}
	\label{fig:corrDistribution}
\end{figure}

For fair comparisons, we generate the IQA indices from 11 state-of-the-art methods
including: (1) six no-reference metrics:
BRISQUE~\cite{DBLP:journals/tip/MittalMB12}, BLIINDS~\cite{DBLP:journals/tip/SaadBC12},
DIVINE~\cite{DBLP:journals/tip/MoorthyB11},
BIQI~\cite{DBLP:journals/spl/MoorthyB10},
CORNIA~\cite{DBLP:conf/cvpr/YeKKD12}, and CNNIQA~\cite{DBLP:conf/cvpr/KangYLD14};
(2) one semi-reference metric: NSSA~\cite{DBLP:conf/icip/YeganehRW12};
and (3) four full-reference metrics:
IFC~\cite{DBLP:journals/tip/SheikhBV05},
SSIM~\cite{DBLP:journals/tip/WangBSS04}, FSIM~\cite{DBLP:journals/tip/ZhangZMZ11}, and PSNR. 
As the no-reference metrics are originally designed
to measure image degradations, e.g., noise, compression and fading, rather than for SR evaluation, we retrain them on our SR dataset using the same validation schemes. 
Note that both the DIVINE and BIQI metrics apply intermediate steps to estimate  
specific types of image degradations \cite{DBLP:journals/tip/SheikhSB06} 
for image quality assessment. 
However, SR degradation is not considered in 
any type of degradations in \cite{DBLP:journals/tip/SheikhSB06}. 
We directly regress the features generated by DIVINE and BIQI methods 
to the perceptual scores but this approach is not effective as
the quality scores for different SR images are almost the same. 
We thus report the original results using the DIVINE and BIQI indices 
without retraining on our dataset. 
We empirically tune the parameters to obtain best performance during retraining. 
The NSSA metric is designed for
evaluating SR images. The other four full-reference metrics are widely
used in SR evaluation although they are not designed for SR. 
Figure~\ref{fig:corrDistribution} shows the correlation between subjective scores and
IQA indices.
Table~\ref{tb:cv},  \ref{tb:leaveimage} and \ref{tb:leavemethod} quantitatively compares the Spearman rank
correlation coefficients. 
In addition, we compare the original results of BRISQUE, BLIINDS, CORNIA, and CNNIQA in Table \ref{tb:notrain} and Figure \ref{fig:corr_notrain}. 
Without retraining on our SR dataset, these metrics generally perform worse. 
This shows the contributions of this work by developing a large-scale SR image dataset 
and carrying out large-scale subject studies on these SR images. 
%
Note that we do not present the results of NSSA in Table \ref{tb:notrain} and Figure \ref{fig:corr_notrain} 
as the learned data file of the NSSA metric is not publicly available. 

\begin{table}
	\caption{Spearman rank correlation coefficients~\cite{hogg2005introduction}
		(metric with higher coefficient matches perceptual score better). The compared no-reference metrics are not retrained on our SR dataset. Bold: best; underline: second best.} 
	\label{tb:notrain}
	\centering
	\resizebox{.9\textwidth}{!}{
	\setlength{\tabcolsep}{0em}
	\small
	\begin{tabular}{| p{6em} || p{5em}<{\centering}|p{5em}<{\centering}|p{5em}<{\centering}||p{5em}<{\centering}|p{5em}<{\centering}|p{5em}<{\centering}|p{5em}<{\centering}|}
		\hline
		&  Ours & Ours & Ours & BRISQUE & BLIINDS  & CORNIA & CNNIQA \\
		& \textit{5-fold CV} & \textit{image-out} & \textit{method-out} & \cite{DBLP:journals/tip/MittalMB12} 
		& \cite{DBLP:journals/tip/SaadBC12}
		& \cite{DBLP:conf/cvpr/YeKKD12}
		& \cite{DBLP:conf/cvpr/KangYLD14}
		\\
		\hline\hline
		~ Bicubic   & \textbf{0.933} & 0.805       & {\ul 0.932}    & 0.850 & 0.929 & 0.893       & 0.927       \\
		~ BP        & {\ul 0.966}    & 0.893       & \textbf{0.967} & 0.934 & 0.953 & 0.938       & 0.931       \\
		~ Shan08    & \textbf{0.891} & 0.800       & {\ul 0.803}    & 0.534 & 0.471 & 0.799       & 0.842       \\
		~ Glasner09 & \textbf{0.931} & 0.867       & 0.913          & 0.677 & 0.805 & 0.817       & {\ul 0.896} \\
		~ Yang10    & \textbf{0.968} & 0.904       & {\ul 0.965}    & 0.834 & 0.895 & 0.914       & 0.938       \\
		~ Dong11    & \textbf{0.954} & 0.875       & 0.932          & 0.774 & 0.780 & 0.917       & {\ul 0.936} \\
		~ Yang13    & \textbf{0.958} & 0.885       & {\ul 0.944}    & 0.716 & 0.845 & 0.911       & 0.934       \\
		~ Timofte13 & \textbf{0.930} & 0.815       & 0.774          & 0.760 & 0.849 & 0.898       & {\ul 0.906} \\
		~ SRCNN     & \textbf{0.949} & 0.904       & 0.933          & 0.771 & 0.806 & {\ul 0.908} & 0.924       \\\hline
		~ Overall   & \textbf{0.931} & {\ul 0.852} & 0.848          & 0.644 & 0.763 & 0.809       & 0.833       \\\hline	
	\end{tabular}
		}
\end{table}

\subsection{Discussion}
As shown in Table~\ref{tb:cv}-\ref{tb:leavemethod} and Figure~\ref{fig:corrDistribution},
the proposed method performs favorably against the state-of-the-art
IQA methods, e.g., the overall quantitative correlation with perceptual
scores is 0.931  under 5-fold cross validation.
The leave-image-out and leave-method-out validations are more challenging since
they take into account the independence of image contents and SR algorithms.
In the leave-image-out setting, the training and test sets do not
contain SR images generated from the same reference image.
In the leave-method-out setting, the SR images in training and test sets
are generated by different SR algorithms.
Table~\ref{tb:leaveimage} and \ref{tb:leavemethod} show that the proposed metric 
performs well against existing IQA methods in these two
challenging validations. 
Note that the proposed metric performs best in the 5-fold cross validation 
as it learns from perceptual scores and favors prior information from 
image contents and SR algorithms for training.

\begin{figure}[!t]
	\centering
	\small
	\setlength{\tabcolsep}{0em}
	\begin{tabular}{ccc}
		\includegraphics[width=.33\textwidth]{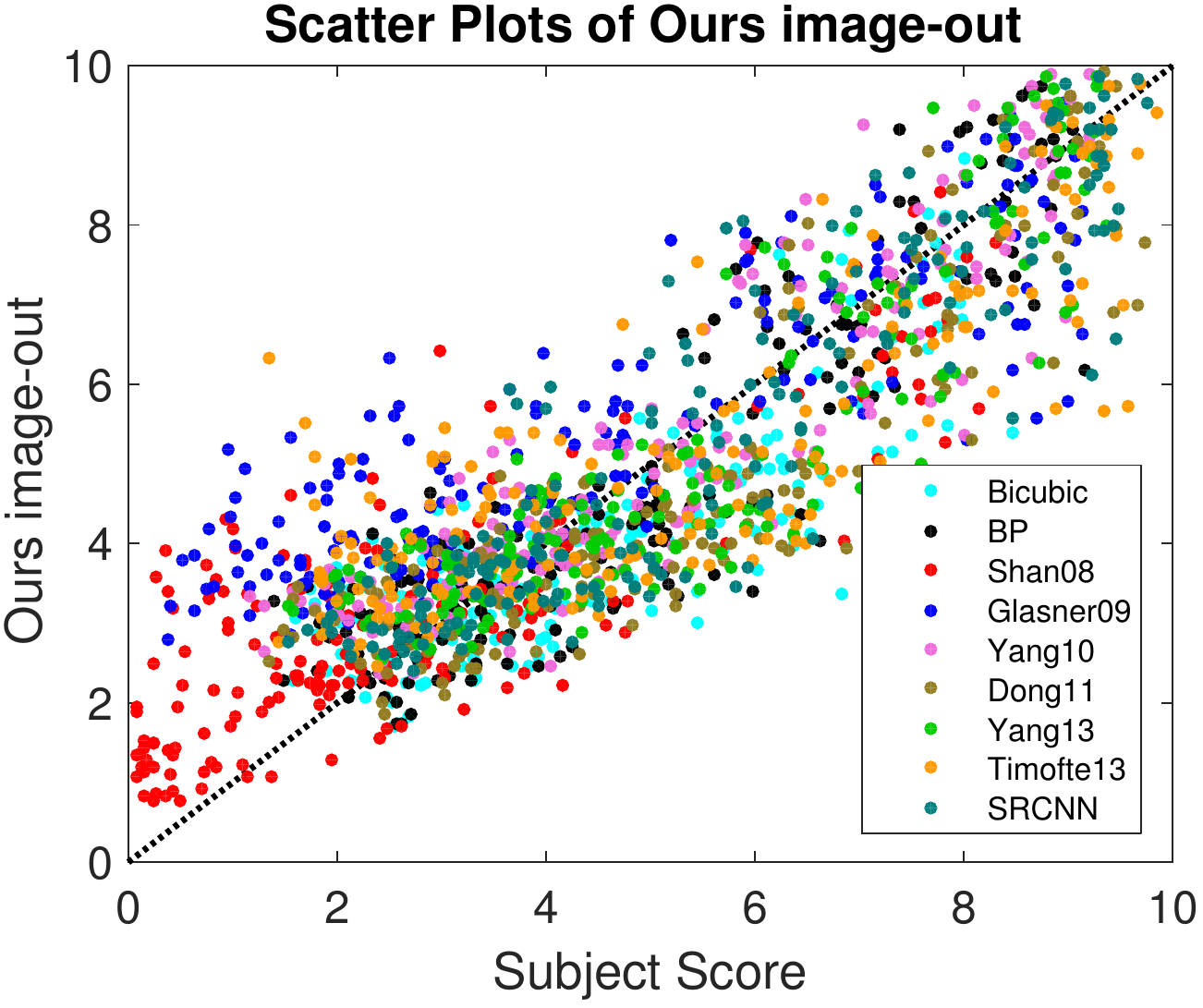}&
		\includegraphics[width=.33\textwidth]{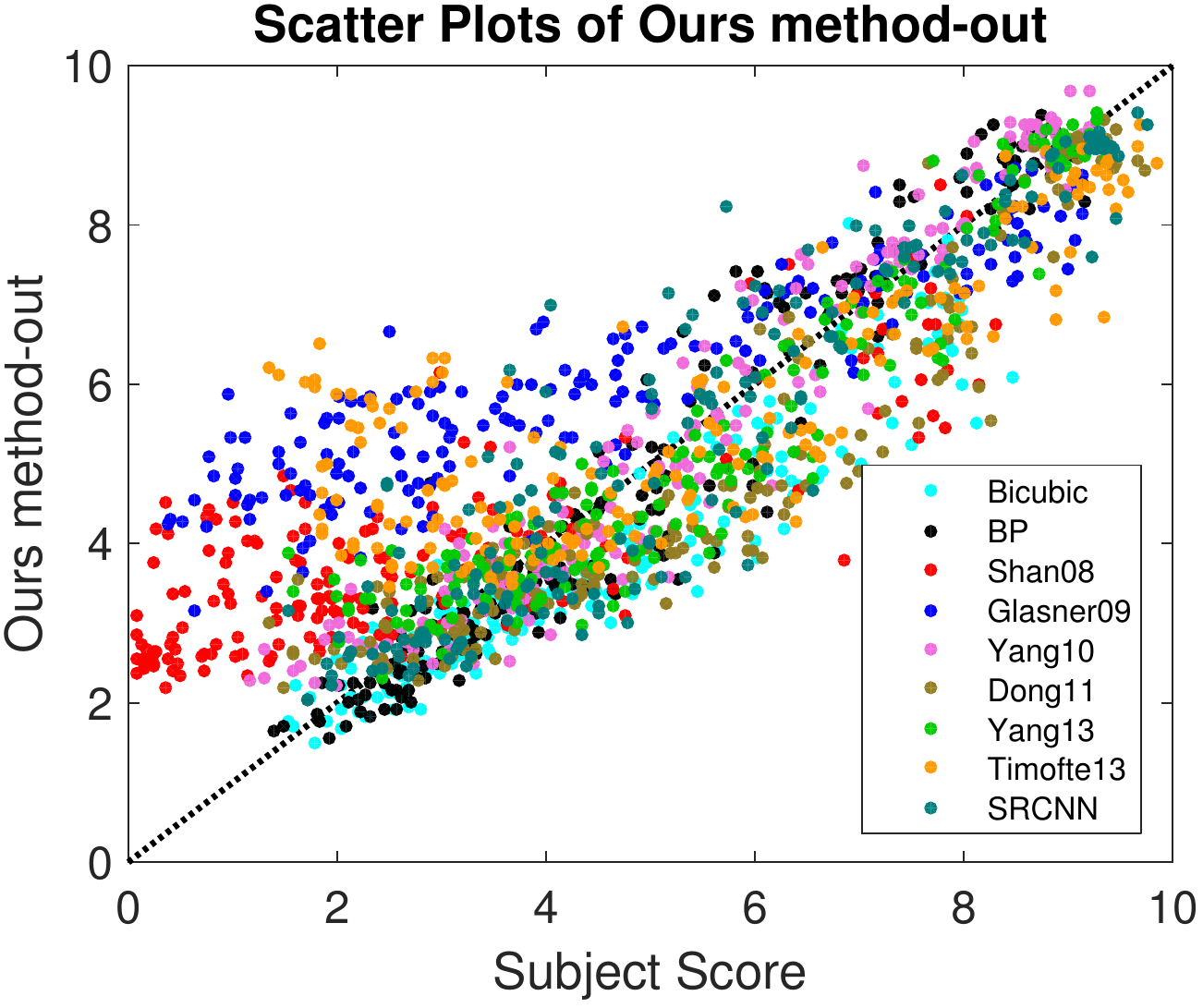}&
		\includegraphics[width=.33\textwidth]{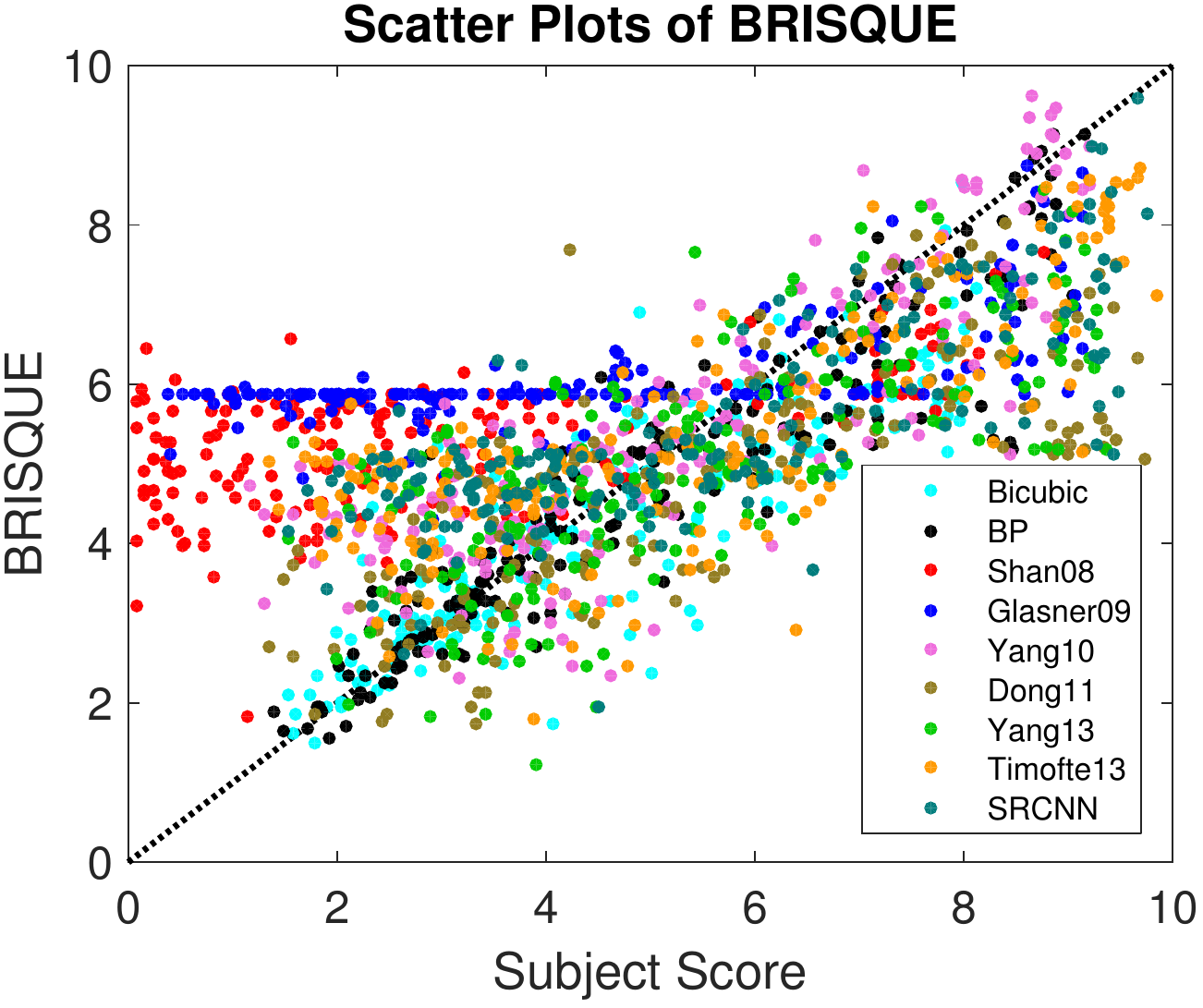} \\
		(a) & (b) & (c) \\
		\includegraphics[width=.33\textwidth]{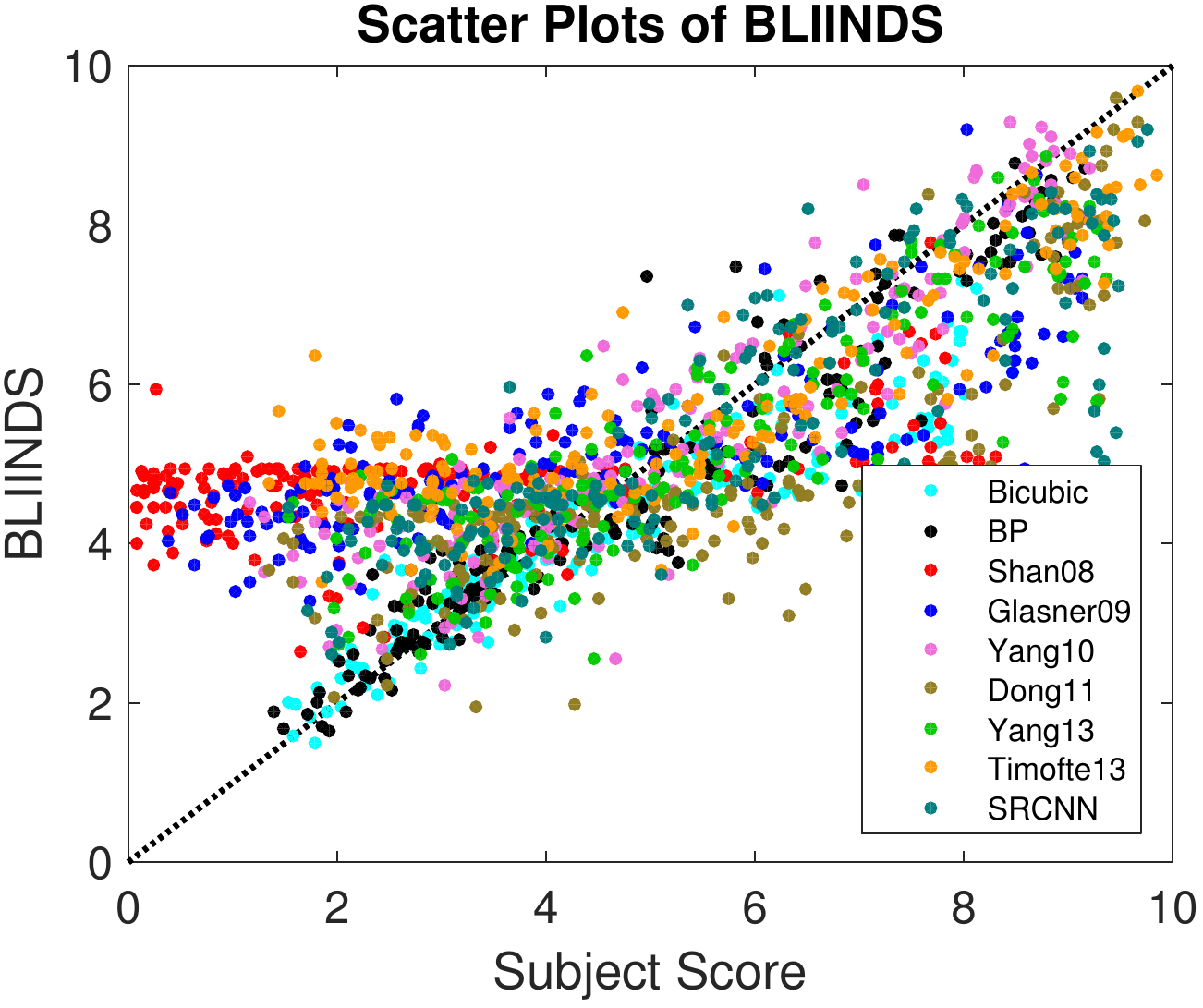} &
		\includegraphics[width=.33\textwidth]{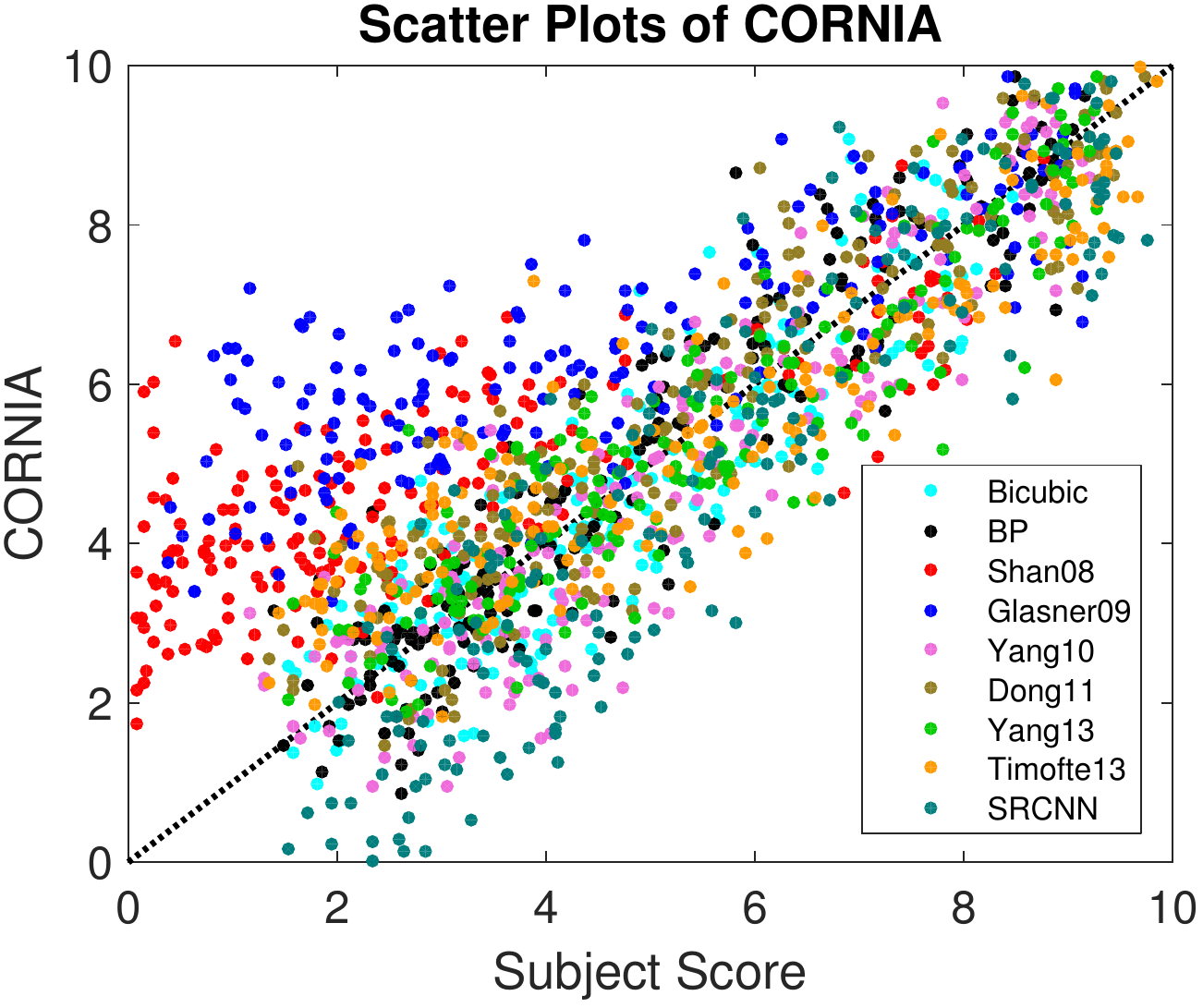} &
		\includegraphics[width=.33\textwidth]{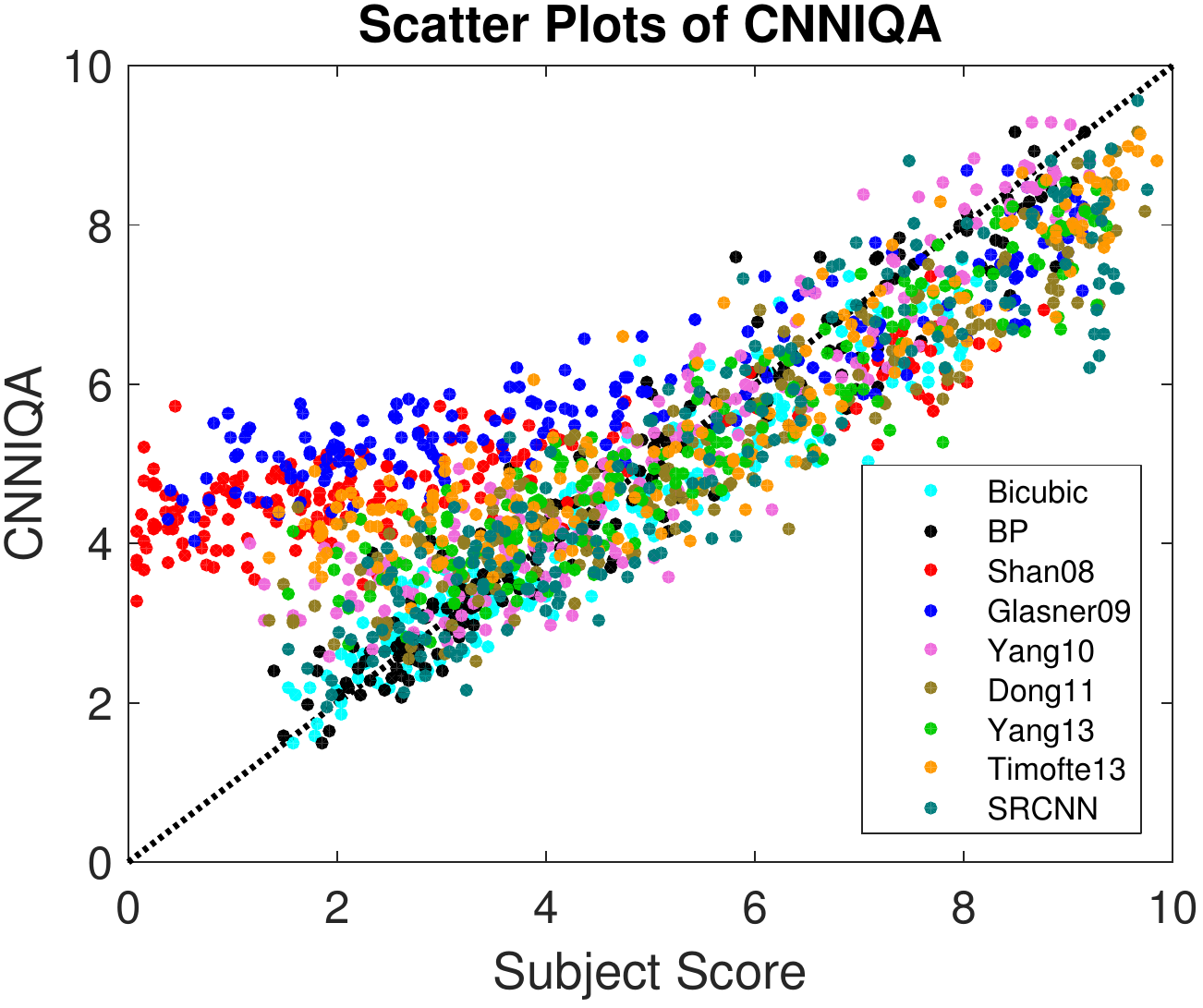} \\
		(d) & (e) & (f) \\
	\end{tabular}
	\caption{Quality indices generated by different methods to perceptual scores.
		(a)-(b): Proposed metric under \textit{leave-image-out} and \textit{leave-method-out} validation schemes. (c)-(f): Original baseline no-reference algorithms without retraining on our SR dataset. The proposed metric under two challenging validations still performs well against state-of-the-art metrics. 
		A metric matches visual perception well if the distribution is compact
		and spread out along the diagonal. 
	}
	\label{fig:corr_notrain}
\end{figure}

The six evaluated no-reference IQA metrics, BRISQUE, BLIINDS, DIVINE,
BIQI, CORNIA, and CNNIQA, are not originally designed for SR. We retrain them (except DIVINE and BIQI) on our own SR dataset.
%
For DIVINE and BIQI, we present 
the reported results as the performance of these methods by retraining on our dataset is significantly worse. 
The reason is that these two metrics apply intermediate steps to quantify specific image distortions in \cite{DBLP:journals/tip/SheikhSB06} rather than SR. 
Table~\ref{tb:cv} shows that for most SR algorithms, the DIVINE or BIQI metrics do not match human perception well.
The retrained BRISQUE and BLIINDS metrics perform well against DIVINE and BIQI.
We note that some of the features used by the BRISQUE and BLIINDS metrics are similar to 
the proposed DCT and GSM features. 
However, both BRISQUE and BLIINDS  metrics 
are learned from one support vector regression (SVR) model \cite{libsvm}, 
which are less robust to the outliers of perceptual scores than the 
random forest regression (RFR) model.  Figure~\ref{fig:corrDistribution} shows that their quality scores scatter more than close to the
diagonal.
The CORNIA method learns a codebook from an auxiliary
dataset~\cite{DBLP:journals/jei/LarsonC10}  
containing various image distortions. 
The coefficients of densely sampled patches from a test image are computed 
based on the codebook as features.
Table \ref{tb:cv} shows that the CORNIA metric achieves 
second best results among all the baseline algorithms. 
The proposed metric performs favorably against CORNIA
due to the effective two-stage regression model based on RFRs. 
While CORNIA only relies on one single SVR. 
The CNNIQA metric uses convolutional neural network to assess the image quality, however,
it does not perform as well as the proposed method.
The can be explained by insufficient amount of training examples.
Overall, the proposed method exploits both global and local
statistical features  specifically designed to account for SR artifacts. 
Equipped with a novel two-stage regression model, i.e., three independent random forests are regressed on extracted three types of
features and their outputs are linearly regressed with perceptual
scores,  our metric is more robust to outliers than the compared IQA
methods, which are based on one single regression model (e.g., SVR or CNN).

Although the semi-reference NSSA method is designed for evaluating SR images 
and extracts both frequency and spatial features,  it does not perform well
as shown in Figure~\ref{fig:corrDistribution} and Table~\ref{tb:cv}-\ref{tb:leavemethod}.
This is because the features used in the NSSA method are two-dimension coefficients and 
their regressor is based on a simple linear model. 
The quality indices computed by weight-averaging two coefficients are 
less effective for evaluating the quality of SR images generated by the state-of-the-art SR
methods.

\begin{figure}[!t]
	\centering
	\setlength{\tabcolsep}{.1em}
	\begin{tabular}{cccc}
		\includegraphics[width=.24\textwidth]{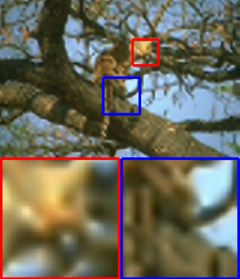} & 
		\includegraphics[width=.24\textwidth]{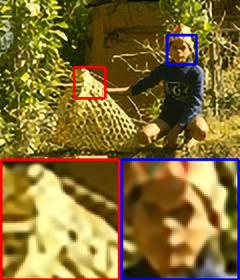} & 
		\includegraphics[width=.24\textwidth]{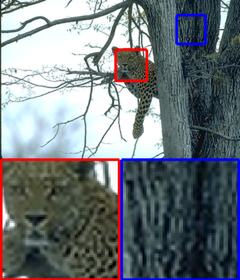} &
		\includegraphics[width=.24\textwidth]{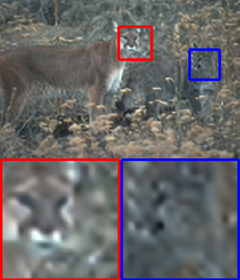} \\
		Shan08~\cite{DBLP:journals/tog/ShanLJT08} & 
		Glasner09~\cite{Glasner2009} & 
		Yang13~\cite{Yang13_ICCV_Fast} &
		Dong11~\cite{DBLP:journals/tip/DongZSW11}\\ 
		2.68 / 2.68 &  4.70 / 4.70 & 8.65 / 8.65 &  5.17 / 5.18\\ 
		$s=6$, $\sigma=1.8$  & $s=4$, $\sigma=1.2$  &  $s=2$, $\sigma=0.8$ & $s=4$, $\sigma=1.2$\\
	\end{tabular}
	\caption{Four best cases using the proposed metric to evaluate the quality of SR images under the 5-fold cross validation. 
		The left / right values under each image are the predicted score and the perceptual score respectively.}
	\label{fig:bw1}
\end{figure}

\begin{figure}[!t]
	\centering
	\setlength{\tabcolsep}{.1em}
	\begin{tabular}{cccc}
		\includegraphics[width=.24\textwidth]{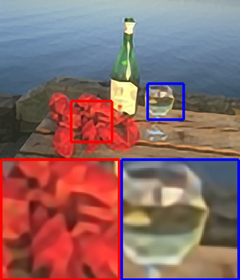} & 
		\includegraphics[width=.24\textwidth]{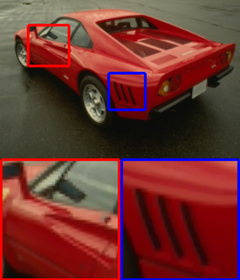} &
		\includegraphics[width=.24\textwidth]{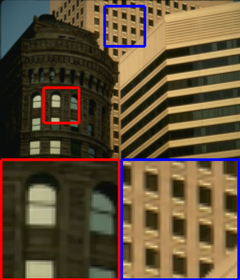} & 
		\includegraphics[width=.24\textwidth]{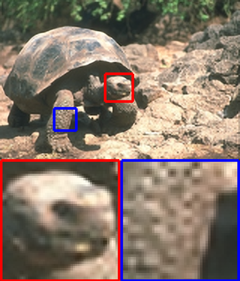}  \\
		Glasner09~\cite{Glasner2009} & Shan08~\cite{DBLP:journals/tog/ShanLJT08} & Shan08~\cite{DBLP:journals/tog/ShanLJT08} & Yang13~\cite{Yang13_ICCV_Fast} \\
		0.95 / 4.95 & 8.15 / 5.30 & 0.93 / 6.85 & 2.55 / 4.48 \\
		$s=5$, $\sigma=1.6$   & $s=2$, $\sigma=0.8$ & $s=2$, $\sigma=0.8$  & $s=3$, $\sigma=1.0$  \\
	\end{tabular}
	\caption{Four worst cases using the proposed metric to evaluate the quality of SR images under the 5-fold cross validation. 
		The left / right values under each image are the predicted score and the perceptual score respectively.}
	\label{fig:bw2}
\end{figure}

For the cases when ground truth HR images are available,
the proposed method performs favorably against four widely used
full-reference quality metrics including
PSNR, SSIM~\cite{DBLP:journals/tip/WangBSS04}, IFC~\cite{DBLP:journals/tip/SheikhBV05}, and FSIM~\cite{DBLP:journals/tip/ZhangZMZ11}. 
The PSNR metric performs poorly since the pixel-wise difference measurement
does not effectively account for the difference in visual perception (See
Table~\ref{tb:cv} and Figure~\ref{fig:corrDistribution}). 
For example, an SR image with slight misalignment from the ground truth
data appears similarly in terms of visual perception, but the PSNR
value decreases significantly. 
The SSIM method performs better than PSNR as it aims to mimic human
vision and computes perceptual similarity between SR and ground truth images by using patches instead of pixels.
However, the SSIM metric
favors the false sharpness on the SR images generated by Shan08 and
Glasner09 and overestimates the corresponding quality scores as shown in
Figure~\ref{fig:corrDistribution}.
The FSIM metric is less effective in evaluating the SR performance either.
The IFC method is also designed to match visual perception and
generally performs well for SR images~\cite{Yang14_ECCV}. 
Nonetheless, its indices are less accurate for some SR images
(Figure~\ref{fig:corrDistribution}). 
This can be explained by the fact that the IFC metric is limited by
local frequency features. 
In other words, the IFC metric does not take global frequency and
spatial properties into account, and fails to distinguish them.
Thus it may underestimate the quality of SR images 
(See the dots cluster below the diagonal in the last sub figure of
Figure~\ref{fig:corrDistribution}).

We present four best and worse cases using our metric with 5-fold cross validation 
to predict the quality of SR images in Figure~\ref{fig:bw1} and Figure~\ref{fig:bw2}. 
The reasons that cause the worst cases in Figure 15 can be explained by several factors. 
For the first, third and fourth SR images, the proposed metric gives low quality scores due to the fact that human subjects do not always favor oversharp SR images (see also the discussion in Table 1 in the manuscript). 
For the second image, the richer high-frequency contents affect 
the proposed metric to compute the high score. 

Overall, the proposed metric performs favorably against the
state-of-the-art methods, which can be attributed to two reasons. 
First, the proposed metric uses three sets of discriminative low-level
features from the spatial and frequency domains to describe SR
images. 
Second, an effective two-stage regression model is more robust to outliers 
for learning from perceptual scores collected in our large-scale
subject studies. 
In contrast, existing methods neither learn from perceptual 
scores nor design effective features with focus on representing SR
images.
%
The proposed metric is implemented in Matlab on a machine with 
an Intel i5-4590 3.30 GHz CPU and 32 GB RAM. 
We report the average run time (in seconds) as follows, 
ours: 13.31, BRISQUE: 0.14, BLIINDS: 23.57, DIVINE: 9.51, BIQI: 1.21,  CORNIA: 3.02, CNNIQA: 12.68, NSSA: 0.28,  IFC: 0.61, SSIM: 0.13, FSIM: 0.18, and PSNR: 0.02.

\begin{figure}
	\centering
	\small
	\setlength{\tabcolsep}{.2em}
	\begin{tabular}{cccc}
		\includegraphics[height=0.395\textwidth]{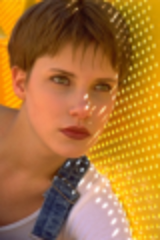} &
		\includegraphics[height=0.395\textwidth]{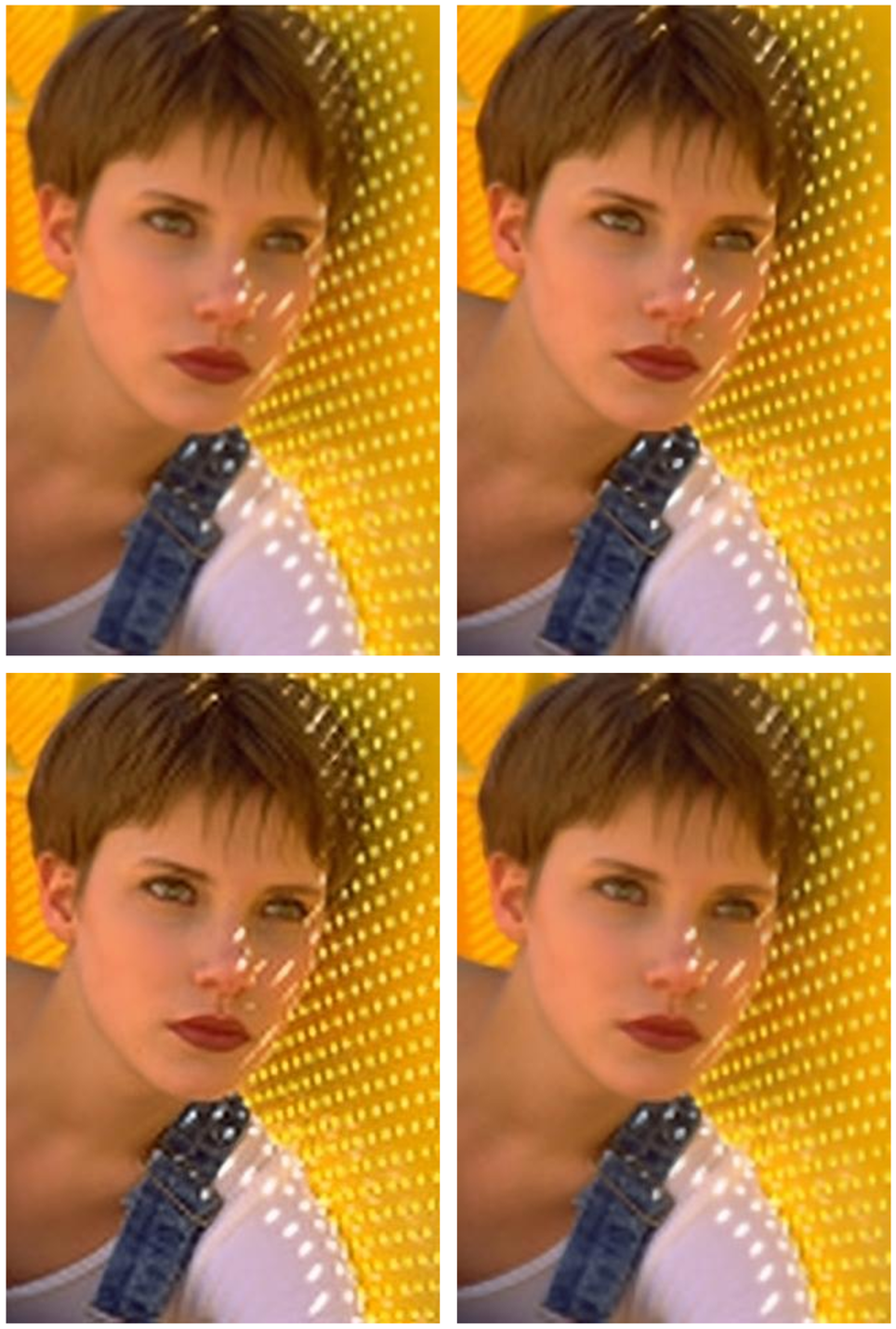} &
		\includegraphics[trim = 0mm 2mm 0mm 0mm, clip, height=0.395\textwidth]{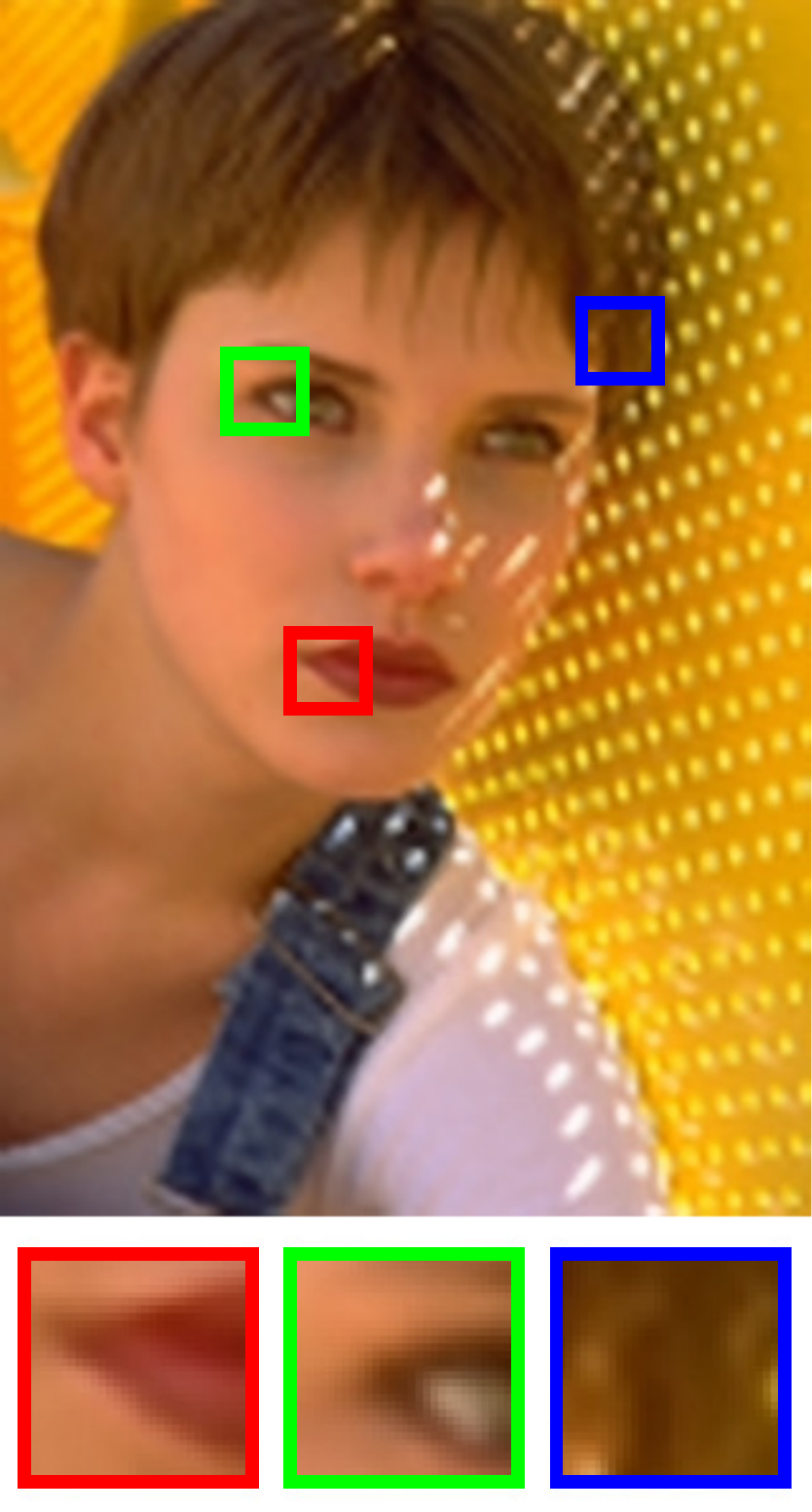} &
		\includegraphics[trim = 0mm 2mm 0mm 0mm, clip, height=0.395\textwidth]{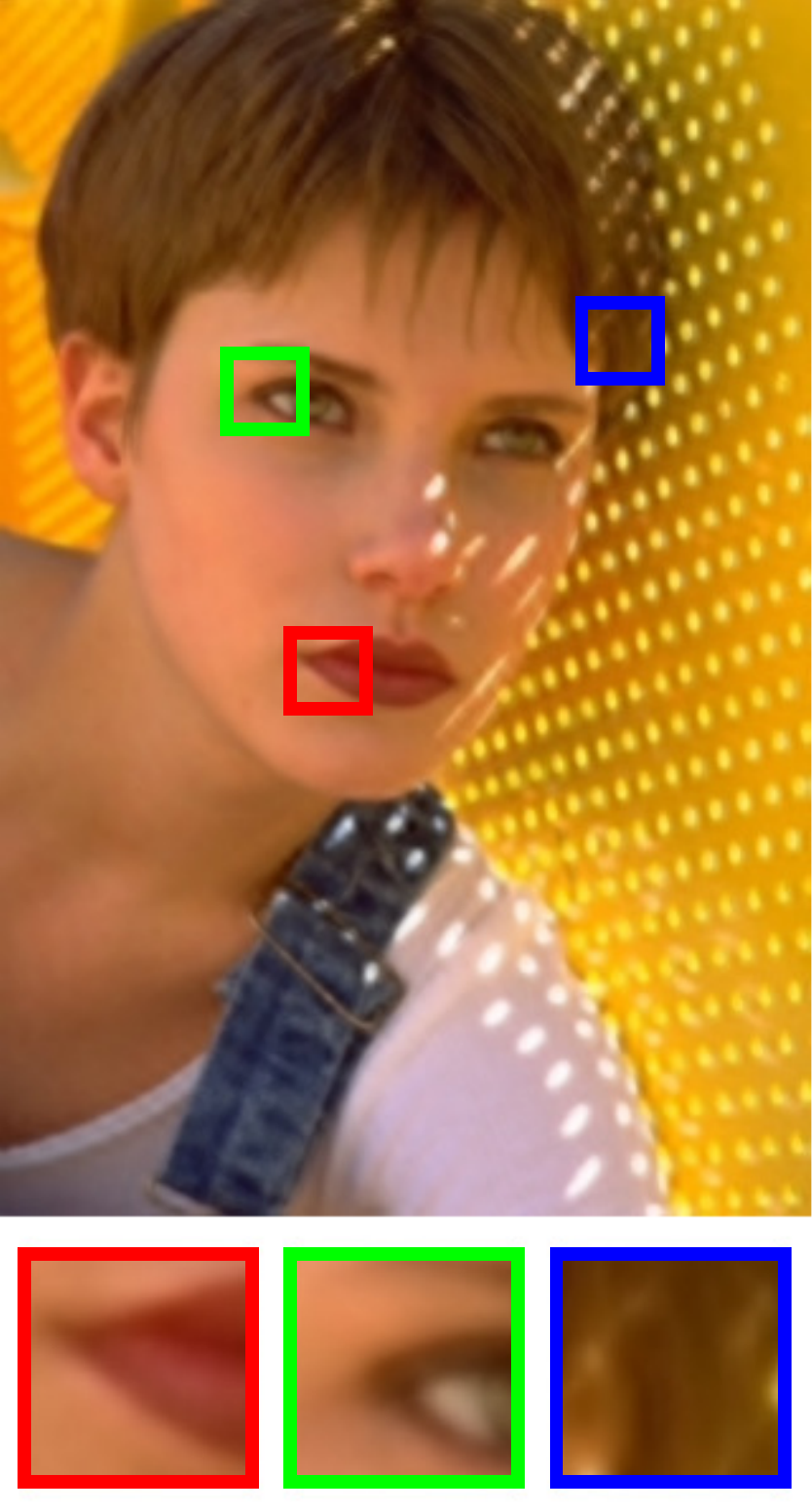} \\
		(a) 4.01 & (b) 4.82/4.70/4.69/4.50 & (c) 4.61 & (d) 4.78 \\
	\end{tabular}
	\caption{Perception guided SR results (best viewed on a high-resolution displayer) 
		with quality scores predicted by the proposed metric.
		(a) Input LR image ($s=4,\sigma=1.8$).
		(b) Selected best SR images with the Dong11, Yang13, Timofte13 
		and Yang10 methods using the proposed metric.
		(c) $3\times 3$ grid integration. (d) Pixel-level integration. }
	\label{fig:srresults}
\end{figure}

\begin{figure} 
	\centering 
	\small
	\setlength{\tabcolsep}{.05em}
	\begin{tabular}{ccc} 
		\includegraphics[width=.32\textwidth]{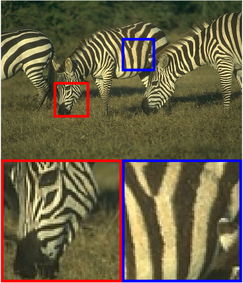} & 
		\includegraphics[width=.32\textwidth]{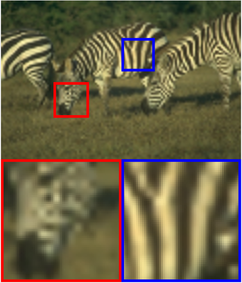} & 
		\includegraphics[width=.32\textwidth]{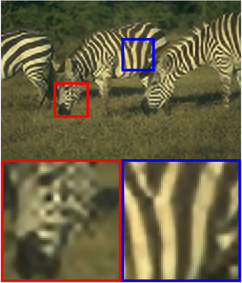} \\ 
		Ground truth & Input LR image & Ours (6.02) \\ 
		\includegraphics[width=.32\textwidth]{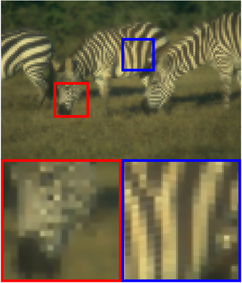} & 
		\includegraphics[width=.32\textwidth]{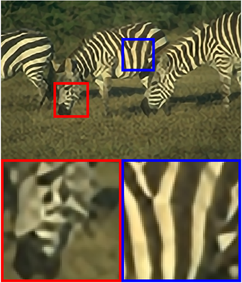} & 
		\includegraphics[width=.32\textwidth]{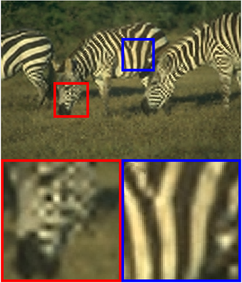} \\ 
		Shan08~\cite{DBLP:journals/tog/ShanLJT08}~(5.72) & SRCNN~\cite{DBLP:conf/eccv/DongLHT14}~(5.32) & Yang13~\cite{Yang13_ICCV_Fast}~(5.16) \\ 
		\includegraphics[width=.32\textwidth]{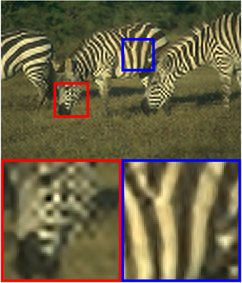} & 
		\includegraphics[width=.32\textwidth]{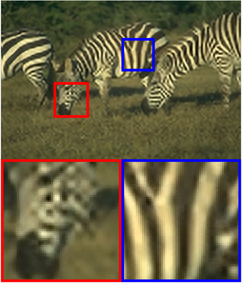} & 
		\includegraphics[width=.32\textwidth]{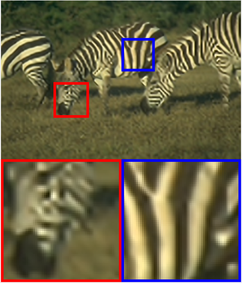} \\ 
		Glasner09~\cite{Glasner2009}~(4.78) & Timofte13~\cite{DBLP:conf/iccv/TimofteDG13}~(4.65) & Dong11~\cite{DBLP:journals/tip/DongZSW11}~(4.61) \\ 
	\end{tabular} 
	\caption{Visual comparison of SR results. The input low resolution images are generated using (1)  with $s=4$ and $\sigma=1.2$. We show the best 6 results based on their quality scores in parentheses predicted by the proposed metric, and select the best 4 algorithms to integrate our SR results.} 
	\vspace{1em}
	\label{fig:sr007} 
\end{figure}

\begin{figure}[!t]
	\centering
	\small 
	\setlength{\tabcolsep}{.1em}
	\begin{tabular}{ccc} 
		\includegraphics[width=.32\textwidth]{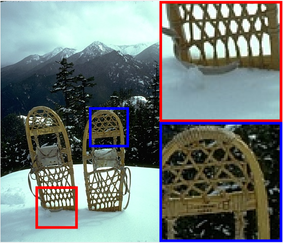} & 
		\includegraphics[width=.32\textwidth]{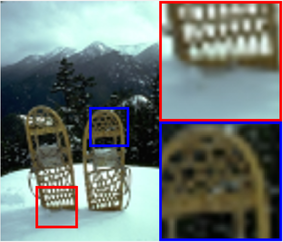} & 
		\includegraphics[width=.32\textwidth]{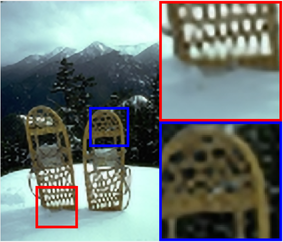} \\ 
		Ground truth & Input LR image & Ours (5.81) \\ 
		\includegraphics[width=.32\textwidth]{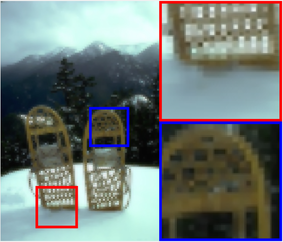} & 
		\includegraphics[width=.32\textwidth]{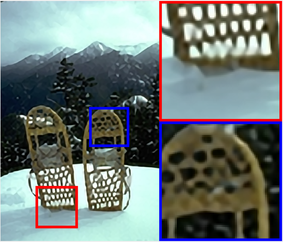} & 
		\includegraphics[width=.32\textwidth]{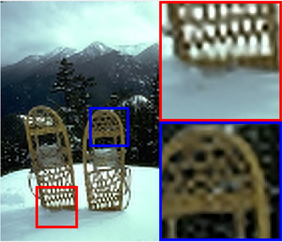} \\ 
		Shan08~\cite{DBLP:journals/tog/ShanLJT08}~(5.48) & Yang10~\cite{DBLP:journals/tip/YangWHM10}~(5.35) & Glasner09~\cite{Glasner2009}~(5.02) \\ 
		\includegraphics[width=.32\textwidth]{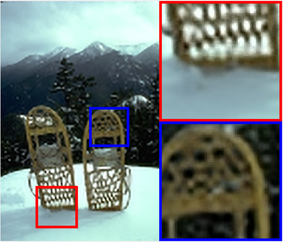} & 
		\includegraphics[width=.32\textwidth]{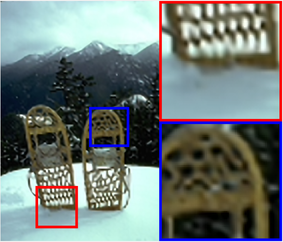} & 
		\includegraphics[width=.32\textwidth]{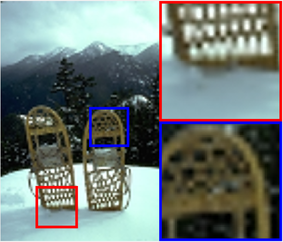} \\ 
		Timofte13~\cite{DBLP:conf/iccv/TimofteDG13}~(4.64) & Dong11~\cite{DBLP:journals/tip/DongZSW11}~(4.63) & BP~\cite{DBLP:journals/cvgip/IraniP91}~(4.58) \\ 
	\end{tabular} 
	\caption{Visual comparison of SR results. The input low resolution images are generated using (1) with $s=4$ and $\sigma=1.2$. We show the best 6 results based on their quality scores in innermost parentheses predicted by the proposed metric, and select the best 4 algorithms to integrate our SR results.} 
	\label{fig:sr010} 
\end{figure} 

\section{Perception Guided Super-Resolution} 
Given an LR input image, we can apply different SR algorithms to
reconstruct HR images and use the proposed metric to 
automatically select the best result. 
Figure~\ref{fig:SRimage} shows such an example where 
the SR image generated by the Timofte13 method has the highest quality 
score using the proposed metric (See Figure~\ref{fig:SRimage}(i)) 
and is thus selected as the HR restoration output. 
Equipped with the proposed metric, we can also select the best
local regions from multiple SR images and integrate them into a
new SR image. 
Given a test LR image, we apply aforementioned 9 SR algorithms 
to generate 9 SR images. 
We first divide each of them into a $3\times3$ grid of regions. 
We compute their quality scores based on the proposed metric and stitch 
the best regions to generate a new SR image (See Figure~\ref{fig:srresults}(c)). 
For better integration, we densely sample overlapping patches of 
$11\times11$ pixels. 
We then apply the proposed metric on each patch and 
compute an evaluation score of each pixel of that SR image. 
For each patch, we select the one from all results with highest quality scores and 
stitch all the selected patches together using the graph cut and
Poisson blending~\cite{DBLP:journals/tog/PerezGB03} method
(See Figure~\ref{fig:srresults}(d)). 
It is worth noting that the proposed metric can be used to select SR
regions with high perceptual scores from which a high-quality HR image
is formed. 
Figure~\ref{fig:sr007} and Figure~\ref{fig:sr010} show two more pixel-level integrated SR results, which retain most edges and render smooth contents as well. 
The integrated SR results effectively exploit the merits of state-of-the-art SR algorithms, 
and show better visual quality.  
\section{Conclusion}

In this paper, we propose a novel no-reference IQA algorithm 
to assess the visual quality of SR images by learning
perceptual scores collected from large-scale
subject studies. 
The proposed metric regress three types of low-level statistical
features extracted from SR images to perceptual scores.
Experimental results demonstrate that the proposed metric performs
favorably against state-of-the-art quality assessment methods
for SR performance evaluation.

\section*{References}

\bibliographystyle{elsarticle-num} 
\bibliography{bib_metric}
\end{document}